\documentclass[lettersize,journal]{IEEEtran}
\usepackage{amsmath,amsfonts}
\usepackage{algorithmic}
\usepackage{algorithm}
\usepackage{array}
\usepackage[caption=false,font=normalsize,labelfont=sf,textfont=sf]{subfig}
\usepackage{textcomp}
\usepackage{stfloats}
\usepackage{url}
\usepackage{verbatim}
\usepackage{graphicx}
\usepackage{cite}

\usepackage[utf8]{inputenc}
\usepackage[T1]{fontenc}
\usepackage{wrapfig}
\usepackage{subfig}
\usepackage{adjustbox}
\usepackage{rotating}
\usepackage{multirow}
\usepackage{array}
\usepackage{ragged2e}
\usepackage{hhline}
\usepackage{colortbl}
\usepackage{tabularx}
\usepackage{xspace}
\usepackage{makecell}
\usepackage{rotating}
\usepackage{multirow}
\usepackage{booktabs}

\usepackage{bm}

\usepackage{color}
\usepackage{rotating}
\usepackage{tabularray}

\newcommand{\thickhline}{%
    \noalign {\ifnum 0=`}\fi \hrule height 1.3pt
    \futurelet \reserved@a \@xhline
}
\newcommand{\ie}{\textit{i.e.}\xspace}
\newcommand{\eg}{\textit{e.g.}\xspace}
\newcommand{\etal}{\textit{et al.}\xspace}

\newcolumntype{?}{!{\vrule width 1.5pt}}
\DeclareMathOperator*{\argmin}{arg\,min}

\begin{document}

\title{Survey of Deep Learning and Physics-Based Approaches in Computational Wave Imaging}

\author{Youzuo Lin$^\star$, Shihang Feng, James Theiler, Yinpeng Chen, Umberto Villa, Jing Rao, \\ John Greenhall, Cristian Pantea, Mark A. Anastasio, and Brendt Wohlberg

\thanks{$^{\star}$\textbf{Corresponding author:} \textbf{Youzuo Lin} (\texttt{yzlin@unc.edu}).}
\thanks{Y. Lin and S. Feng are with the University of North Carolina at Chapel Hill, Chapel Hill, NC 27599, USA.}
\thanks{J. Theiler, J. Greenhall, C. Pantea, and B. Wohlberg are with Los Alamos National Laboratory, Los Alamos, NM, 87545, USA.}
\thanks{Y. Chen is with Google DeepMind, Seattle, WA, 98103, USA.}
\thanks{U. Villa is with the University of Texas at Austin, Austin, TX, 78712, USA.}
\thanks{J. Rao is with University of New South Wales, Canberra, ACT 2600, Australia.}
\thanks{M. Anastasio is with Washington University in St. Louis, St. Louis, MO, 63130, USA.}

}



\maketitle

\begin{abstract}
Computational wave imaging (CWI) extracts hidden structure and physical properties of a volume of material by analyzing wave signals that traverse that volume. Applications include seismic exploration of the  Earth's subsurface, acoustic imaging and non-destructive testing in material science, and ultrasound computed tomography in medicine. Current approaches for solving CWI problems can be divided into two categories: those rooted in traditional physics, and those based on deep learning. Physics-based methods stand out for their ability to provide high-resolution and quantitatively accurate estimates of acoustic properties within the medium. However, they can be computationally intensive and are susceptible to ill-posedness and nonconvexity typical of CWI problems. Machine learning-based computational methods have recently emerged, offering a different perspective to address these challenges. Diverse scientific communities have independently pursued the integration of deep learning in CWI. This review discusses how contemporary scientific machine-learning (ML) techniques, and deep neural networks in particular, have been developed to enhance and integrate with traditional physics-based methods for solving CWI problems.
We present a structured framework that consolidates existing research spanning multiple domains, including computational imaging, wave physics, and data science. This study concludes with important lessons learned from existing ML-based methods and identifies technical hurdles and emerging trends through a systematic analysis of the extensive literature on this topic. 
\end{abstract}


\section{Introduction}

\begin{figure}
\centerline{
\subfloat[]{%
    \includegraphics[width=0.5\textwidth]{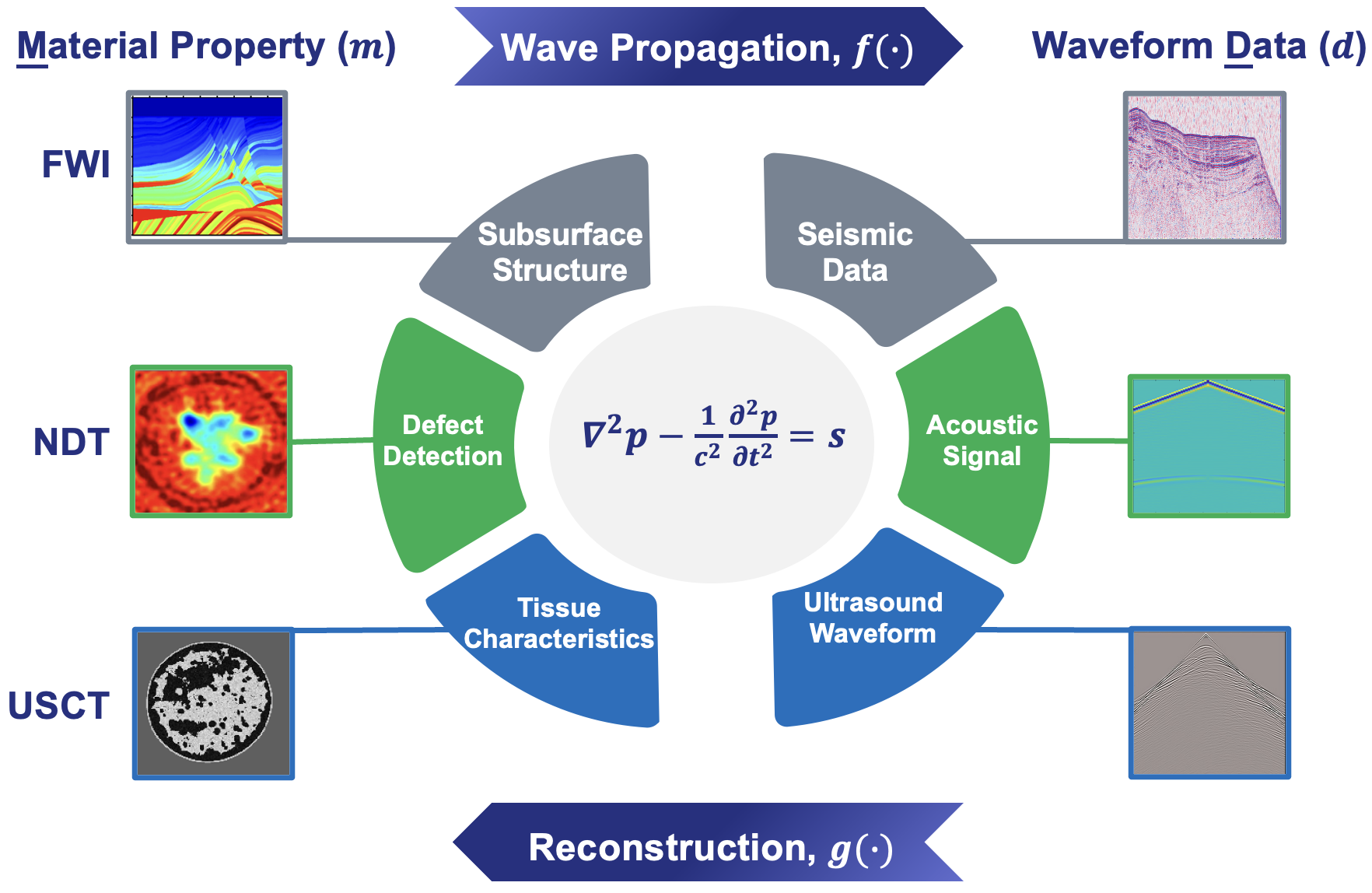}%
    \label{fig:WaveImaging}}}
\centerline{
\subfloat[]{%
    \includegraphics[width=0.5\textwidth]{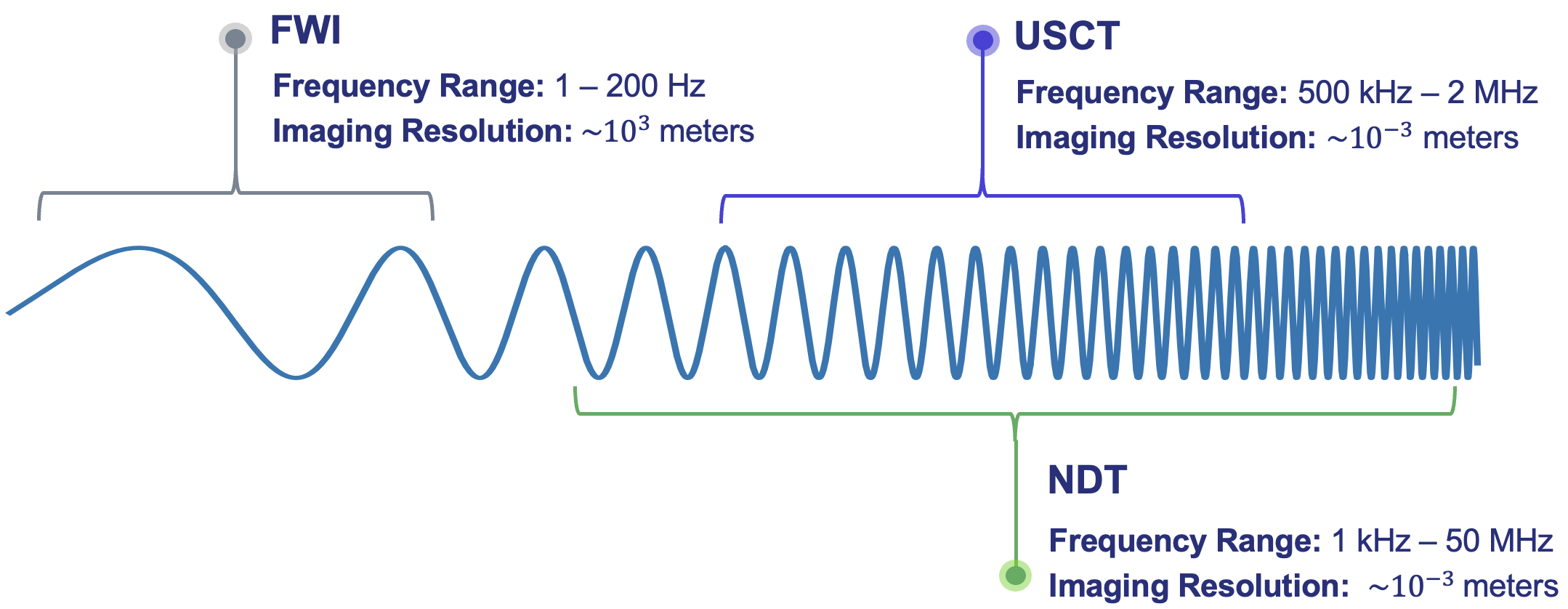}%
    \label{fig:Spectrum}}}
\caption{An illustration of (a) CWI problems covered in this paper and (b) specific CWI applications differentiated by their source spectrum. The common feature is the forward wave equation, and the common goal is the reconstruction of physical properties from wave measurements. We will discuss various approaches for achieving this reconstruction in the context of three application areas:  seismic Full-Waveform Inversion~(FWI), acoustic and industrial ultrasound including Non-Destructive Testing~(NDT), and medical Ultrasound Computed Tomography~(USCT).}
\label{fig:CWI}
\end{figure}
Waves are ubiquitous in our world; prominent examples include the transmission of sound through the atmosphere, the motion of water on the surface of the ocean, and the traveling of seismic vibrations through the earth. Well-established physical models provide detailed descriptions of how these waves are influenced by the materials through which they propagate. Computational wave imaging~(CWI) aims to invert these models, and to extract detailed information about these materials, including properties such as density and bulk modulus, using measurements of wave signals (amplitudes, phases, and travel-times) taken on the outer boundaries of these spaces. 

This review will focus on the three distinct application areas shown in Fig.~\ref{fig:WaveImaging}: seismic Full-Waveform Inversion~(FWI), acoustic and industrial ultrasound imaging (encompassing Non-Destructive Testing, or NDT), and medical Ultrasound Computed Tomography~(USCT). Despite the differences in the frequency of the propagating wave (Fig.~\ref{fig:Spectrum}), and the scale of the imaging volumes, these applications share several common underlying physical processes. Each involves both a forward wave propagation model and an inverse imaging or reconstruction problem, as depicted in Fig.~\ref{fig:WaveImaging}. ``Wave Propagation'' is described by the governing wave equation, and is generally the more straightforward problem; it often requires the use of numerical methods to simulate waves in complex media, as analytical solutions to wave equations are only attainable in idealized conditions.  Conversely, the more challenging ``reconstruction'' task is to estimate the properties of the intervening medium based on measurements (\eg, of acoustic pressure) taken outside the region of interest. We have observed the great potential of machine learning~(ML) to enhance the accuracy and efficiency of solutions to CWI problem in various fields, such as subsurface energy applications~\cite{araya2018deep, Deep-2021-Yu, Physics-2023-Lin}, material characterization~\cite{ Wang2022MSSP, Ryu2023, Quantitative-2023-Rao}, and medical ultrasound imaging~\cite{liu2021ultrasound, stanziola2023learned, Learned-2023-Lozenski}. Such improvements to CWI can be impactful, as these techniques are not only tools for problem-solving but also catalysts for scientific discovery and engineering innovation. Indeed, they can even lead to the discovery of new physical insights~\cite{AutoLinear-2024-Feng}. Furthermore, we have found that integrating physics into ML models contributes to improved model generalization and robustness, and makes more effective use of limited training data~\cite{Jin-2021-Unsupervised, Image-2023-Chen}. Our objective is to provide an overview of data science and ML techniques used for CWI challenges, and to underscore the significant advantages of incorporating the fundamental underlying physics into the design of ML models.

A common set of computational challenges emerges from the shared foundation of wave physics across diverse imaging problems. One prominent challenge is the pervasive ill-posedness encountered in CWI problems, arising from the inherently limited data coverage; \ie, attempting to characterize a 3D volume with measurements confined to its 2D surface. To mitigate ill-posedness, conventional physics-based methods introduce techniques such as regularization, but selecting an appropriate regularization term can be a non-trivial task. Additionally, numerical optimization methods for solving CWI problems have their own difficulties: (1) the accuracy of the solution can be highly contingent on the initial guess; and (2) the computational cost can be prohibitive. Traditional physics-based CWI solvers often require many evaluations of the forward model before achieving convergence, presenting substantial computational burden. Furthermore, cycle-skipping is a well-documented issue in CWI~\cite{Tackling-2019-Yao} that occurs when the timing of the modeled waveform (predicted by the inverted solution) deviates by more than half a cycle from that of the observed waveform.  The goal of this paper is to provide a review of contemporary methods in CWI, with a focus on common challenges and methods. In particular, we explore both the opportunities and difficulties inherent in the application of ML to enhance and expedite the solution of CWI problems.
 
ML has significantly influenced the evolution of CWI techniques. This synergy is particularly evident in the emerging trend of integrating fundamental physics principles with ML algorithms to advance CWI methods. Here we have compiled over 200 papers from three CWI communities, and have observed a clear and consistent upward trend in the volume of publications dedicated to ML within this domain (as shown in Fig.~\ref{fig:counts} in the Appendix). Furthermore, each of these scientific communities has produced high-quality review papers, a selection of which is presented in Table~\ref{tab:summary2} of the Appendix. These reviews can be categorized into three primary groups: physics-based methods, machine-learning approaches, and hybrid methods that integrate data-driven techniques with fundamental physics principles.

\section{Governing Wave Physics}

Wave-based imaging relies on partial differential equations (PDEs) to describe the propagation of mechanical waves through physical media. The choice of wave model varies across application domains and is dictated by the properties of the medium and the nature of the wave. In Earth sciences, the acoustic wave equation is commonly used in seismic exploration~\cite{Virieux-2009-Overview}, while elastic models are essential for simulating earthquake processes~\cite{Tromp-2020-Seismic}. In medical imaging, acoustic wave models are standard for ultrasound computed tomography~\cite{Quantitative-2023-Ruiter}, with elastic or viscoelastic models employed in applications such as elastography and transcranial ultrasound~\cite{Ultrasound-2017-Sigrist}. In nondestructive testing (NDT), both acoustic and elastic wave equations are used, depending on the geometry and material complexity of the structures being interrogated~\cite{Quantitative-2023-Rao}.


\subsection{Acoustic Wave Equation}
\label{sec:acoustic}

The acoustic wave equation models the propagation of compressional (P) waves through fluid-like media, where shear stresses are negligible. It is widely used in seismic exploration~\cite{Virieux-2009-Overview}, ultrasound imaging~\cite{Quantitative-2023-Ruiter}, and NDT~\cite{Quantitative-2023-Rao} scenarios involving air or water. This model also serves as the default choice in many learning-based CWI frameworks due to its mathematical simplicity and computational efficiency.

The constant density form of the acoustic wave equation is:
\begin{equation}
\nabla^2 p(\bm{r}, t) - \frac{1}{V_P^2(\bm{r})} \frac{\partial^2 p(\bm{r}, t)}{\partial t^2} = s(\bm{r}, t),
\label{eq:Forward}
\end{equation}
where $p(\bm{r}, t)$ denotes the pressure field, $V_P(\bm{r})$ is the P-wave velocity, $\bm{r}$ represents spatial coordinates, and $s(\bm{r}, t)$ specifies the spatiotemporal source. This form assumes uniform density, simplifying numerical treatment but neglecting key physical effects in heterogeneous media.

The variable-density acoustic wave equation introduces spatial variability in density $\rho(\bm{r})$~\cite{mulder2004comparison,liu2009new}:
\begin{equation}
\nabla \cdot \left( \frac{1}{\rho(\bm{r})} \nabla p \right) - \frac{1}{\rho(\bm{r}) V_P^2(\bm{r})} \frac{\partial^2 p}{\partial t^2} = s(\bm{r}, t).
\label{eq:Forward_rho}
\end{equation}
This model offers a more accurate description of wave propagation in complex environments such as sedimentary basins or soft tissues. While computationally more demanding, it remains widely used due to its flexibility and compatibility with both inversion and learning-based algorithms.

Overall, the acoustic wave equation plays a foundational role in CWI. It supports synthetic data generation, enables physics-informed learning, and offers a physically grounded framework for designing loss functions and priors in modern deep learning pipelines. Equations~\eqref{eq:Forward} and \eqref{eq:Forward_rho} describe linear acoustic wave propagation, derived under a small-amplitude assumption such that material properties are independent of the pressure field and superposition holds. Unless otherwise noted, the remainder of this paper adopts the constant density formulation without loss of generality.

\subsection{Extensions to Complex Wave Physics}
\label{sec:AdvancedWaveModels}

Although the acoustic wave equation provides a foundational model widely used across many CWI applications, real-world propagation environments often involve more complex physical phenomena. These complexities arise from material anisotropy, energy dissipation, and the presence of both compressional and shear wave modes. To address these effects, various extensions to the acoustic model have been developed and are employed in geophysics, medical imaging, and nondestructive testing~\cite{levander1988fourth,alkhalifah2000acoustic,blanch1995modeling}.

\textbf{Elastic Wave Equation.~} In solid media, wave propagation involves both compressional (P) and shear (S) waves. The elastic wave equation models this behavior by jointly solving for velocity and stress tensors, using Lamé parameters to describe the material’s elastic properties. It is essential in applications such as earthquake seismology and materials characterization~\cite{Tromp-2020-Seismic}.

\textbf{Viscoacoustic Wave Equation.~} To model attenuation in lossy or dissipative media, viscoacoustic formulations introduce memory variables and relaxation mechanisms that simulate the frequency-dependent absorption of wave energy~\cite{blanch1995efficient}. These models are critical in biomedical and seismic contexts, where frequency-dependent attenuation and dispersion lead to amplitude decay and spectral loss, thereby degrading signal-to-noise ratio and indirectly affecting effective image resolution and reconstruction accuracy.

\textbf{Anisotropic Wave Equation.~} Many materials exhibit directionally dependent wave speeds due to internal structure or stress fields. Anisotropic wave equations, including simplified pseudo-acoustic forms such as Vertical Transverse Isotropy (VTI), enable modeling of wave propagation in these directionally varying environments~\cite{alkhalifah2000acoustic,duveneck2008acoustic}. Such models are widely used in layered earth models and anisotropic composites.

\textbf{Other Variants.~} Additional extensions include poroelastic models for fluid-solid interactions and coupled elastic-acoustic systems~\cite{meyers2008mechanical}. These formulations offer greater fidelity but also introduce increased complexity in both numerical simulation and learning-based inversion.

While this review focuses on the acoustic wave equation for clarity and broad applicability, many of the learning methods described in the following sections can be extended to handle these more complex wave physics. Readers are referred to  Appendix~\ref{sec:Appendix-A} for detailed mathematical formulations of these advanced wave equations.

\begin{figure*}[t]
\centering
    \includegraphics[width=\linewidth]{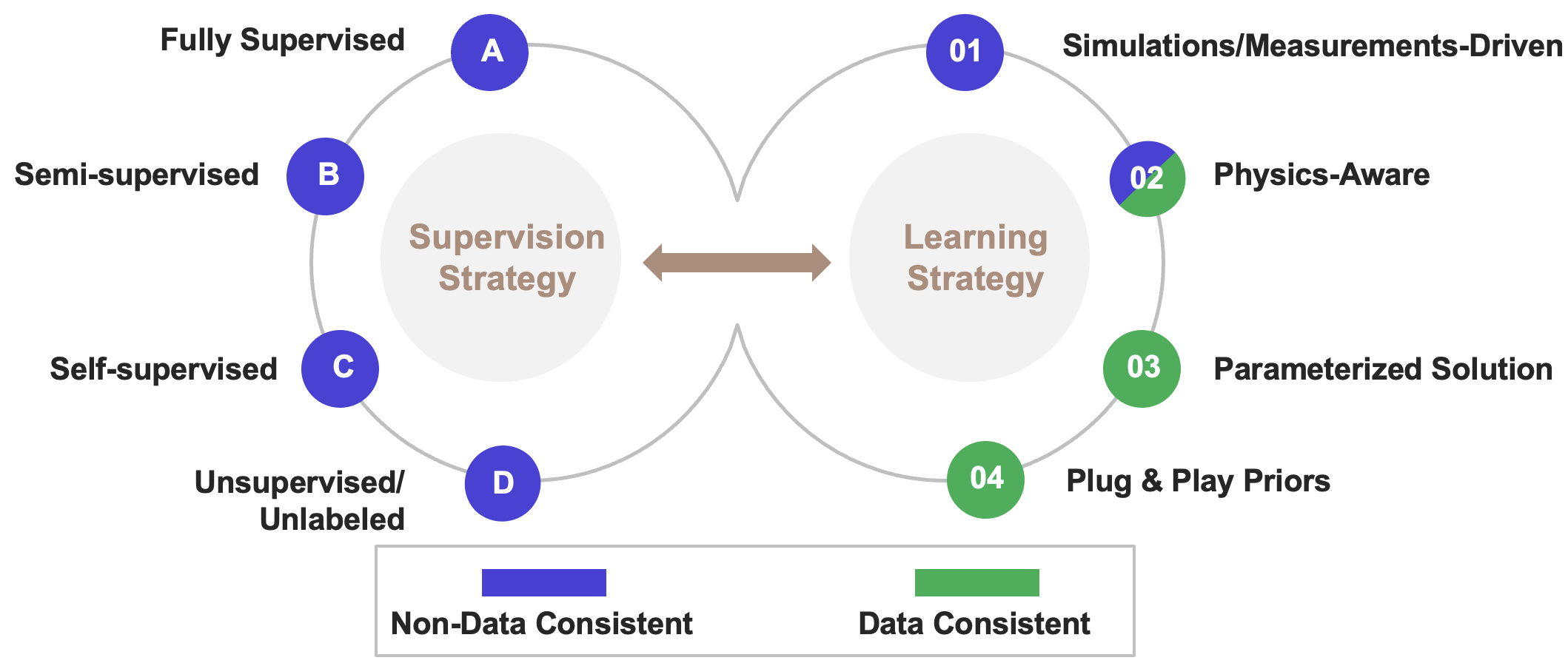}%
\caption{A taxonomy of ML methods for CWI, based on two categorizations. Under supervision strategy, we distinguish fully supervised, semi-supervised, self-supervised, and unsupervised/unlabeled. Under learning strategy, we categorize methods as simulations/measurements-driven, physics-aware, parameterized solutions, and plug-and-play priors. We further differentiate these methods by ``Data Consistent''~(in green) and ``Non-Data Consistent''~(in blue) based on their adherence to data fidelity consistency. In Appendix~\ref{sec:Appendix-B}, we offer visualizations illustrating trends in ML methods for CWI, along with a breakdown categorizing over 200 papers into eight method groups. This section also includes a curated table highlighting selected papers from each category. }
\label{fig:trend_breakdown}
\end{figure*}

\section{Computational Wave Imaging and Physics-based Methods}

\subsection{Forward Modeling}

Physics-based methods are broadly implemented in various CWI applications. Generally speaking, these methods are computationally expensive, and require careful regularization. However, they tend to be more generalizable and robust to measurements contaminated with noise than data-driven methods.
The forward modeling of CWI can be written
\begin{equation}
\bm{d} = f(\bm{m}) + \boldsymbol{\eta} \,,
\label{eq:ForwardLinearM}
\end{equation}
where $\bm{d}$ represents the observed data, $f(\cdot)$ is the forward model, $\bm{m}$ represents the material properties we wish to infer, and $\eta$ represents measurement noise.

Particularly for acoustic wave modeling, which is the focus of this manuscript, the observed data $\bm{d}$ typically consists of time traces of the pressure wavefield $p(\bm{r}, t)$ recorded at discrete receiver locations $\bm{r}$, while $\bm{m}$ corresponds to the spatially varying velocity map $V_P(\bm{r})$. In many practical settings, the forward model also depends on additional physical or algorithmic nuisance parameters---quantities that influence the data but are not the primary inversion targets. These parameters may be set to nominal values, estimated through calibration, or treated as additional unknowns to be jointly or implicitly handled during inversion. Common examples include the source location, source time function, boundary conditions, sensor calibration parameters, or, in variable-density formulations, the density field when it is not explicitly reconstructed. While not the primary focus of CWI, these factors must be carefully handled to ensure accurate reconstruction.

A variety of numerical techniques have been employed for forward modeling. Finite difference methods~\cite{kelly1976synthetic, moczo2007finite}, which discretize the domain into regular grids and approximate derivatives using difference formulas, work well for simple geometries. Staggered-grid finite difference methods~\cite{virieux1984sh, virieux1986p, graves1996simulating} can enhance numerical stability and reduce numerical dispersion. Finite element methods~\cite{kuhlemeyer1973finite, de2009new,komatitsch2010high} are well suited to irregular geometries and complex material properties. Pseudospectral methods~\cite{kosloff1982forward, fornberg1988pseudospectral, tabei2002k, cox2007k, firouzi2012first} have proven effective for problems that require high precision in spatial derivatives. Finally, spectral element methods~\cite{komatitsch2002spectralI, komatitsch2002spectralII, komatitsch2005spectral, kudela2007modelling} combine the advantages of spectral methods and finite elements, offering both high accuracy and flexibility in complex scenarios.


\subsection{Physics-based Wave Imaging}

Computatioanl wave imaging aims to estimate the material property $\bm{m}$ from the observed data~$\bm{d}$; that is, it aims to find an inversion operator $g$ so that $\widehat{\bm{m}}  = g(\bm{d})$ approximates the actual property.  A minimal requirement for this approximate model is that the forward propagation based on this model should approximate the observed data; that is, $f(\widehat{\bm{m}})\approx\bm{d}$. This implies that $f(g(\bm{d}))\approx \bm{d}$ and for this reason, $g$ is informally referred to as the \emph{inverse} of $f$. But directly inverting $f$ is generally infeasible due to the ill-posed nature of CWI. To address these challenges, CWI is often formulated as the solution to an optimization problem. The unknown material properties are then estimated as the minimizers $\widehat{\bm{m}}$ of a penalized nonlinear least squares problem and can be written
\begin{equation}
\widehat{\bm{m}} = \underset{\bm{m}}{\operatorname{argmin}}~\left[
 \left \| \bm{d} - f(\bm{m})\right \| _2 ^2 + \gamma \, r(\bm{m})\right],
\label{eq:MisFit}
\end{equation}
where the first term enforces data fidelity and the second term, weighted by $\gamma$, incorporates prior information with a regularization function $r(\bm{m})$, which encodes assumptions about the underlying medium~\cite{Lin-2013-Ultrasound, Lin-2014-Ultrasound, Acoustic-2015-Lin, Quantifying-2015-Lin}. Classical choices include Tikhonov regularization, $r(\bm{m}) = \| H \bm{m} \|_2^2$, where $H$ is typically a derivative (high-pass) operator so that the penalty suppresses high-frequency components, thereby promoting smoothness, and total variation (TV) regularization, $r(\bm{m}) = \| \nabla \bm{m} \|_1$, which encourages piecewise smooth structures and preserves edges\cite{Osher-1992-Nonlinear}.

Beyond these, many advanced regularization strategies have been developed to better adapt to specific medium structures and improve reconstruction quality. For instance, modified TV formulations have been used to enhance the delineation of high-contrast interfaces~\cite{Acoustic-2015-Lin}; adaptive schemes such as the proximal Newton–ADMM framework with pretrained denoisers allow dynamic balancing between smooth and blocky regions~\cite{aghamiry2020hybrid}; and constraint-based approaches like salt flooding techniques have proven effective in complex subsurface environments~\cite{kalita2019Regularized}. Recently, hybrid strategies that decompose the model into smooth and blocky components, such as the adaptive Tikhonov–TV regularization method~\cite{Robust-2025-Aghazade}, demonstrated strong robustness across varying realistic settings.

\section{Deep Learning for Computational Wave Imaging Problems}

In this section, we discuss ML-based methods for CWI, categorizing them by supervision and learning strategies as illustrated in Fig.~\ref{fig:trend_breakdown}. Supervision strategy pertains to the use of labels and data, while learning strategy involves integrating physics into ML model design. Techniques are typically classified by supervision strategy in computer vision and ML, and by learning strategy in computational imaging and physics. We present these categorizations side-by-side for a comprehensive understanding across different communities. Additionally, we classify methods based on data fidelity consistency: data-consistent methods ensure model outputs closely align with actual measurements, enhancing trustworthiness and generalizability, while non-data consistent methods may produce quantitatively incorrect yet realistic estimates.

\subsection{Machine Learning Methods by Supervision Strategy}

Here, we categorize ML methods for solving CWI problems by supervision strategy as shown in Fig.~\ref{fig:trend_breakdown}. In what follows, we distinguish between the learning paradigm, which specifies how supervision is obtained during training, and the learning task, which specifies the mapping being learned. 
The learning paradigms discussed below apply to the supervision strategy itself and are independent of the specific task being considered.

\begin{itemize}
    \item \textbf{Fully Supervised.}~Both samples and associated labels are available, and all labels are used to supervise the training of the models. The labels may represent medium parameters, waveform data, or other task-specific targets, depending on the learning objective.
    
    \item \textbf{Semi-Supervised.}~Some labeled samples are available, while additional unlabeled samples are also used during training. The labeled samples correspond to task-specific targets, while unlabeled data are incorporated during training to improve generalization. Semi-supervised methods often leverage pseudo-labeling, consistency regularization, domain adaptation, or generative modeling to exploit unlabeled data when labeled data are scarce.

    \item \textbf{Self-Supervised.}~Explicit labels are not provided. Instead, supervision is obtained through internally constructed learning signals derived from the data itself, such as consistency objectives, contrastive learning, or latent-space similarity constraints. These auxiliary objectives provide guidance during training and are distinct from directly solving the original forward or inverse problem.
    
    \item \textbf{Unsupervised/Unlabeled.}~Training is performed without labeled samples or auxiliary proxy objectives. Learning is driven either by directly enforcing physical data consistency using forward models. Medium parameters are inferred solely by minimizing the mismatch between simulated and observed measurements under the governing physical equations.

\end{itemize}

\begin{figure*}[t]
\centerline{
    \includegraphics[width=0.8\textwidth]{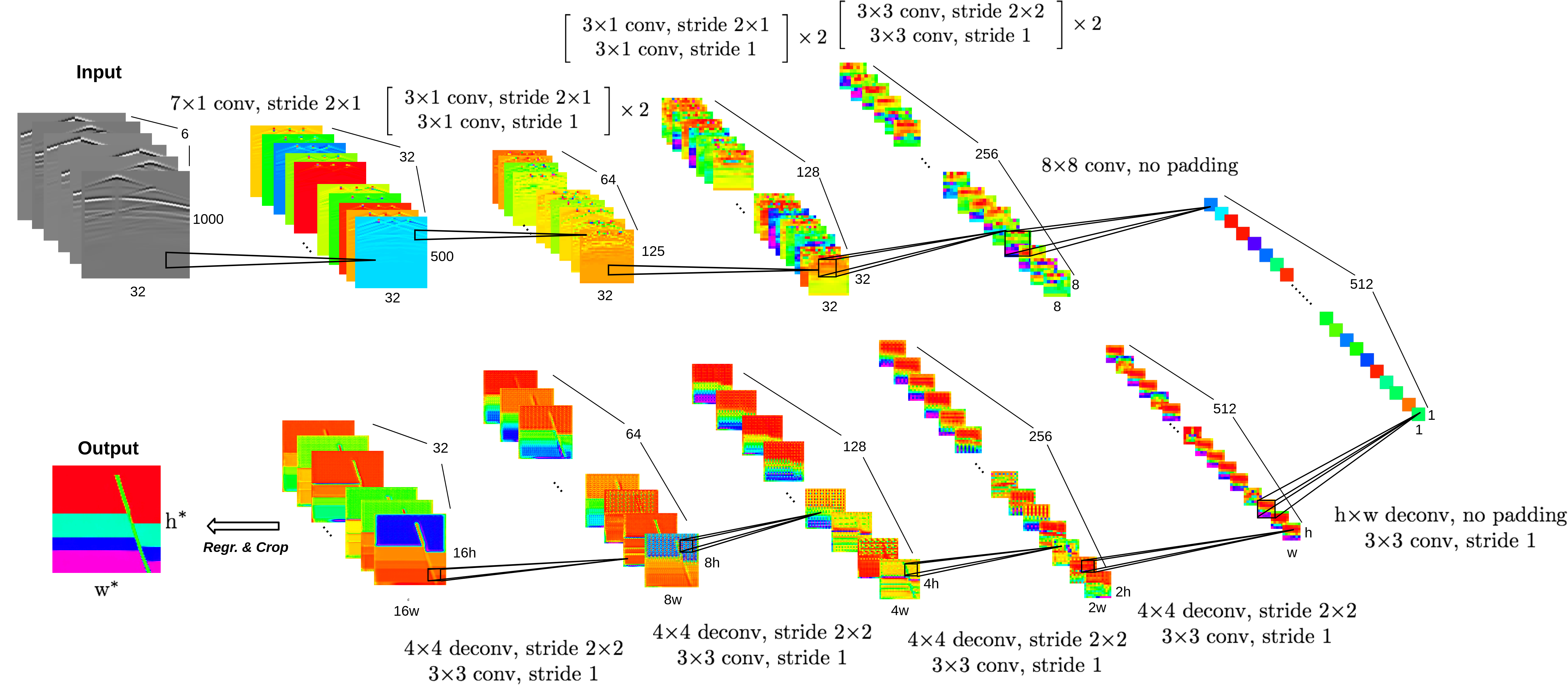}%
    }
\caption{Network architecture of InversionNet, a fully-supervised method~\cite{wu-2019-inversionnet}.}
\label{fig:inversionnet}
\end{figure*}

\subsubsection{Fully Supervised Methods}

Fully supervised methods are among the most straightforward and widely used approaches in solving CWI problems~\cite{araya2018deep, wu-2019-inversionnet, yang2019deep, Quantitative-2023-Rao}. The basic idea is to obtain a large training set $\bm{\Phi}$ of pairs $(\bm{d}_i,\bm{m}_i)$ and to use that data to train a deep neural network that instantiates the inversion operator $g_\theta(\cdot)$. Here, we employ the ``InversionNet'' method, developed by~\cite{wu-2019-inversionnet}, as an example to illustrate how to solve CWI problems using a fully supervised learning strategy. Mathematically, InversionNet can be expressed as  
\begin{align}
\label{eq:encoder-decoder}
	    \widehat{\bm{m}} & = g_{\boldsymbol  \theta^{*}}(\bm{d})\; \\
\nonumber
        \mathrm{where}\; \boldsymbol \theta^{*} & = \argmin_{\boldsymbol \theta} \sum_{(\bm{d}_i, \bm{m}_i) \in \bm{\Phi}} \mathcal{L}(g_{\boldsymbol  \theta}(\bm{d_i}),\, \bm{m}_i) \,,
\end{align}
where $\bm{m}$ is the predicted velocity map, $\boldsymbol \theta$ represents the trainable weights in the inversion network, $g_{\boldsymbol \theta}(\cdot)$ corresponds to the network itself,  $\bm{\Phi}$ is the training dataset with paired samples of $(\bm{d}_i, \bm{m}_i)$, and the loss function $\mathcal{L}(\cdot, \,\cdot)$ is commonly defined in terms of the $\ell_1$   or $\ell_2$ norm, comparing the difference of $\bm{m}_i$ with~$\widehat{\bm{m}}$. As shown in Fig.~\ref{fig:inversionnet}, InversionNet utilizes an encoder-decoder structure, processing both input waveform data and output velocity maps as images. The labeled samples (\ie, $\bm{d}, \bm{m}$ pairs) are typically synthetically generated to reflect realistic medium profiles, often numbering in the tens of thousands. Various methods, such as the geostatistical approaches detailed by~\cite{Azevedo-2022-Model}, are used to create geologically significant $\bm{m}_i$-images for seismic inversion. Each waveform $\bm{d}$ is computed using the forward operator $f$, as outlined in Eq.~(\ref{eq:ForwardLinearM}). While training requires a large number of labeled samples, a fully trained network can quickly evaluate the inverse operator $g_{\bm{\theta}^*}(\cdot)$, an advantage in monitoring scenarios requiring frequent assessments. However, inference on out-of-distribution data remains challenging, and the network tends to generate solutions that closely mirror the training data, thus limiting generalization.

In addition to InversionNet, several other works build on the concept of fully supervised learning strategies, employing various backbone networks. \cite{araya2018deep} utilizes fully connected networks, while \cite{Quantitative-2023-Rao} develops an encoder-decoder network structure similar to InversionNet to enhance acoustic imaging. \cite{yang2019deep} implements a specific type of encoder-decoder network known as U-net~\cite{Ronneberger-2015-UNet}, which includes skip connections linking the encoder and decoder paths, to facilitate learning. The choice of backbone network architecture can significantly impact the network's properties. Generally, convolution-based networks are more parameter-efficient due to weight sharing, making them particularly suitable for processing high-dimensional waveform data and velocity maps while reducing the size of the required training dataset. In contrast, fully connected networks, due to their high parameter count, may become impractical for large-scale applications.

\subsubsection{Semi-supervised Methods}
\label{Sec:Semi-supervised}

\begin{figure}[t]
\centerline{
    \includegraphics[width=0.5\textwidth]{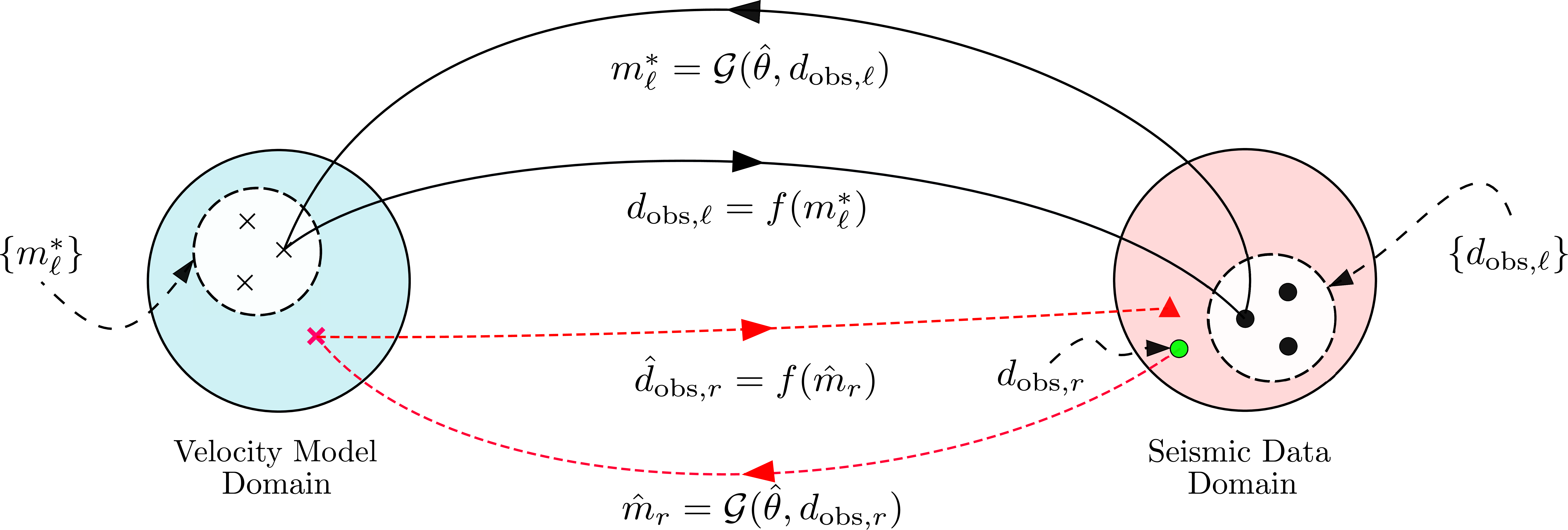}%
    }
\caption{Semi-supervised learning framework for computational wave imaging~\cite{Renan-2022-Physics}, combining labeled data, unlabeled observations, and physics-based forward modeling to generate and refine pseudo-labels.}
\label{fig:Semisupervised}
\end{figure}

Fully supervised learning provides a direct means to apply machine learning to CWI, but its utility is limited by the scarcity of labeled data. Acquiring ground-truth labels, such as velocity models, can be costly and time-consuming, whereas collecting unlabeled waveform data is typically much more feasible. For example, in seismic FWI, distributed acoustic sensing (DAS)~\cite{Li-2021-Distributed} allows efficient acquisition of large volumes of unlabeled seismic data. To address the imbalance between labeled and unlabeled data availability, various semi-supervised learning approaches have been developed~\cite{Renan-2022-Physics, Cai-2022-Semi, Feng-2022-Exploring}.

A typical strategy in semi-supervised learning is to generate ``pseudo-labels'' from unlabeled data or auxiliary measurements. Specifically, \cite{Renan-2022-Physics} illustrates how semi-supervised learning can be adapted to solving CWI as shown in Fig.~\ref{fig:Semisupervised}. The high-level idea is to adapt the training set to the particular waveform $\bm{d}$ that was measured, by incrementally generating more data samples that are close to the measured data.  The first step is to learn a reconstruction operator $g_0$ from the waveform-data manifold to velocity-map manifold using the original pairwise data. From the measured data, use $g_0(\bm{d}+\eta)$ to create a new batch of models $\bm{m}$, and then apply $f$ to those models to create a new batch of labeled samples.  The training set is augmented with the new labeled samples, and a new regression $g_1$ is fit.  With each iteration, more labeled samples are in the regime of the measured sample, and the performance of the resulting $g$ is improved for the waveform $\bm{d}$ of interest. Please refer to Appendix~\ref{sec:Appendix-D} for more implementation details. 

Similarly,\cite{Cai-2022-Semi} also generates pseudo-labels from waveform data but incorporates a CycleGAN architecture to improve the fidelity of the inferred labels.  \cite{Feng-2022-Exploring} utilizes available well-log data to construct pseudo-labels for velocity profiles. Together, these methods highlight the flexibility of semi-supervised methods in balancing data availability and model generalization in CWI.



\subsubsection{Self-Supervised Methods}

\begin{figure}[t]
\centerline{
    \includegraphics[width=0.5\textwidth]{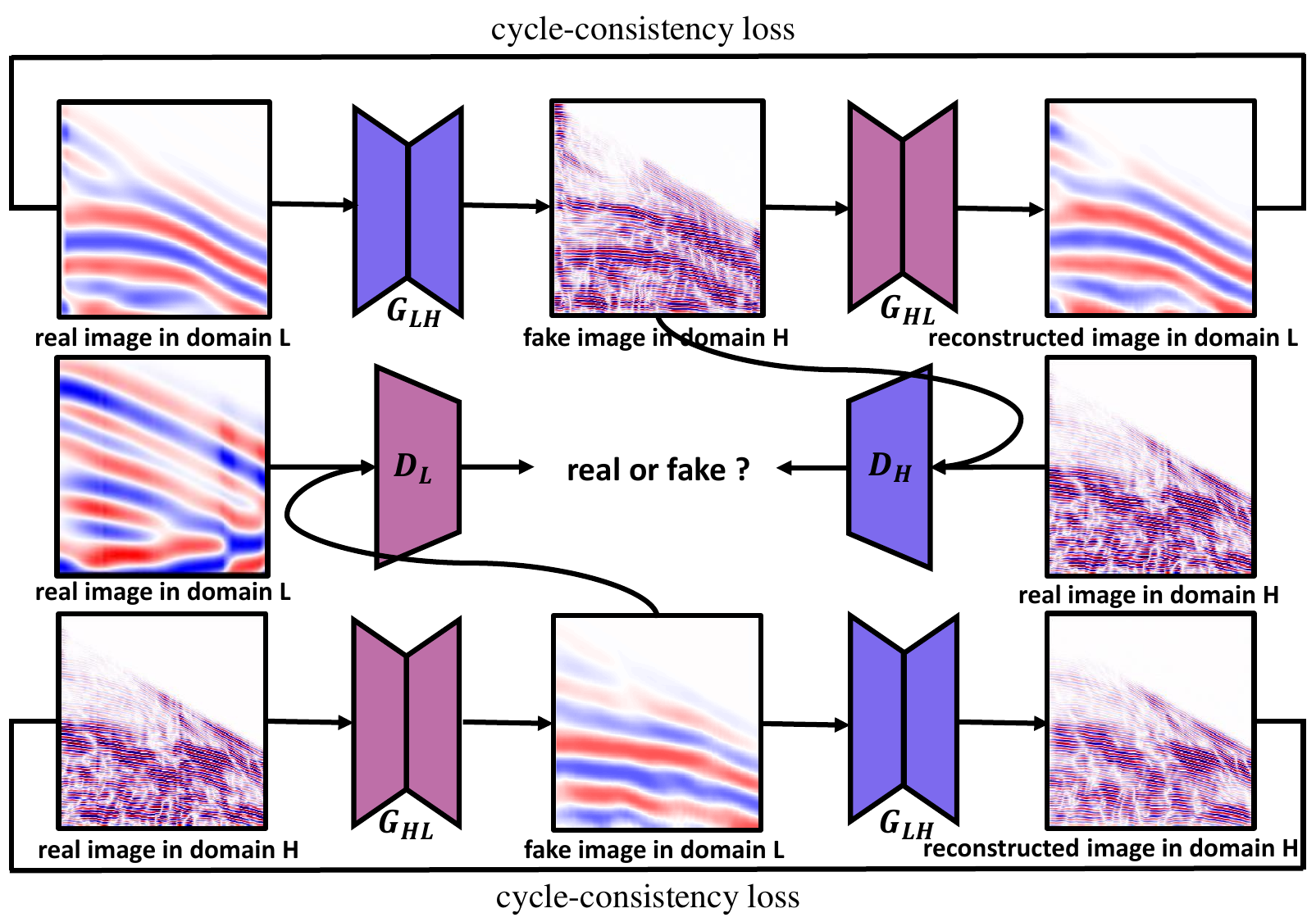}%
    }
\caption{Schematic illustration of self-supervised learning strategies in CWI. In the CycleGAN-based low-frequency extrapolation method~\cite{Learning-2023-Sun}, the network learns to translate between unpaired high-frequency field data and synthetic low-frequency data using adversarial and cycle-consistency losses. Image courtesy of the authors of \cite{Learning-2023-Sun}.}
\label{fig:Selfsupervised}
\end{figure}

Self-supervised learning involves the model generating its own supervision signals from unlabeled data. The model creates tasks, such as predicting parts of the data from other parts, where labels are implicitly generated through the data itself. Its key idea is to allow models to learn from measurements themselves without the need for a vast amount of labels. Self-supervised learning remains an emerging area within CWI, with growing interest in its application to low-frequency extrapolation, which is an essential step in overcoming the lack of low-frequency content in seismic data that limits FWI performance~\cite{Self-2024-Cheng, Learning-2023-Sun, Self-2020-Wang}. Particularly, as shown in Fig.~\ref{fig:Selfsupervised}, \cite{Learning-2023-Sun} proposes to train a CycleGAN model using unpaired data from two domains: band-limited field shot gathers (typically 4--10\,Hz), denoted \( h \in \mathcal{H} \), and synthetic low-frequency data (0--4\,Hz), denoted \( l \in \mathcal{L} \), derived from known subsurface models. Rather than relying on paired training data, the model learns mappings between domain distributions using adversarial learning and cycle-consistency constraints. The architecture comprises two generators, \( G_{HL}: \mathcal{H} \rightarrow \mathcal{L} \) and \( G_{LH}: \mathcal{L} \rightarrow \mathcal{H} \), along with two discriminators, \( D_L \) and \( D_H \), trained jointly with adversarial, identity, and cycle-consistency losses. Waveform structure is preserved by enforcing the cycle-consistency conditions:
\[
G_{LH}(G_{HL}(h)) \approx h \quad \text{and} \quad G_{HL}(G_{LH}(l)) \approx l,
\]
allowing the network to produce plausible low-frequency extrapolations even in the absence of ground-truth labels. Once trained, \( G_{HL} \) is applied to real field data to generate low-frequency components, which are then used to enhance inversion quality in a subsequent two-stage full-waveform inversion~(FWI) pipeline. The method is successfully validated on both synthetic and real marine field datasets.

One early effort~\cite{Self-2020-Wang} introduces an self-supervised learning method that trains a neural network using paired seismic datasets with shared high-frequency content but varying low-frequency components. The model learns to reconstruct missing low-frequency information, although its performance is inherently constrained by the frequency range present in the original data. Furthermore, the required downsampling can degrade high-frequency resolution, limiting time precision. To address these challenges, \cite{Self-2024-Cheng} proposes a more effective two-stage self-supervised learning framework for low-frequency extrapolation. The warm-up stage initializes training using high-pass filtered inputs and original data as pseudo-labels. In the subsequent iterative data refinement stage, the model refines its predictions by using outputs from the previous epoch as pseudo-labels, with their high-pass filtered versions serving as new inputs. This iterative refinement progressively closes the gap between predictions and the ideal low-frequency content, enhancing extrapolation accuracy and robustness.

\subsubsection{Unsupervised/Unlabeled Methods}



\begin{figure}
\centerline{
\subfloat[]{%
    \includegraphics[width=\linewidth]{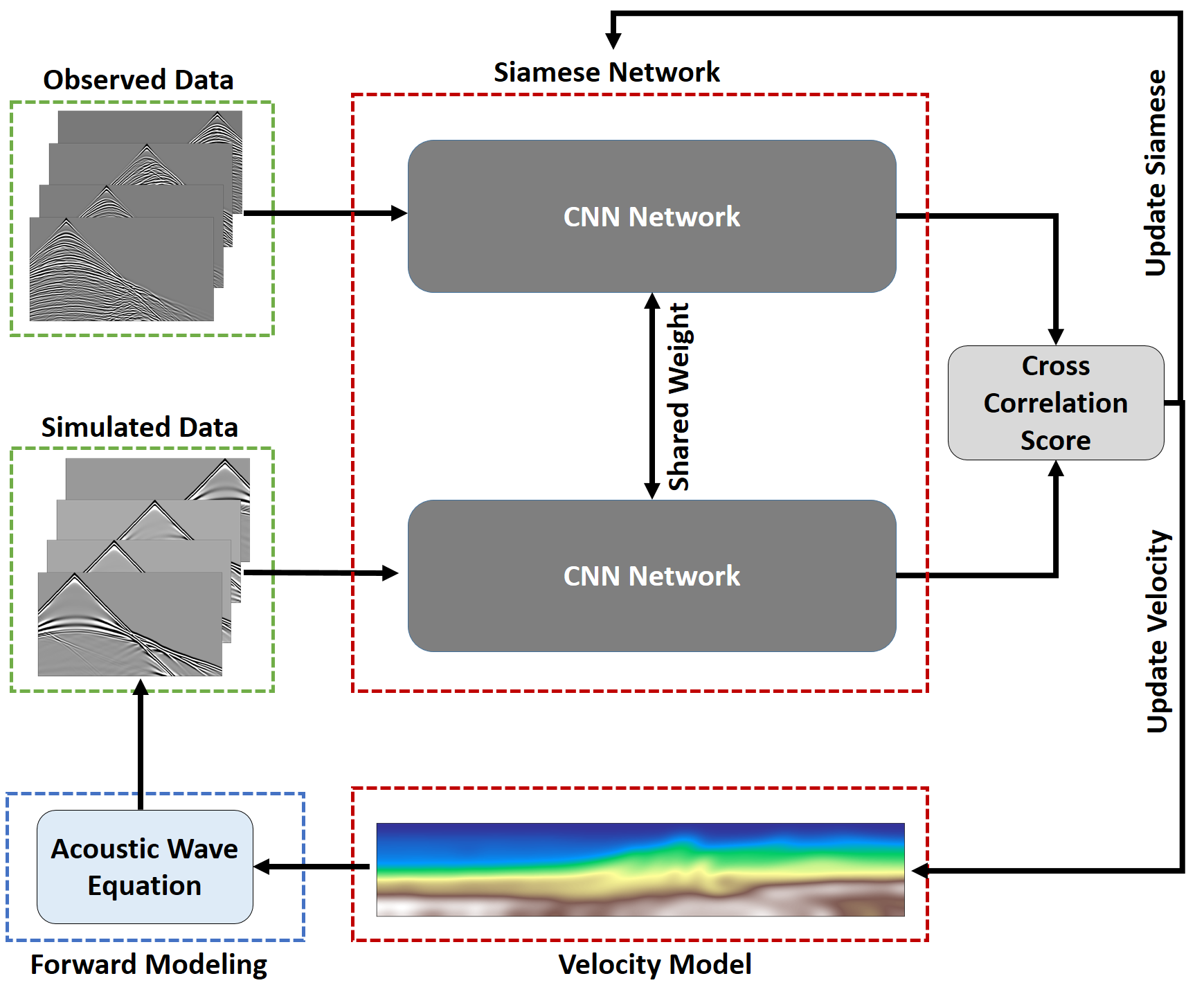}%
    \label{fig:SiameseFWI}}}
\vspace{1em} 
\centerline{
\subfloat[]{%
    \includegraphics[width=\linewidth]{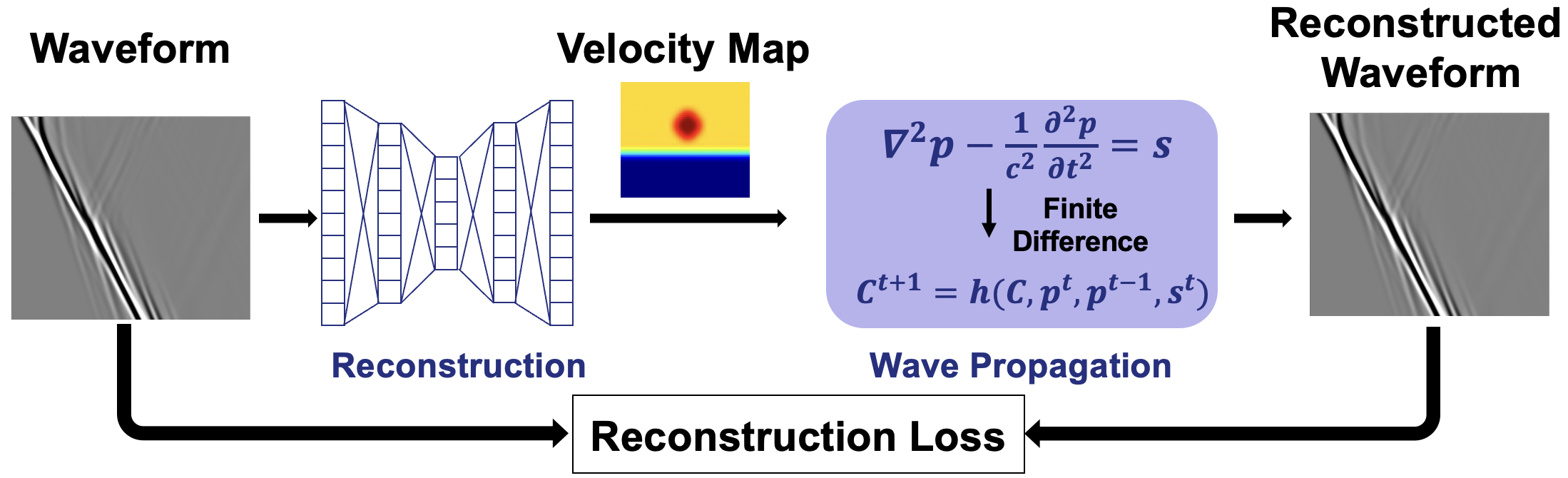}%
    \label{fig:upfwi}}} 
    \caption{Schematic overview of two representative unsupervised/unlabeled methods. (a) The SiameseFWI~\cite{SiameseFWI-2024-Saad}, a self-supervised learning framework for full waveform inversion. The model uses a Siamese neural network to compare simulated and observed seismic data in a shared latent space, enhancing sensitivity to key structural features. (b) The UPFWI architecture, a physics-informed unlabeled method that maps waveform data to velocity models by minimizing the mismatch between observed and simulated waveforms via a differentiable wave equation~\cite{Jin-2021-Unsupervised}. Image (a) courtesy of the authors of \cite{SiameseFWI-2024-Saad}.}
\label{fig:Unsupervised}
\end{figure}

This category encompasses methods that operate without paired supervision; \ie, where only waveform samples are available and no velocity labels are required. Methods in this category focus on mapping waveform data to velocity models by directly minimizing the mismatch between observed and simulated waveforms, using the full wave equation as a physical constraint. 

A representative example is the SiameseFWI~\cite{SiameseFWI-2024-Saad}, which is a novel self-supervised framework for FWI that uses a Siamese neural network to enhance the comparison between simulated and observed seismic data, as shown in Fig.~\ref{fig:SiameseFWI}. Unlike traditional pixel-wise loss functions such as the $\ell_2$ norm, SiameseFWI transforms both data types into a shared latent space via two identical convolutional neural networks with shared weights. This design enables the network to focus on meaningful features (\eg, P-wave arrivals), and it minimizes the Euclidean distance between the learned representations to guide velocity model updates. The method achieves robust inversion performance on synthetic benchmarks and field data from Western Australia, demonstrating greater accuracy and resilience to noise, poor initial models, and unknown source wavelets.

Another example is the \textit{Unsupervised Physics-informed Full-Waveform Inversion~(UPFWI)} framework (shown in Fig.~\ref{fig:upfwi}) developed in \cite{Jin-2021-Unsupervised}. UPFWI couples a neural encoder-decoder architecture with a differentiable wave-equation simulator, enabling end-to-end training with a loss defined in the data space:
\begin{align}
	\label{eq:UPFWI}
	    \widehat{\bm{m}} & = g_{\boldsymbol  \theta^{*}}(\bm{d})\;,\\  \mathrm{where}\; \boldsymbol  \theta^{*} & = \argmin_{\boldsymbol  \theta} \sum_{\bm{d}_i \in \bm{\Phi}_u} \mathcal{L}(f(g_{\boldsymbol  \theta}(\bm{d}_i)),\,\bm{d}_i) \,. \nonumber
\end{align}
Here, $\bm{\Phi}_u$ denotes a dataset of unlabeled waveform measurements, $g{\boldsymbol_\theta}$ is the network mapping waveforms to velocity models, and $f$ represents the forward modeling operator. The simulated waveforms are compared to the observed data to guide the network's learning. Methods in this category can be further divided into two subtypes:
\begin{itemize}
    \item \textbf{Per-sample methods}, such as those in \cite{Implicit-2023-Sun} and \cite{he2021reparameterized}, reparameterize the model and optimize its parameters individually for each input, mimicking traditional inversion routines.
    \item \textbf{Per-domain methods}, like UPFWI~\cite{Jin-2021-Unsupervised} and \cite{Seismic-2025-Jia}, train a single model across a dataset and apply it to similar examples, enabling faster inference but possibly inheriting dataset-specific biases.
\end{itemize}

Together, these methods reflect a growing effort to decouple learning-based inversion from the need for explicit supervision, while still leveraging physical constraints and inductive priors to achieve accurate reconstructions.


\subsubsection{Comparative Overview of Supervising Strategies for CWI}

\textbf{Fully Supervised Learning.~} Fully supervised learning provides the most direct application of machine learning to CWI by training models on datasets composed of paired waveform measurements and corresponding physical property maps. This strategy has been explored across various domains---including geophysics, medical imaging, and materials characterization---to approximate complex mappings from time-domain signals to spatial distributions of physical parameters such as velocity, impedance, or elastic moduli. Its appeal lies in its conceptual simplicity and fast inference: once trained, these models can generate real-time predictions without iterative solvers or numerical simulations.

However, the theoretical appeal of this approach belies practical challenges. The need to implicitly learn the underlying physics from data alone makes these models highly data-intensive. They often require orders of magnitude more labeled examples than physics-informed or hybrid approaches to achieve comparable generalization. Label acquisition is especially difficult in CWI contexts, where high-fidelity ground-truth models are either unavailable (as in seismic exploration), invasive to obtain (as in biomedical applications), or destructive (as in nondestructive testing of materials). Moreover, supervised models are vulnerable to domain shift: models trained on synthetic or idealized data may not transfer well to field conditions unless the training data closely mimics real-world complexity.

Despite these limitations, there are notable cases where fully supervised learning has achieved success, typically under carefully designed conditions. For example, \cite{Quantitative-2023-Rao} demonstrates accurate ultrasonic defect reconstruction in multilayered materials by tailoring the synthetic training data to match the sensor setup and propagation environment used in experiments. Similarly, \cite{One-2025-Jiang} trains a Transformer-based network on over 20~million examples from a curated dataset of velocity models, enabling their model to generalize across diverse geophysical structures. These successes underscore that fully supervised approaches can work when domain fidelity and data scale are carefully managed---but they also highlight the inherent difficulty of deploying such methods directly in practical scenarios.

Ultimately, while fully supervised learning serves as a foundational paradigm for data-driven CWI, its dependency on exhaustive labeled datasets and lack of embedded physics limits its standalone applicability. These limitations have led to increased interest in semi-supervised, self-supervised, and physics-informed frameworks that can leverage unlabeled data or incorporate domain knowledge to improve generalization and interpretability.

\textbf{Semi-Supervised Learning.}~Semi-supervised learning represents a pragmatic approach to bridging the gap between the data efficiency of supervised learning and the flexibility of unlabeled methods. It is particularly suited for CWI applications, where waveform measurements are abundant but reliable ground-truth labels, such as velocity models, are difficult to obtain. The key strength of semi-supervised learning lies in its ability to exploit the information embedded in unlabeled data to guide or regularize model training, thereby enhancing generalization with limited supervision. Unlike fully supervised methods, semi-supervised frameworks can be designed to integrate physics-based constraints or exploit structural relationships within the data to generate consistency constraints or domain-adaptive feature mappings, rather than relying exclusively on ground-truth labels. In CWI, this enables networks to extract meaningful representations even in the absence of dense annotations. When properly configured, semi-supervised methods can achieve performance close to their fully supervised counterparts, while being more robust to overfitting and more scalable to large, real-world datasets. By accommodating both labeled and unlabeled sources, semi-supervised provides a flexible and data-efficient strategy for learning in complex wave-based imaging tasks.

\textbf{Self-Supervised Learning.~}Self-supervised learning is an emerging paradigm that enables models to learn meaningful representations directly from unlabeled data by formulating surrogate tasks in which the labels are derived from the data itself. In the context of CWI, self-supervised learning offers a powerful means to exploit the inherent structure and redundancy in waveform measurements, without relying on explicit supervision. Recent advances have demonstrated its potential in tasks such as low-frequency extrapolation and feature embedding. For example, self-supervised strategies have been used to learn missing low-frequency waveform components from band-limited observations by enforcing cycle consistency or predictive reconstruction objectives, thereby alleviating the lack of low-frequency data commonly encountered in full-waveform inversion~\cite{Self-2024-Cheng, Learning-2023-Sun, Self-2020-Wang}. In addition, self-supervised learning has been applied to map simulated and observed waveforms into a shared latent space and enforce representation consistency, enabling robust misfit evaluation without the need for pixel-wise loss functions~\cite{SiameseFWI-2024-Saad}. The core advantage of self-supervised methods lies in their ability to harness large-scale unlabeled datasets for pretraining, which can subsequently enhance performance on downstream inversion tasks, particularly when labeled data is scarce. However, the success of self-supervised learning depends heavily on the design of pretext tasks and their consistency with the physics of wave propagation, making it an active and promising area for further exploration in CWI.


\textbf{Unsupervised/Unlabeled Methods.} Unsupervised/unlabeled learning strategies provide a label-efficient alternative for tackling CWI problems by either extracting latent patterns directly from data or leveraging physics-based constraints to supervise the learning process. 
Methods in this category embed physical models, such as the wave equation, into the learning objective, allowing end-to-end inversion from measurements to physical parameters without explicit supervision. These techniques are especially advantageous in applications where unlabeled waveform data are abundant, but acquiring ground-truth models remains difficult. Despite the benefit of bypassing labeled data, these methods often require careful regularization, reliable forward modeling, and principled training strategies to ensure convergence and physical plausibility. Together, unsupervised/unlabeled methods offer a flexible and scalable paradigm for model training under realistic data constraints.

   

\subsection{Machine Learning Methods by Learning Strategy}

Computational wave imaging, unlike typical ML problems such as natural language processing, is governed by underlying physics. We are thus presented with an opportunity to incorporate critical physics and prior knowledge to potentially improve the performance of ML models in solving CWI problems~\cite{Physics-2023-Lin}. As illustrated in Fig.~\ref{fig:trend_breakdown}, different approaches have been developed to incorporate physics into ML; we distinguish the following four:
\begin{itemize}
    \item \textbf{Simulations/Measurements-Driven.~}The method is built on a data-driven loss as in Eq.~(\ref{eq:encoder-decoder}), and physics knowledge is implicitly represented via simulations and/or measurements.
    
    \item \textbf{Physics-Aware.~}The method is built on a data-driven loss, but physics knowledge is introduced through explicit domain-aware regularization. 
    
    \item \textbf{Parameterized Solution.~}The method is built on a physics-based loss as in (Eq.~\ref{eq:MisFit}) and ML is used to learn a parameterization of the physical property. 
    
    \item \textbf{Plug and Play Priors.~}The method typically makes use of a physics-based loss function, together with a generic denoising operator (possibly ML-based) that may not be designed/trained for the specific problem of interest.
\end{itemize}

\subsubsection{Simulations/Measurements-Driven Methods}

\begin{figure}[t]
\centerline{
    \includegraphics[width=0.5\textwidth]{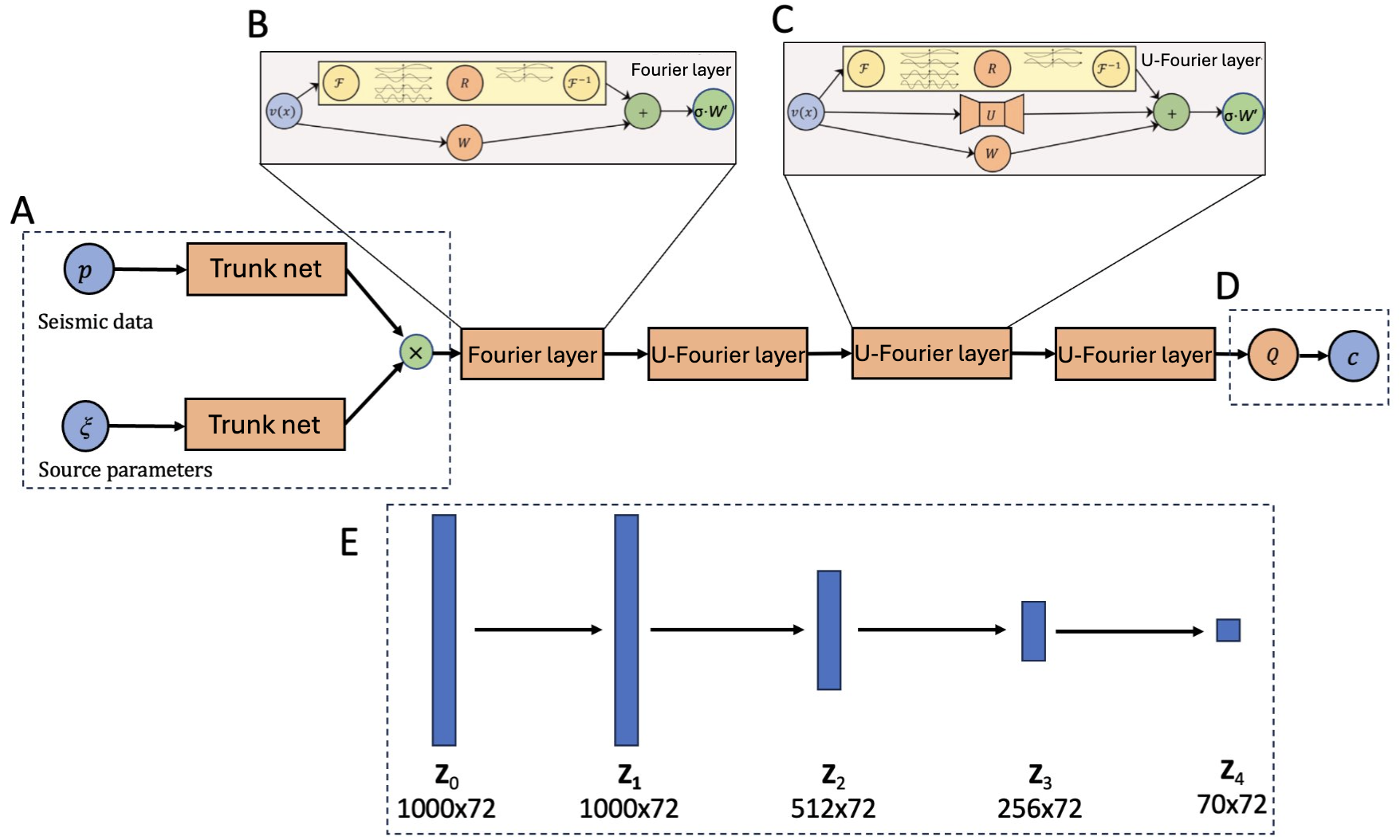}%
    }
\caption{Schematic illustration of Fourier-DeepONet, a simulations/measurements-driven method~\cite{Zhu-2023-DeepONet}. Image courtesy of the authors of \cite{Zhu-2023-DeepONet}.}
\label{fig:F-DeepONet}
\end{figure}

Full-physics simulations and measurements from experiments represent a major source of data for many ML methods for solving physics problems including CWI. Here, simulation data refers to synthetic waveforms  generated by virtual imaging. This is achieved by applying the forward model in Eq.~(\ref{eq:Forward}) to various realizations of the medium properties spanning the physical space of interest. The measurements are these data being acquired through lab or controlled experimental setup. Particularly, given the significant recent increase in available computing power, a very large number of pairwise realizations and waveforms becomes available, enabling fully supervised models as provided in Eq.~(\ref{eq:encoder-decoder}). These simulation/measurement-driven methods learn the underlying physics indirectly through data, by observing how changes in physical parameters affect the resulting wavefields. In this way, they do not encode physics through explicit equations, but rather through exposure to physics-rich data generated via forward modeling.

Here, we use the model~(called ``Fourier-DeepONet'') developed by \cite{Zhu-2023-DeepONet} as an example to illustrate how simulations can be utilized to train ML models for solving CWI problems as shown in Fig.~\ref{fig:F-DeepONet}. Fourier-DeepONet is a representative example of this class because it explicitly leverages frequency-domain information embedded in simulation data to drive learning. It was developed to enhance the generalization ability of waveform sources including both frequency range and source locations. It builds on two major components: DeepONet (deep operator networks)~\cite{Lu-2021-DeepONet} and U-FNO (a block combining Fourier neural operator and U-net)~\cite{Wen-2022-UFNO}. Particularly, source parameters (\ie, frequencies and source locations) are employed as the inputs of the trunk net to facilitate the data-driven model with varying source frequencies and locations. The branch net and trunk net encode the waveform data and source parameters, respectively. They are linear transformations lifting inputs to high dimensional space. A few U-FNO blocks were then utilized as decoders to produce high-resolution imaging output. This use of frequency-domain learning makes Fourier-DeepONet particularly well-suited to handle variability in waveform sources, thus exemplifying how simulation-driven approaches can exploit structured physics knowledge implicitly embedded in data.

Fully supervised learning strategies, as discussed in the section on \textbf{Fully Supervised Methods}, prove highly effective in utilizing the large volume of simulation-data pairs. Beyond the above-mentioned method, a substantial body of research has explored simulation/measurement driven methods for FWI~\cite{araya2018deep, wu-2019-inversionnet,yang2019deep, InversionNet3D-2022-Zeng, Yang-2022-Making}, USCT~\cite{donaldson21, Learned-2023-Lozenski, Li20213Dstochastic,  Jeong2023deep,jeong2023investigating, BrainPuzzle-2025-Chen, Robust-2025-Chen}, and NDT~\cite{Rachman2021, Sergio2022, Quantitative-2023-Rao}. However, a common issue with these approaches is their limited generalization ability; that is, the models' suboptimal predictability when faced with  unseen data significantly different from those in the training set. Thus, additional effort (such as transfer learning, active learning, etc.)~\cite{zhang2020data, feng2021multiscale, Renan-2022-Physics } may be required to overcome the distribution shift between simulations and real-world measurements. 
Another method that has proven effective in enhancing generalization is the ``CNN Correction after Direct Inversion'' approach, as proposed by \cite{Deep-2017-Jin}. This method initially applies model-based inversion to closely approximate the physics model of the system, producing an estimate near the true solution. Subsequently, a CNN utilizes this estimate along with high-quality labels to learn how to eliminate artifacts and thus improve reconstruction quality. \cite{jeong2023investigating} implement a similar strategy to address the USCT problem. In \cite{Robust-2025-Chen}, authors propose a deep learning framework for USCT that is trained entirely on simulation data and demonstrates successful generalization to real physical phantom experiments. A key component of the approach is the transformation of raw ultrasound waveform signals into time-shift maps, which serve as the input representation for the learning model.

\subsubsection{Physics-Aware Methods}

\begin{figure}[t]
\centerline{
    \includegraphics[width=0.5\textwidth]{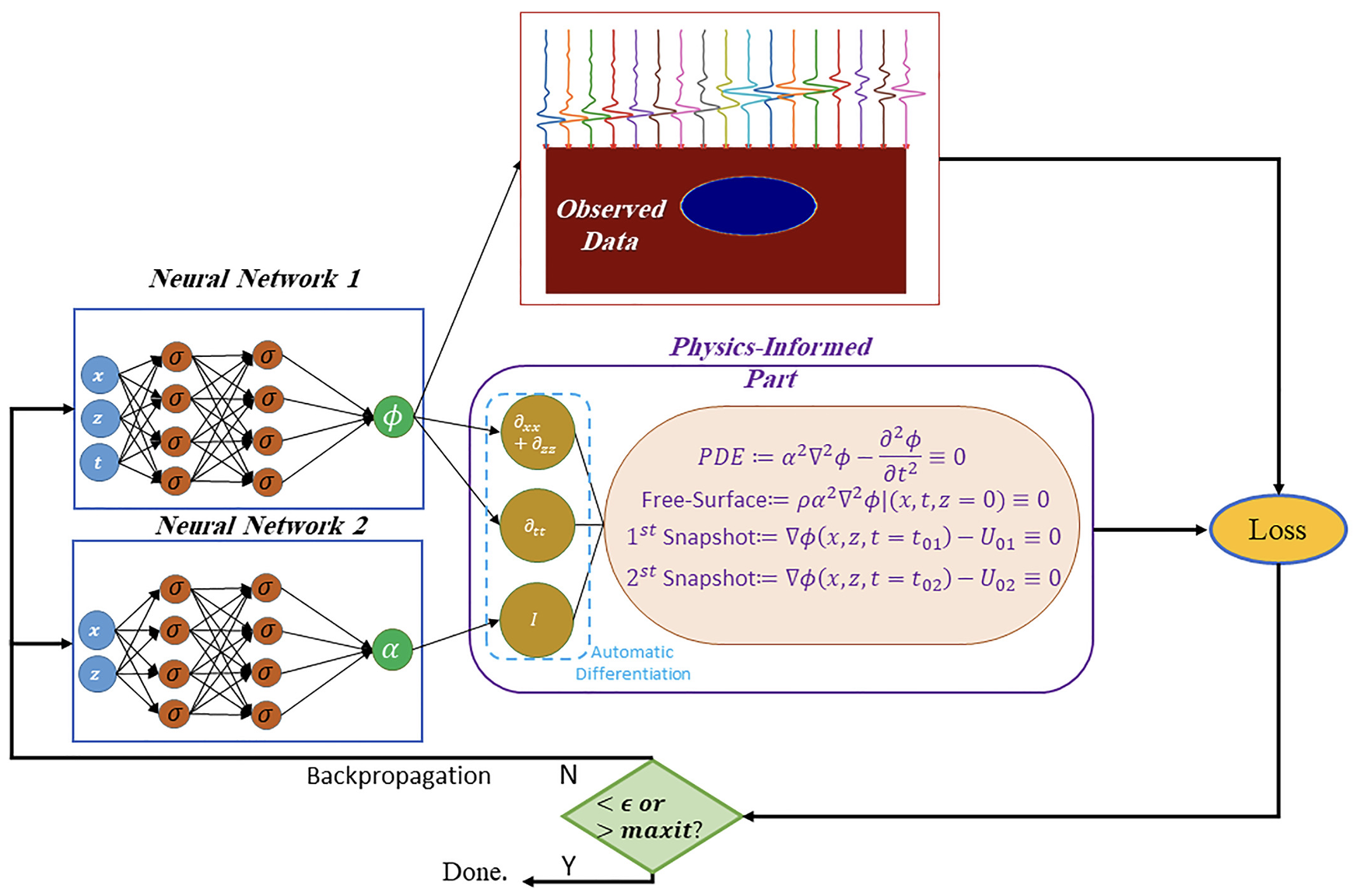}%
    }
\caption{Schematic illustration of a physics-informed neural network (PINN)~\cite{Physics-2022-Behesht}, a physics-aware method for CWI that jointly learns the wavefield and velocity model by embedding the acoustic wave equation into the network’s loss function. Image courtesy of the authors of \cite{Physics-2022-Behesht}.}
\label{fig:PhysicsAware}
\end{figure}

This category of methods integrates physical principles—most notably, the wave equation—directly into the learning framework through explicit constraints or regularization. A growing body of work has shown that enforcing the wave equation can significantly enhance inversion quality and physical interpretability in CWI~\cite{Physics-2020-Ren, Jin-2021-Unsupervised, Physics-2022-Behesht, Solving-2023-Gupta, Liu-2023-Physics, Yang-2023-FWIGAN, Elastic-2023-Dhara}.

A representative example~\cite{Physics-2022-Behesht} is the physics-informed neural network (PINN) approach~\cite{Physics-2019-Raissi}, shown in Fig.~\ref{fig:PhysicsAware}. This model jointly performs wave propagation and inversion by embedding the governing wave equation into the loss function of a neural network. The architecture includes two neural networks: the first learns the wavefield \( \phi(x, z, t) \) as a function of space and time, and the second learns the velocity model \( \alpha(x, z) \), which serves as a key physical parameter in the wave equation.

The Physics-Informed Part imposes the 2D acoustic wave equation \( \alpha^2 \nabla^2 \phi - \partial^2 \phi / \partial t^2 = 0 \), along with free-surface boundary conditions and snapshot constraints that align predicted wavefields with observed data at specific times. All differential operators are computed via \textit{automatic differentiation}, allowing the model to enforce physical constraints without requiring meshing or numerical solvers. The total loss function combines the wave equation residual, boundary losses, and waveform mismatch, and is minimized via backpropagation until convergence. This mesh-free framework provides a flexible and differentiable solution for solving inverse problems, particularly in settings with limited prior information or complex subsurface structures.

Beyond PINNs, several physics-aware models adopt alternative network structures and optimization schemes. For instance,~\cite{Physics-2020-Ren} introduces SWINet, a waveform inversion model that integrates wave equation solvers as differentiable layers and treats the velocity map as a trainable parameter. Similarly,~\cite{Yang-2023-FWIGAN} proposes FWIGAN, which fuses the wave equation with a Wasserstein GAN formulation to recover subsurface models in a generative framework. While these methods effectively embed physical constraints, they often rely on good initializations and face scalability challenges due to the high computational cost of solving wave equations, especially in 3D settings.

Physics-based regularization has also been incorporated into end-to-end learning pipelines. \cite{Jin-2021-Unsupervised, Physics-2022-Dhara, Solving-2023-Gupta, Elastic-2023-Dhara} propose encoder–decoder networks that embed physical constraints to guide learning. \cite{Solving-2023-Gupta}, in particular, develops a framework that jointly performs forward modeling and inversion using an invertible neural network. These approaches strike a balance between physical interpretability and computational tractability, and are promising for tackling multiparameter and elastic wave imaging tasks.

\subsubsection{Parameterization Solution Methods}

\begin{figure}[t]
\centerline{
    \includegraphics[width=0.5\textwidth]{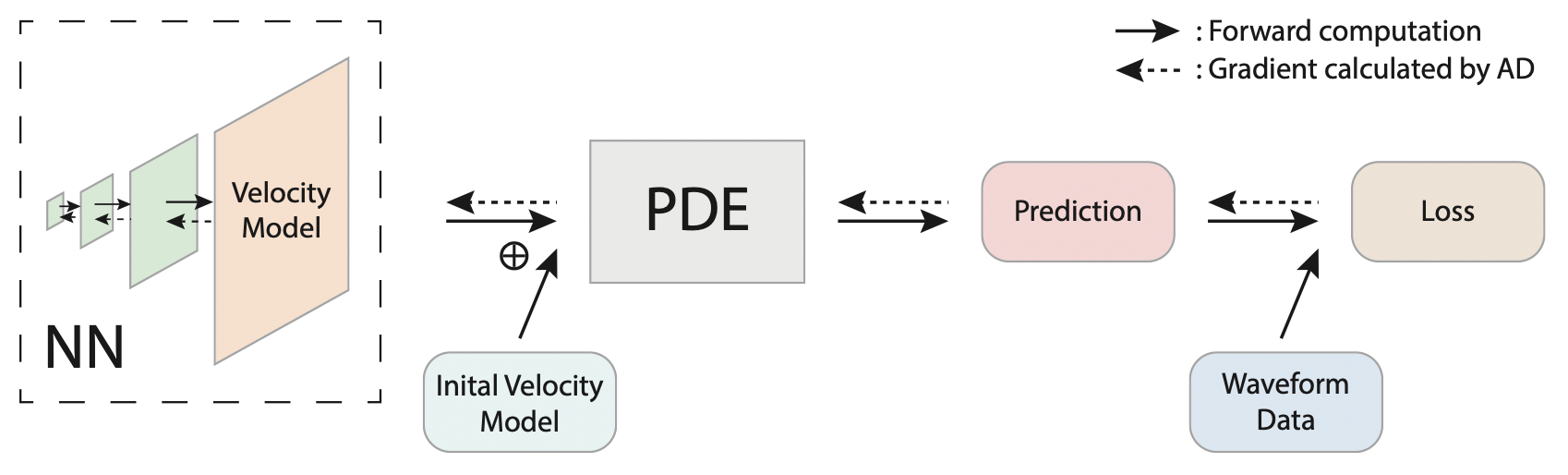}%
    }
\caption{Schematic illustration of parameterization-based method~\cite{Integrating-2021-Zhu}. Image courtesy of the authors of \cite{Integrating-2021-Zhu}.}
\label{fig:Parameterization}
\end{figure}

Physics knowledge can be represented by the parameter space of the neural networks. \cite{Deep-2020-Ulyanov} develops the deep image prior, a type of convolutional neural network, to learn useful prior knowledge via re-parameterization. A similar idea has been developed to solve CWI problems as shown in Fig.~\ref{fig:Parameterization}. In particular, velocity maps are parameterized while the forward model is used to calculate the associated waveform and obtain the data misfit. Mathematically, parameterization can be posed as
\begin{align}
    \label{eq:reparameterization}
        \widehat{\bm{m}} & = g_{\boldsymbol \theta^{*}(\bm{d})}(\bm{\beta}),\; 
        \mathrm{where}\; \boldsymbol \theta^{*}(\bm{d}) =\argmin_{\boldsymbol \theta} \mathcal{L}(f(g_{\boldsymbol \theta}(\bm{\beta})), \bm{d}) \,, 
\end{align}
where $\bm{\beta}$ is a random tensor drawn from a normal distribution; \ie, $\bm{\beta} \sim \mathcal{N}(0, \sigma^2)$ and $g_{\boldsymbol \theta^{*}}(\cdot)$ is a network model used to parameterize the velocity maps. Unlike conventional physics-based methods that optimize velocity models directly in a high-dimensional voxel space, parameterization solution methods use a neural network to implicitly represent the velocity field. This has two key advantages. First, the network architecture acts as a strong implicit prior, promoting spatial smoothness, continuity, and structural plausibility—thereby regularizing the ill-posed inverse problem. Second, although the total number of network parameters can be large, the optimization is carried out over a lower-dimensional latent space (\eg, via $\bm{\beta}$ or a compact representation), and the network maps this to a structured velocity model. This effectively restricts the solution space to a subset of physically plausible models, improving convergence stability and robustness to noise.

Parameterization solution methods, building on conventional physics-based approaches, inherit similar challenges. Typically, these methods require a well-defined initial velocity map, and factors like regularization and optimization techniques (\eg, gradients and step sizes) significantly influence their convergence. Specifically, crucial to Eq.~(\ref{eq:reparameterization}) are the selection of $\bm{\beta}$ and the network structure as in $g_{\boldsymbol \theta^{*}}(\cdot)$. Early work by \cite{wu2019parametric} samples $\bm{\beta}$ from noise and used CNNs. \cite{he2021reparameterized} adopts a similar approach but with a deeper theoretical focus. Both methods involve a two-step process: network pre-training with a specific initial guess, followed by fine-tuning. The discretization assumption limits the output to fixed sizes, restricting transferability. These limitations have motivated recent developments to improve flexibility and physical fidelity. Overcoming these drawbacks, \cite{Integrating-2021-Zhu} develops a new strategy that integrates the wave equation for regularization, eliminating the pre-training step and addressing the ill-posed nature of CWI problems. More recently, \cite{Implicit-2023-Sun} introduces an approach that employs a continuous implicit function via neural networks to represent velocity maps, accommodating arbitrary sizes.

\begin{figure*}[ht]
\centerline{
    \includegraphics[width=\textwidth]{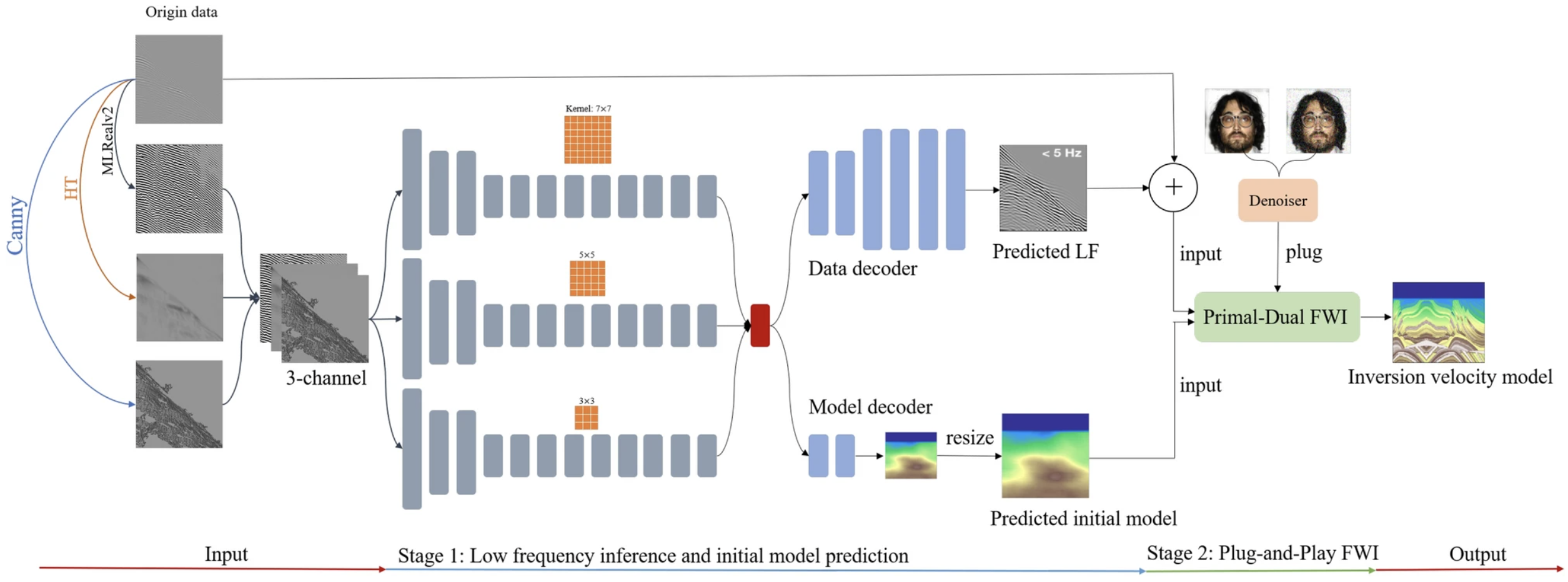}%
    \label{fig:PPP_CWI}}
\caption{Schematic illustration of the network architecture of PPP-CWI~\cite{Progressive-2025-Zhang}, a Plug-and-Play Priors framework for CWI that alternates between physics-consistent updates and learned prior denoising. Image adapted from~\cite{Progressive-2025-Zhang}.}
\label{fig:ppp}
\end{figure*}

\subsubsection{Plug and Play Priors Methods}

The Plug-and-Play Priors (PPP) framework, initially developed in the computational imaging community~\cite{venkatakrishnan2013, sreehari2016TCI, kamilov-2023-plug}, provides a flexible and powerful approach for solving inverse problems by integrating advanced image denoisers into iterative optimization algorithms. 
In this framework, the denoiser effectively replaces an explicitly designed regularization term, thereby acting as an implicit data-driven prior rather than a hand-crafted analytic regularizer.
Originally designed to incorporate classical denoisers such as BM3D~\cite{DabovBM3D07}, PPP has since evolved to support modern learning-based models like DnCNN~\cite{zhang2017beyond}, enabling the use of expressive neural network priors without requiring problem-specific retraining.

PPP is rooted in proximal optimization techniques~\cite{parikh2014proximal}, particularly the Alternating Direction Method of Multipliers~(ADMM)~\cite{glowinski1975approximation, gabay1976dual, boyd2011distributed}, which solves inverse problems of the form:
\[
\min_{\mathbf{m}} \; h(\mathbf{m}) + r(\mathbf{m}),
\]
where \( h(\cdot) \) typically represents a data fidelity term and \( r(\cdot) \) serves as a regularization prior.  Within the PPP formulation, this explicit regularization term is not specified analytically; instead, its proximal operator is replaced by a denoising operator. The proximal operator, defined in~\cite[Ch. 6]{beck2017first}, is interpreted as a maximum a posteriori (MAP) estimator. This interpretation allows replacing it with advanced denoisers, including learned networks, while preserving algorithmic structure. Although this substitution breaks the strict optimization interpretation, it has been shown to converge under certain conditions~\cite{sreehari2016TCI, Chan2017, ryu2019plugandplay}.

In CWI, PPP has been adapted to handle the challenges of ill-posedness and data inconsistency. A representative framework, PPP-CWI~\cite{Progressive-2025-Zhang}, proceeds in two stages (shown in Fig.~\ref{fig:ppp}): an initial estimate is obtained from low-frequency data, followed by iterative refinement via a PPP-based scheme. The objective is:
\[
\min_{\bm{m}} \; g(\bm{m}) + r(\bm{m}), \text{with } g(\mathbf{m}) = \frac{1}{2} \| f(\bm{m}) - \bm{d} \|_2^2,
\]
where \( f(\cdot) \) is the forward modeling operator and \( \bm{d} \) denotes observed data.

The PPP-CWI algorithm iterates over three steps:

\begin{enumerate}
    \item Model update (physics-consistent step):
    \begin{equation}
\mathbf{m}^{(k+1)} = \arg \min_{\mathbf{m}} \; g(\mathbf{m}) + \frac{\rho}{2} \left\| \mathbf{m} - \mathbf{y}^{(k)} + \mathbf{u}^{(k)} \right\|_2^2,
\end{equation}
typically solved using a first-order update:
\begin{equation}
\mathbf{m}^{(k+1)} = \mathbf{m}^{(k)} - \eta \left( \nabla g(\mathbf{m}^{(k)}) + \rho (\mathbf{m}^{(k)} - \mathbf{y}^{(k)} + \mathbf{u}^{(k)}) \right),
\end{equation}
where \( \eta \) is the learning rate and \( \nabla g(\mathbf{m}) \) is computed using the adjoint-state method.

    \item Denoising update (prior-consistent step):
\begin{equation}
\mathbf{y}^{(k+1)} = \mathcal{D}_\theta \left( \mathbf{m}^{(k+1)} + \mathbf{u}^{(k)} \right),
\end{equation}
where \( \mathcal{D}_\theta \) is a pretrained denoising network (\eg, DRUNet~\cite{Zhang2022Plug} or DnCNN~\cite{zhang2017beyond}).

    \item Dual update:
\begin{equation}
\mathbf{u}^{(k+1)} = \mathbf{u}^{(k)} + \mathbf{m}^{(k+1)} - \mathbf{y}^{(k+1)}.
\end{equation}

\end{enumerate}

This iterative scheme decouples the enforcement of physical consistency from the enforcement of prior knowledge, allowing both to be handled in a modular and flexible manner. Such decoupling is particularly valuable in CWI applications, where the forward model is highly complex, and the inversion problem is severely ill-posed. Traditional physics-based CWI methods often suffer from poor convergence due to non-convexity, high sensitivity to the initial model, and instability in the presence of noise or incomplete data. PPP mitigates these issues by incorporating powerful, data-driven priors, which is often learned from large datasets, directly into the inversion loop. These priors guide the solution toward physically plausible reconstructions, reduce reliance on handcrafted regularization terms, and improve robustness to data imperfections. Furthermore, by leveraging pretrained denoisers, PPP methods enhance generalization across different domains while reducing the need for extensive manual tuning. This makes them a promising and scalable alternative to conventional CWI, especially in large-scale or real-time imaging scenarios.

Several recent works have successfully applied PPP in CWI. \cite{aghamiry2020hybrid} combines BM3D regularization with a proximal Newton solver to improve robustness in FWI, although performance degrades under complex noise. Learning-based alternatives have shown promise: \cite{Fu-2023-Efficient} demonstrates that deep CNN priors outperform handcrafted regularization such as total variation, and \cite{Izzatullah-2023-Plug} introduces a CNN-driven posterior sampling scheme to enhance uncertainty quantification in CWI.

\subsubsection{Comparative Overview of Learning Strategies for CWI}

This subsection provides a concise synthesis of the four learning strategies discussed in this section, namely, \textit{Simulations/Measurements-Driven}, \textit{Physics-Aware}, \textit{Parameterized Solution}, and \textit{Plug-and-Play Priors}. Each of these approaches brings distinct advantages and challenges. Summarizing their characteristics helps guide practitioners in selecting or combining appropriate strategies for specific CWI tasks.

\textbf{Simulations/Measurements-Driven.~}These methods rely on supervised learning with paired data obtained from simulations or experimental measurements. Their primary strength lies in their ability to directly learn complex mappings from waveforms to physical properties when a large, diverse training dataset is available. However, their performance can degrade significantly under distribution shifts, particularly when real-world data differ from synthetic training conditions. These methods are most effective in well-controlled environments where high-quality paired data can be curated.

\textbf{Physics-Aware.~}Physics-aware approaches incorporate physical knowledge, such as wave equations, either as soft constraints (\eg, through loss functions) or embedded in the architecture itself. By doing so, they reduce reliance on labeled data and improve generalization to unseen scenarios. However, these methods often involve more complex optimization procedures and may incur higher computational costs. They are particularly advantageous when physical models are well established, but labeled data are limited or expensive to obtain.

\textbf{Parameterized Solution.~}In this strategy, a neural network is used to parameterize the velocity model, which is then optimized through a physics-based loss involving the forward model. The network acts as an implicit prior, constraining the inversion process to a lower-dimensional, structured space. This can improve convergence and regularization. Nonetheless, the approach is sensitive to the choice of network architecture, initialization, and training dynamics. It is well-suited for applications where capturing global structure or enforcing smoothness is critical.

\textbf{Plug-and-Play Priors.~}Plug-and-play methods decouple the physics and data-driven components, allowing for the use of pretrained denoisers or generative models as implicit priors within iterative solvers. This modularity enables flexible integration and leverages powerful priors without altering the physics simulator. The trade-off, however, lies in computational overhead and the need to carefully balance the influence of priors and data consistency. These methods are promising in scenarios where data priors are rich, but direct supervision is limited.

\textbf{Toward Strategy Selection and Integration.~}Each learning strategy offers a distinct balance among generalization ability, interpretability, data efficiency, and computational cost. Selecting an appropriate strategy depends not only on theoretical considerations but also on the practical constraints of the specific CWI application. For instance, in data-rich but poorly modeled environments, simulation-driven methods may suffice. Conversely, in data-sparse domains with reliable physics, physics-aware or parameterized methods are often more effective.

Recent trends suggest that hybrid approaches can leverage the strengths of multiple strategies. For example, plug-and-play frameworks can incorporate physics-aware denoisers or generative priors, such as score-based diffusion models, to improve inversion robustness without retraining the entire pipeline. Similarly, parameterized solvers can benefit from pretraining on synthetic data followed by fine-tuning with physics-constrained optimization, improving generalization while maintaining physical consistency.

Ultimately, strategy selection should consider factors such as the availability and quality of labeled data, the fidelity and computational cost of the forward model, the required level of interpretability, and tolerance to distribution shift. Designing adaptive systems that can switch between or combine strategies during training or inference remains an open but promising area for future research in CWI.

\subsection{Emerging Trends in Machine Learning Techniques for Computational Wave Imaging}

In this section, we examine two important advancements in ML-enhanced CWI methods: Uncertainty Quantification (UQ) and Generative AI (GenAI) Models. These developments offer significant opportunities to further enhance CWI techniques.

\subsubsection{Uncertainty Quantification~(UQ)}

Uncertainty is a fundamental feature of CWI and of inverse problems at-large: not only do we wish to infer the model-parameters, but we also want to quantify the uncertainty associated with this inference, reflecting the degree of confidence we have in the inversion. 
CWI problems are typically ill-posed and non-convex, often yielding non-unique or suboptimal solutions. Furthermore, real-world waveform measurements are generally compromised by acquisition noise from equipment, environmental disturbances, and other sources. These uncertainties are often categorized as epistemic (model-related) and aleatoric (data-related) uncertainties~\cite{Single-2019-Tagasovska}. Epistemic uncertainty arises from unknown or unmodeled physics, such as the excitation pulse, detector characteristics, acoustic attenuation, density variations, and nonlinear effects. In contrast, aleatoric uncertainty reflects the inherent variability in the data itself. UQ mitigates these issues by providing predictive distributions and assessing prediction confidence. 

In recent years, the Bayesian inverse problem has emerged as the most comprehensive and systematic framework for formulating and solving inverse problems with quantified uncertainties~\cite{Tarantola05, KaipioSomersalo05, Stuart10, OdenMoserGhattas10, BieglerBirosGhattasEtAl11}. However, the solution of Bayesian inverse problems is extremely challenging; when the forward model is complex and the parameter dimension is large, Bayesian inversion becomes prohibitive with standard methods. Exploration of the posterior distribution using Markov chain Monte Carlo (MCMC) sampling is often computationally infeasible for CWI and often requires advanced sampling algorithms that leverage curvature information (geometric MCMC)\cite{MartinWilcoxBursteddeEtAl12,Bui-ThanhGhattasMartinEtAl13,UlrichBoehmZuninoetal22,bates2022probabilistic}. Variational inference approaches introduce a parametrized approximation of the posterior distribution and solve a high dimensional optimization problem to minimize the Kullback-Leibler (KL) divergence between the approximated and true posterior\cite{BleiKucukelbirMcAuliffe17}. However, in the context of CWI, variational inference approaches have been limited to two-dimensional synthetic problems \cite{ZhangCurtis20} and entail significant computational cost due to the ``curse of dimensionality'' in approximating a probability distribution and its expectations for a high-dimensional parameter space.

ML offers new opportunities to enable UQ for CWI by use of novel network architectures and approaches, such as normalizing flows\cite{rezende2015variational,papamakarios2021normalizing}, deep probabilistic imaging \cite{sun2021deep}, and generative AI (see next section), that can help to break the curse of dimensionality and enable high-dimensional UQ. Here, we explore UQ estimation using neural networks, specifically through a method known as InvNet\_UQ~\cite{Enhanced-2023-Liu}. This approach focuses on aleatoric uncertainty and incorporates Simultaneous Quantile Regression~(SQR) within a CNN framework~\cite{Single-2019-Tagasovska}. InvNet\_UQ employs a novel joint quantile regression loss function to enhance CWI by improving model reliability and quantifying prediction confidence. The network parameters of InvNet\_UQ are obtained through the following optimization
\begin{align}
    &\boldsymbol{\theta}^* \in \argmin_{\theta} \nonumber\\
    &
    \left \{ \frac{1}{pqr} \sum_{k = 1}^{p}\sum_{i = 1}^{q}\sum_{j = 1}^{r} E_{\tau \approx U[0,1]} [\mathcal{L}_{\tau} (g_{\boldsymbol{\theta}}(\mathbf{d}_k, \tau)(i,j), \mathrm{\mathbf{m}}_k(i,j))] \right \},
\end{align}
where $k$ denotes $k$th sample in the dataset, $(i,j)$ denotes the pixel location, and the pinball (quantile) loss $\mathcal{L}_{\tau}$, defined below, is designed to capture the discrepancy between predicted and actual values across different quantile levels of ${0 \leq \tau \leq 1}$. 
\begin{equation}
    \mathcal{L}_{\tau}(\mathbf{m}(i,j), \hat{\mathbf{m}}(i,j)) = 
    \begin{cases}
        \tau (\mathbf{m}(i,&j) - \hat{\mathbf{m}}(i,j)) \\
        &    \text{if ${\mathbf{m}(i,j)- \hat{\mathbf{m}}(i,j))  \geq 0}$ },\\
        (1-\tau)& (\hat{\mathbf{m}}(i,j)- \mathbf{m}(i,j))\\
        &\text{else}.
    \end{cases}
\end{equation}
Provided with a significance level ${\alpha}$, the estimated imaging uncertainty can be computed from the (${1-\alpha}$) prediction interval around the median
\begin{equation}
    u_\alpha(\mathbf{d}):=\hat{g}\left(\mathbf{d}, 1-\frac{\alpha}{2}\right) - \hat{g}\left(\mathbf{d}, \frac{\alpha}{2}\right).
    \label{eq:UQ}
\end{equation}

\cite{Enhanced-2023-Liu} shows that data-driven models with UQ excel in handling contaminated data, including noisy or erroneous measurements. Their approach not only boosts confidence in inversion tasks but also refines training sets, enhancing model performance. Unlike Bayesian neural networks, InvNet\_UQ does not necessarily require the image prior to be a Gaussian distribution, potentially providing more accurate uncertainty bounds compared to other methods. Beyond the use of quantile regression for UQ, Monte Carlo dropout---a technique favored for its simplicity---is also widely used in CWI. \cite{Integrating-2021-Zhu} implements it by adding dropout layers to their network. Similarly, \cite{Um-2022-Deep} uses a U-net architecture, employing both Monte Carlo dropout and bootstrap aggregating for uncertainty estimation. \cite{Uncertainty-2023-Yablokov} uses a multi-layer fully connected network and Monte Carlo simulations to estimate uncertainties.

\subsubsection{Generative AI (GenAI)}

GenAI models excel in representing probability distributions across diverse data domains and present new opportunities for CWI, and computational imaging in general. By learning an accurate characterization of the model parameter prior distribution, GenAI can 1) enhance the solution of the Bayesian inverse problem and UQ~\cite{Parameterizing-2020-Rizzuti, Deep-2022-Siahkoohi, Prior-2023-Wang, Yang-2024-EdGeo}; 2) reduce ill-posedness of CWI by constraining the model parameter to belong to a high-probability manifold\cite{mosser2020stochastic, zhang2020data,DiGiT-2024-Yang}; and 3) boost the performance of learning methods by generating synthetic data to augment training sets\cite{Yang-2022-Making}. 

GenAI approaches can be broken down into two broad categories: sample-based approaches and score-based approaches.   
Sample-based approaches, such as Variational Auto-Encoders ~\cite{Kingma-2014-Auto} (VAEs) and Generative Adversarial Models~\cite{Goodfellow-2014-Generative} (GANs) learn the push forward from an easy-to-sample latent distribution (\eg, a multivariate \emph{i.i.d.} Gaussian) to the data distribution. Once trained, they allow for \emph{direct} (and computationally efficient) sampling. GANs, for example, were successfully employed to reduce the ill-posedness of CWI in the work of \cite{zhang2020data} and \cite{mosser2020stochastic} by leveraging the GAN latent space as a nonlinear dimension reduction technique, while VAEs were successfully employed in \cite{Yang-2022-Making} for data augmentation. However, it is worth noticing that GAN and VAE approaches are generally limited to small/medium scale problems and may suffer from the so-called \emph{mode collapse} phenomenon~\cite{kodali2017convergence, srivastava2017veegan, yacoby2020failure}, which prevents them from fully capturing the distribution of the data.
Score-based approaches overcome this by directly learning the Stein score of the data distribution (\ie, the gradient of the logarithm of the probability distribution function), which can be used in iterative sampling algorithms such as MCMC~\cite{neal2011mcmc}. In particular, diffusion models \cite{Sohl-2015-Diffusion, Ho-2020-Denoising, Song-2022-Solving} are now recognized as state-of-the-art in GenAI because of their high-quality sample generation, scalability to high-dimensional distributions, stability, and simplicity.

\cite{Prior-2023-Wang} uses a diffusion model for solving CWI problems with quantified uncertainties. This method extends \cite{Song-2022-Solving} for solving linear inverse problems (like CT and MRI image reconstruction) using diffusion models to the more challenging CWI problem, which is governed by a highly non-linear forward model. In this approach, a diffusion model is first constructed from velocity model data. Diffusion models consist of two components: a \emph{forward} diffusion process that uses a stochastic ordinary differential equation (SDE) to map the data distribution to a fixed noise distribution, and a \emph{reverse} diffusion process that maps the noise distribution to the data distribution. In particular, \cite{Prior-2023-Wang} adopts the variance preserving (VP) SDE \cite{song2021scorebased} for the forward diffusion process, which can be written as
\begin{equation}
    d\mathbf{m}_t = -\frac{\beta_t}{2}\mathbf{m}_t dt + \sqrt{\beta_t} d \mathbf{w}_t, \quad t \in (0, T)
\end{equation}
where the parameter $\beta_t$ controls the noise variance, the initial condition $\mathbf{m}_0$ of the SDE is the velocity model, and $d\mathbf{w}_t$ is the noise introduced. The corresponding reverse diffusion process depends on the Stein score of the probability density of $\mathbf{m}_t$, which can be approximated by training the parameters $\theta$ of a vector-valued neural network $s_{\theta}(\mathbf{m}_t, t)$ using the cost function
 \begin{equation}
\frac{1}{T}\int_0^T \lambda (t) \mathbb{E}_{\mathbf{m}_t | \mathbf{m}_0} \mathbb{E}_{\mathbf{m}_0} \left[ 
\left\| s_{\theta}(\mathbf{m}_t, t) + \frac{\mathbf{m}_t - e^{-\frac{1}{2} \int_{0}^t \beta(s) ds }\mathbf{m}_0}{1 - e^{- \int_{0}^t \beta(s) ds}}\right\|_2^2
\right]\, dt,
 \end{equation}
 where $\lambda$ is a positive weighting function. 
 Once trained, the network can generate new samples from the prior distribution of the model parameter, as the solution of another SDE (reverse diffusion process) as
\begin{equation}
    d\mathbf{m}_t = \left(-\frac{\beta_t}{2}\mathbf{m}_t  - \beta_t s_{\theta}(\mathbf{m}_t, T- t)\right) dt + \sqrt{\beta_t} d \mathbf{w}_t, \quad t \in (0, T),
\end{equation} 
using white noise at the initial condition.
To condition the sample generation to the measurement data, \cite{Prior-2023-Wang} include an FWI update after each step of discretized SDE solver to enhance data consistency. 

\section{Computational Wave Imaging Applications}

CWI is widely used in various scientific and engineering fields. This section provides examples of employing these methods to solve real-world problems. Appendix~\ref{sec:Appendix-E} includes source codes, packages, and training datasets, offering practical examples and a foundation for readers to address their own CWI challenges.

\subsection{Seismic Full-Waveform Inversion}

Seismic inversion, a tool for visualizing subsurface structures, is crucial in identifying and characterizing underground reservoirs. It also aids in diverse geological and environmental studies. This method involves creating precise velocity maps, which are essential for advanced post-processing techniques like reverse-time migration, to extract high-resolution details from subsurface data.

\begin{figure*}[t]
	\centerline{
	\subfloat[]{\includegraphics[width=.25\linewidth]{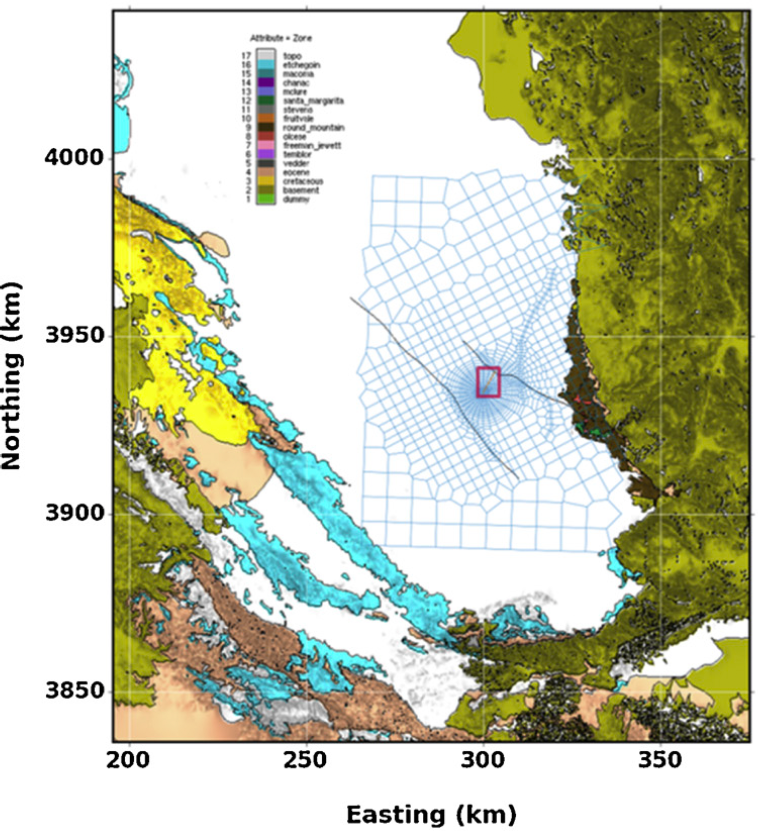}}
	\quad \quad \quad \quad
	\subfloat[]{\includegraphics[width=.15\linewidth]{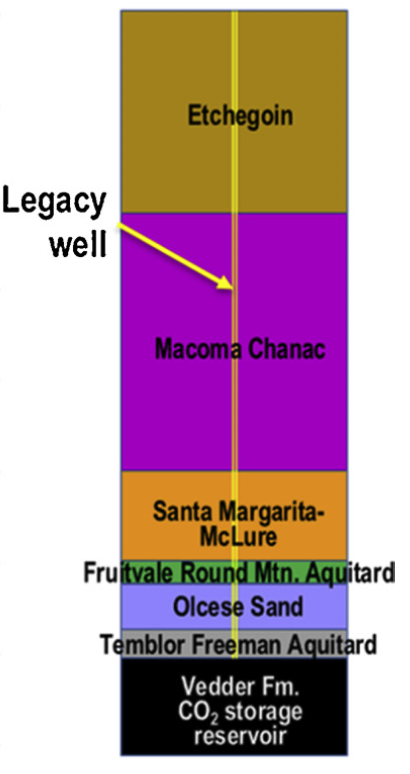}}
	\quad \quad \quad \quad
	\subfloat[]{\includegraphics[width=.32\linewidth]{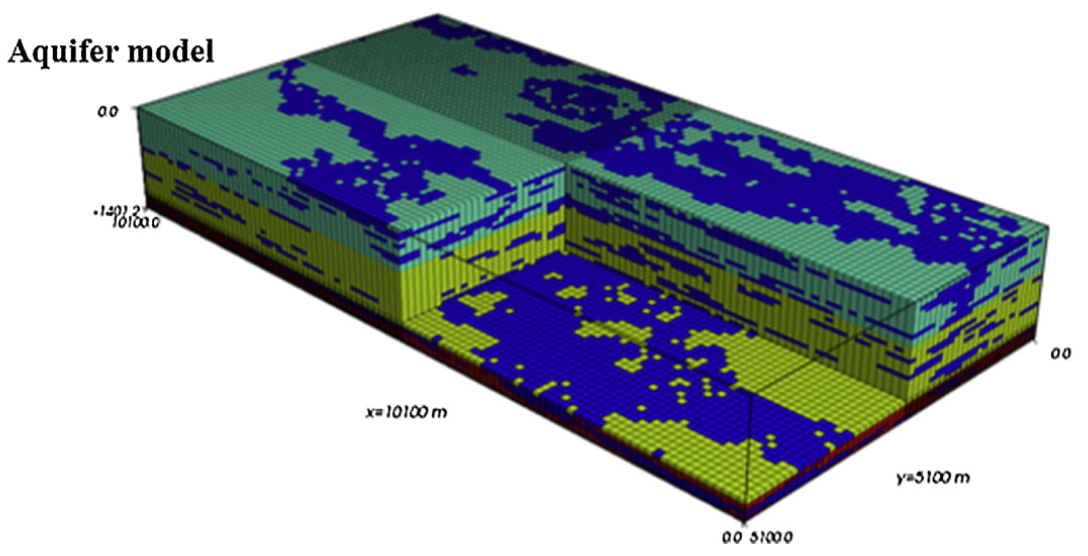}}}
	\centerline{
	\subfloat[]{\includegraphics[width=\linewidth]{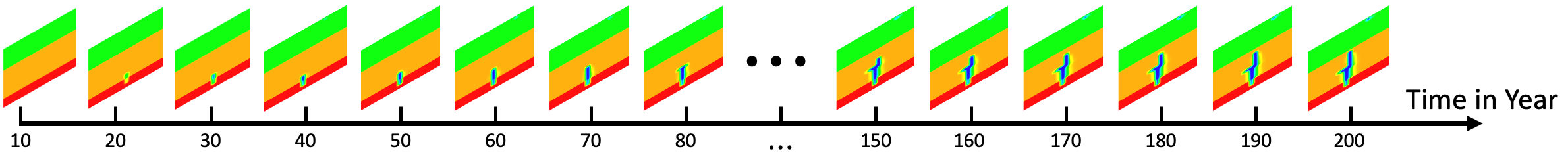}}}
	\centerline{
	\subfloat[]{\includegraphics[width=\linewidth]{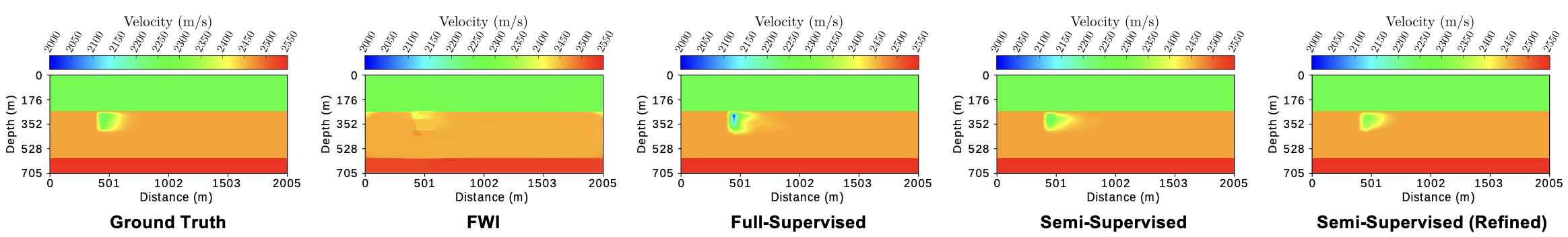}}}
	\caption{Illustration of the Kimberlina dataset and three modeling modules used to generate the simulated velocity maps. (a) CO$_{2}$ storage reservoir model, (b) wellbore leakage model, (c) multi-phase flow and reactive transport models of CO$_{2}$ migration in aquifers~\cite{Downhole-2019-Buscheck, Assessment-2019-Yang}, and (d) Illustration of a set of simulation with 20 velocity maps for 200 years. A CO$_2$ leakage will result in a decrease in the velocity value in the location where the leak happens. (e) Visualizations of the early CO$_2$ leakage plume (from left to right): ground truth, results of physics-based FWI, fully-supervised model~\cite{wu-2019-inversionnet}, semi-supervised model~\cite{Renan-2022-Physics}, and a refined semi-supervised model~\cite{Renan-2022-Physics}.}
	\label{fig:kimberlina}
\end{figure*}

\subsubsection{Leakage Detection for Carbon Sequestration}

Geologic carbon sequestration~(GCS), crucial in climate change mitigation, involves capturing and storing CO$_2$. Seismic imaging is key for early detection of CO$_2$ leakage, essential for project success. Effective leakage detection hinges on identifying leak presence, location, and magnitude. While traditional seismic methods are commonly used, they often lack early detection precision. Deep learning has brought significant improvements. \cite{Data-2019-Zhou} develops a method for direct CO$_2$ leakage quantification from seismic data. \cite{Renan-2022-Physics} introduces a semi-supervised approach targeting small leaks and imbalanced data scenarios. \cite{Yang-2022-Making} employs a generative model to enhance rare event detectability. Moving beyond indirect inference from seismic data, \cite{zhong2020inversion} uses a generative adversarial network for mapping dynamic saturation changes. \cite{leong2022estimating} create a network for direct CO$_2$ saturation mapping from seismic data. \cite{Um-2022-Real,Um-2022-Deep} combine seismic, electromagnetic, and gravity data for a comprehensive saturation model with uncertainty estimation. \cite{Feng-2022-Exploring} links seismic and electromagnetic data for improved CO$_2$ saturation modeling.

The Kimberlina dataset, generated by the National Risk Assessment Partnership project of the U.S. Department of Energy, is based on the geologic structure of a commercial-scale GCS reservoir at the Kimberlina site in the southern San Joaquin Basin, CA, USA. The simulation includes four modules: a CO$_2$ storage reservoir model (Fig.~\ref{fig:kimberlina}(a)), a wellbore leakage model (Fig.~\ref{fig:kimberlina}(b)), multi-phase flow and reactive transport models of CO$_2$/brine migration in aquifers (Fig.~\ref{fig:kimberlina}(c)), and a geophysical model. 

The P-wave velocity maps used in this study, which form part of the geophysical model, are constructed from the stratigraphic and lithologic information at the Kimberlina GCS site (Fig.~\ref{fig:kimberlina}(b))~\cite{Downhole-2019-Buscheck, Assessment-2019-Yang}. Fig.~\ref{fig:kimberlina}(d) shows a set of simulated velocity maps across 20 time steps spanning 200 years. To assess different inversion strategies, Fig.~\ref{fig:kimberlina}(e) compares the reconstructed velocity fields using three methods: physics-based seismic full waveform inversion (FWI)~\cite{Acoustic-2015-Lin}, a fully supervised learning approach~\cite{wu-2019-inversionnet}, and a semi-supervised learning method incorporating adaptive data augmentation~\cite{Renan-2022-Physics}. 

The results highlight the benefits of the adaptive augmentation strategy, which can be applied iteratively to refine CWI reconstructions. In each iteration, the current model generates approximate velocity maps from the measured waveform data, which are then used to synthesize new labeled waveform-velocity pairs via forward modeling. These new samples are added to the training set, and the model is retrained. This cycle is repeated, with each round progressively tailoring the training data to the target recorded waveform, thereby improving inversion accuracy. As shown in Fig.~\ref{fig:kimberlina}(e), this process helps better localize and delineate the CO$_2$ leakage plume, particularly in its early stages. Additional implementation details of the adaptive augmentation procedure are provided in Appendix~\ref{sec:Appendix-D}. In summary, the physics-based FWI lacks the resolution to capture small-scale leakage features; the fully supervised method suffers from shape distortion; whereas the semi-supervised approach delivers more accurate and physically consistent reconstructions.


\begin{figure*}[t]
	\centerline{\includegraphics[width=0.75\linewidth]{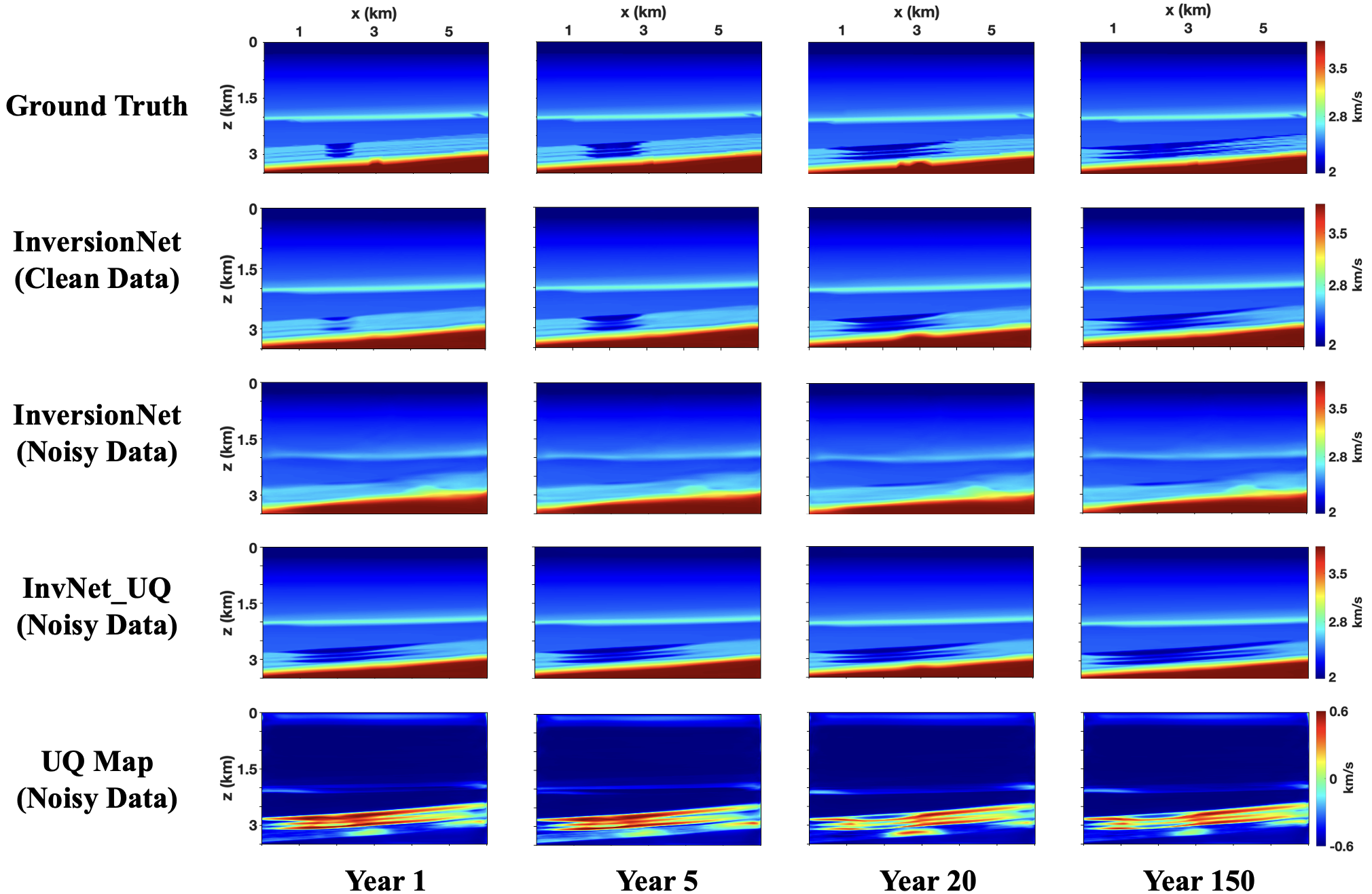}}
	\caption{Illustration of seismic full-waveform inversion with uncertainty quantification applying to noisy data. Ground Truth (Row 1); Results obtained using InversionNet~\cite{wu-2019-inversionnet} on clean data (Row 2); Results obtained using InversionNet~\cite{wu-2019-inversionnet} on noisy data with SNR=10 (Row 3);  Results obtained using InvNet\_UQ~\cite{Enhanced-2023-Liu} on noisy data with SNR=10 (Row 4); Uncertainty map obtained using \cite{Enhanced-2023-Liu} (Row 5).}
	\label{fig:FWI_UQ}
\end{figure*}

\subsubsection{Seismic Full-Waveform Inversion with Uncertainty Quantification}

Robustness to noise is essential for CWI and broader inversion methods. It is important to distinguish ``adversarial noise''---deliberate perturbations meant to deceive ML models---from ``measurement noise,'' which naturally occurs during data collection. Since the latter is far more relevant to CWI, we consider adversarial scenarios to be beyond the scope of this review. In our second seismic FWI example, we specifically address measurement noise. To improve ML models' resistance to this, researchers have developed strategies such as UQ~\cite{Enhanced-2023-Liu}, advanced physics-guided regularization~\cite{Jin-2021-Unsupervised}, and training with noisy data. We focus on the UQ method InvNet\_UQ by \cite{Enhanced-2023-Liu}.

The effectiveness of InvNet\_UQ is demonstrated in Fig.~\ref{fig:FWI_UQ}, where it is applied to data contaminated with noise at a Signal-to-Noise Ratio (SNR) of 10. This is compared to results from InversionNet~\cite{wu-2019-inversionnet}, revealing a significant performance decline for InversionNet under noisy conditions. In contrast, InvNet\_UQ maintains higher accuracy in its predictions, showcasing its superior robustness to noise. 

In addition to improved prediction quality, InvNet\_UQ also generates uncertainty maps, as shown in the bottom row of Fig.~\ref{fig:FWI_UQ}. These maps quantify the spatial confidence of the inversion output and serve multiple purposes. They can help practitioners identify regions of high uncertainty, informing risk-aware decision-making in geophysical and subsurface applications. Furthermore, these uncertainty estimates can guide the adaptive selection of training data or even act as a supervisory signal to train auxiliary models. Beyond interpretability, the ability to model predictive uncertainty also contributes directly to more robust inversion results. By estimating a conditional distribution over possible outputs, rather than a single point estimate, InvNet\_UQ mitigates the risk of overfitting to noise or making overconfident errors in ill-posed regions. This allows the model to “hedge” its predictions in ambiguous areas, producing smoother and more stable reconstructions. The uncertainty maps shown in Fig.~\ref{fig:FWI_UQ}, calculated using Eq.~(\ref{eq:UQ}), thus provide not only interpretability but also a mechanism to improve the inversion pipeline itself. As \cite{Enhanced-2023-Liu} have shown, this uncertainty information can be leveraged to enrich training data, potentially using generative AI models, thereby improving overall model performance.


\subsection{Acoustic and Industrial Ultrasonic Imaging}

Acoustic and industrial ultrasonic imaging is widely used for inspecting and monitoring engineered structures. Its applications range from detecting small defects in 3D printed materials to identifying structural issues in large airplane wings and locating pipeline leaks. The process involves using transmitters to emit signals and receivers to capture these waves post-propagation through the structure. Analyzing signal characteristics like arrival time, amplitude, and distortion through computed tomography helps create images representing structural properties such as sound speed, density, and temperature.

\subsubsection{Acoustic and Industrial Ultrasonic Imaging for Defect Detection and Characterisation}

As industry moves towards digitization, the resulting large datasets make ML-based inspection techniques increasingly relevant. Although useful and cost-effective for automating routine NDT processes, the use of ML is nascent in key industrial areas. Notably, in industrial ultrasonic inspection, most ML methods prioritize data-driven over physics-based models~\cite{Rachman2021, Sergio2022}. In the NDT supervised method, \cite{Ryu2023} proposes a regression-based multilayer perceptron, which was formulated to estimate the plastic properties (\eg, yield strength) of aluminum alloys, utilizing data from ultrasonic tests. \cite{wang2022ultrasonic} utilizes CNN for an ultrasonic framework to accurately pinpoint corrosion damage dimensions and locations, despite some forecasting inaccuracies. They also suggested reconstructing sparse signals from limited transducers through compressed sensing and CNN-based deep learning~\cite{Wang2022MSSP}. In addition, a semi-supervised anomaly detection model was introduced to establish a baseline anomaly detection score for the ultrasonic testing dataset~\cite{Rudolph2021}. Authors made adjustments to these networks to enhance their efficiency in addressing specific challenges. \cite{Shukla2020JNE} presents a physics-informed neural network~(PINN) to detect surface-breaking cracks in metal plates based on ultrasonic data. Compared with supervised methods, there is very limited literature available on semi-supervised and unsupervised methods for NDT. \cite{Luiken2023} uses a self-supervised denoiser to develop a method for flexible ultrasonic data acquisition, enhancing speed while preserving image quality.

\begin{figure*}[t]
	\centerline{
	\subfloat[]{\includegraphics[width=0.9\linewidth]{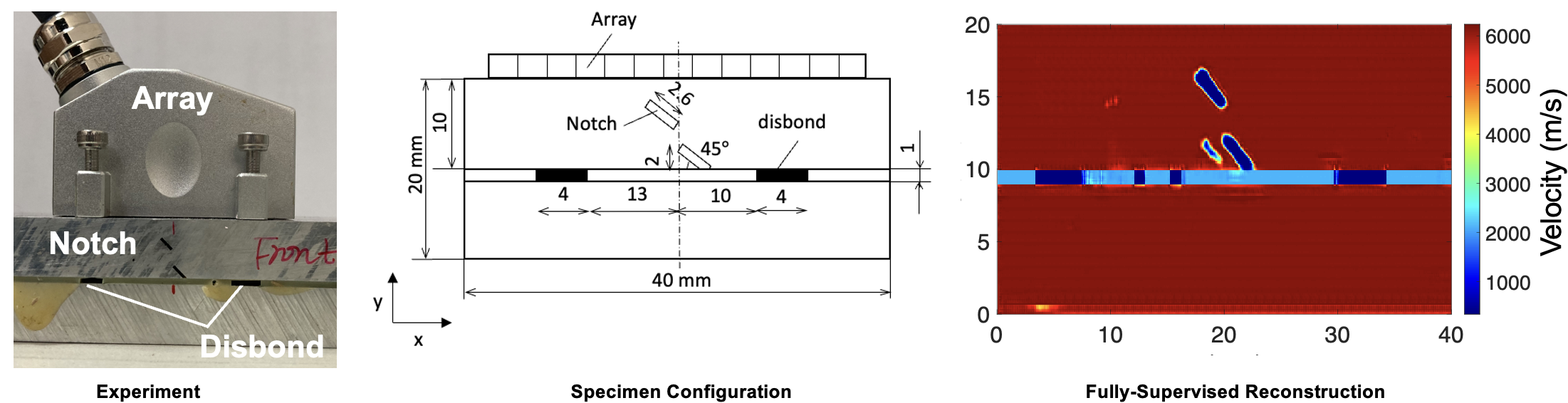}}}
 	\centerline{
	\subfloat[]{\includegraphics[width=0.9\linewidth]{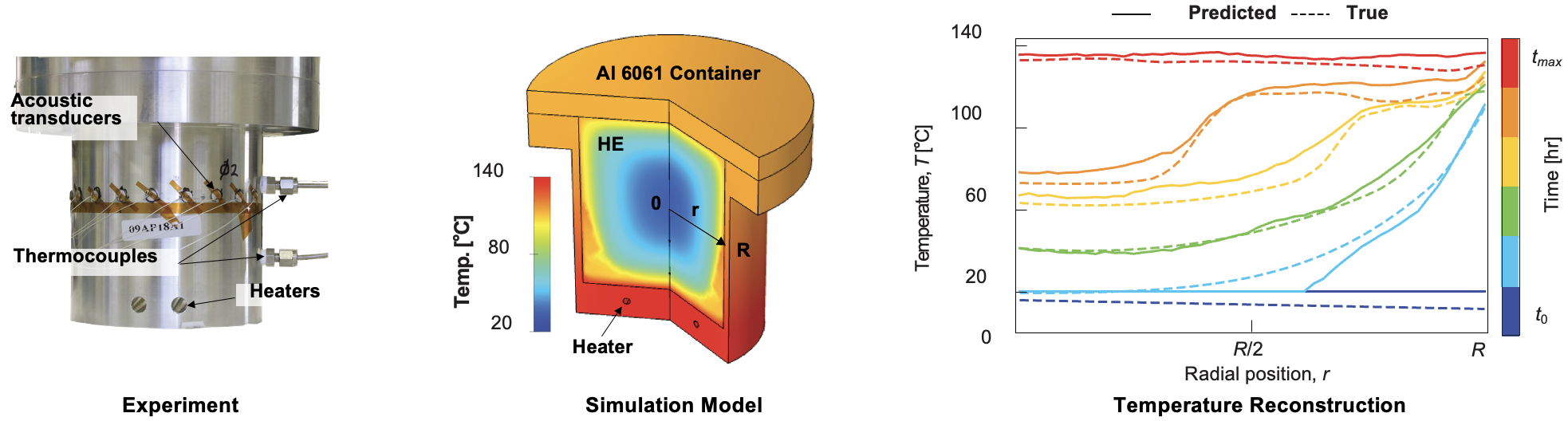}}}
	\caption{Illustration of (a) ultrasonic imaging to identify high-contrast defects in multi-layered bonded structures and (b) Noninvasive acoustic temperature imaging. In (a), (Left Panel): Experimental setup consists of a CTS-PA322T portable ultrasonic phased array flaw detector, a linear phased array and a specimen; (Central Panel): Configuration of the specimen with two notches in the top layer and two disbonds in the 1 mm-thick adhesive layer; (Right Panel): FCN-based reconstruction (Colour bar shows the magnitude of velocity (m/s)). In (b), hybrid data collection with (Left Panel) experimental acoustic measurements and (Central Panel) simulated temperature profiles were used to train a CNN to compute the temperature profile within a container filled with high explosives (HE). In (b), (Right Panel) the predicted and true radial temperature profiles at several different times as the HE is heated from ambient (blue) to detonation (red).}
	\label{fig:Acoustic_Imaging}
\end{figure*}

While numerous studies in ML-based NDT have focused on defect detection, \cite{Quantitative-2023-Rao} develops a novel approach. Their study introduces a quantitative ultrasonic imaging technique using a fully-supervised network to identify high-contrast defects in multi-layered bonded structures. This technique employs the fully-connected network (FCN) to convert ultrasonic data into longitudinal wave velocity models, which are then used to reconstruct defects in new data sets. In their experimental setup, as shown in the left panel of Fig.~\ref{fig:Acoustic_Imaging}(a), a CTS-PA322T portable ultrasonic phased array flaw detector was utilized alongside a linear phased array. The phased array had a central frequency of 5 MHz and a pitch of 0.6 mm. The specimen design is detailed in the central panel of Fig.~\ref{fig:Acoustic_Imaging}(a). The right panel of the same figure displays the reconstruction of this specimen using experimental data. The results demonstrate a high degree of accuracy in identifying the locations and shapes of two notches and disbonds. The reconstructed quantitative information closely matches the known specimen configuration, with artifacts being nearly undetectable.

\subsubsection{Acoustic temperature imaging in high explosives}

Acoustic waves are an attractive choice for noninvasive measurement of internal temperature for a range of applications such as monitoring chemical reactions or metal/plastic casting operations, assessing thermal energy storage devices, identifying abnormal heating in batteries, or determining if high explosives (HE) are safe to handle. 

\cite{Greenhall_arxiv_2023} develops a fully-supervised, simulation/measurement-driven temperature imaging technique that measures temperature distribution directly from the acoustic measurements using a CNN and a hybrid experiment/simulation training process illustrated in Fig.~\ref{fig:Acoustic_Imaging}(b). The temperature imaging technique was demonstrated on cylindrical containers filled with HE (pentolite 50/50) as they were heated externally from ambient until ignition. During heating, 16 acoustic transducers around the container circumference transmitted/received (Tx/Rx) acoustic bursts centered at 350 kHz and two thermocouples measured temperatures at the wall and center to collect temperatures for training data (Fig.~\ref{fig:Acoustic_Imaging}(b, Left Panel)). 

To acquire training data temperature profiles within the container without excessive thermocouples disrupting the heating, the HE melting process was simulated numerically using an axisymmetric finite element model (COMSOL) and a radial temperature profile $T(r)$ was measured (Fig.~\ref{fig:Acoustic_Imaging}(b, Central Panel)). To account for modeling errors, the simulated temperatures were corrected to agree with experimental thermocouple measurements. Training/testing samples comprised of data: acoustic waveforms $X$ measured from a single transmitter, by $N_{Rx}$ receivers and a label: a radial temperature profile $T$. The samples $(X,T)$ were then used to train/test a CNN (see \cite{Greenhall_arxiv_2023} for details).

Data were acquired from three nominally identical experiments/simulations and used to train and test the temperature imaging technique. Fig.~\ref{fig:Acoustic_Imaging}(b, Right Panel) shows the predicted (dashed lines) and simulated ``true'' radial temperature profiles (solid lines) at various times as the cylinder was heated from approximately 20°C at $t_0$ (dark blue) until ignition temperature of approximately 138°C was reached at $t_{max}\approx3.5$ hr (red). Good agreement was observed between the predicted and true temperatures. From repeated training/testing (10 times), they measured a mean error of 14°C, which is approximately 12\% of the range of temperatures throughout the experiments.

The temperature imaging technique based on a CNN and hybrid data provides considerable benefits over traditional temperature measurement techniques. CNN temperature imaging is noninvasive and does not require penetrating the container, unlike thermocouples and thermistors;\cite{Mcgee_temperature_1988} provides an internal temperature profile, unlike infrared thermography, which is limited to surface temperatures;\cite{gaussorgues_infrared_1993} works for multi-phase, attenuating materials, unlike existing physics-based acoustic temperature imaging techniques, which require mounting transducers on the container interior and/or are limited to single-phase materials with low acoustic attenuation\cite{Lu_MST_2000,  Kudo_AP_2003, Mizutani_JJAP_2006, Modlinski_Energy_2015, Zhang_ATE_2015,yu_temperature_2022}.

\subsection{Medical Ultrasound Computed Tomography}

USCT is an imaging method that uses tomographic principles for quantitative assessment of acoustic properties like sound speed, density, and acoustic attenuation~\cite{schreiman1984ultrasound, pratt2007sound,  duric2007detection, li2009vivo, wang2015waveform}. It has become widely adopted, especially in breast imaging applications~\cite{greenleaf1977quantitative, carson1981breast, schreiman1984ultrasound, andre1997high, duric2007detection,nam2013quantitative,  sandhu2015frequency, malik2016objective, duric2018breast, taskin20203d, duric2020using, wiskin2020full, javaherian2020refraction}, which involves submersion of the patient's breast in water, surrounded by ultrasound transducers that emit and receive acoustic pulses, allowing full breast insonification.

USCT offers several advantages over traditional breast imaging methods like mammography, such as higher sensitivity to invasive cancer, cost-effectiveness, and being non-radiative and compression-free~\cite{ruiter20123d, duric2013breast}. Recently approved by the U.S. FDA for breast cancer screening in women with dense breasts and as a general diagnostic tool, USCT still faces significant algorithmic challenges. Key among these is improving image quality and reducing the lengthy image reconstruction times, which currently require high-performance, possibly GPU-accelerated computers~\cite{zhang12, lucka21}. This computational demand not only limits USCT's clinical deployment but also raises costs, hindering its adoption in resource-limited settings. To overcome these challenges, innovative ML methods are being explored. \cite{fan2022model} proposes a primal-dual gradient descent method for efficient data mapping, \cite{stanziola2023learned} employs loop unrolling for wave equation modeling, and \cite{liu2021ultrasound} utilizes physics-informed neural networks to determine sound speed.

\subsubsection{Ultrasound Tomography for Breast Cancer Detection using ring-array systems}

In this section, we briefly describe three learning approaches tailored to ring-array USCT breast imaging. This practical system employs a circular ring array of elevation-focused ultrasonic transducers to achieve volumetric imaging by translating the array orthogonally to the imaging plane. Image reconstruction utilizes a slice-by-slice~(SBS) method, reconstructing the three-dimensional~(3D) volume by stacking two-dimensional~(2D) images for each ring array position.

The cited works~\cite{Li20213Dstochastic, donaldson21, Jeong2023deep,jeong2023investigating} explore a fully supervised simulation/measurement-driven approach, where a post-processing network learns to map preliminary velocity model estimates to high-resolution versions. These studies are distinguished by their use of anatomically and physiologically realistic numerical breast phantoms to train and assess the methods~\cite{Li20213Dstochastic} (refer to Appendix~\ref{sec:Appendix-E} for more details). Two methods~\cite{Li20213Dstochastic, donaldson21} use an early-stopped USCT-derived velocity model as input for a U-net architecture, producing visually enhanced yet potentially inaccurate models. Conversely, \cite{Jeong2023deep,jeong2023investigating} show that using diverse input maps like travel time estimates and reflectivity maps leads to more accurate and reliable velocity models, although some fine structures may be erroneously introduced, as the authors acknowledge \cite{9424044}. \cite{Poudel2019compensation} and \cite{li2024learning} demonstrate a fully supervised approach that addresses image reconstruction artifacts and inaccuracies arising from the mismatch between 3D wave propagation in the physical system and the simpler 2D imaging model often employed in breast USCT imaging.

\begin{figure}[t!]
\centering
\includegraphics[width=.5\textwidth]{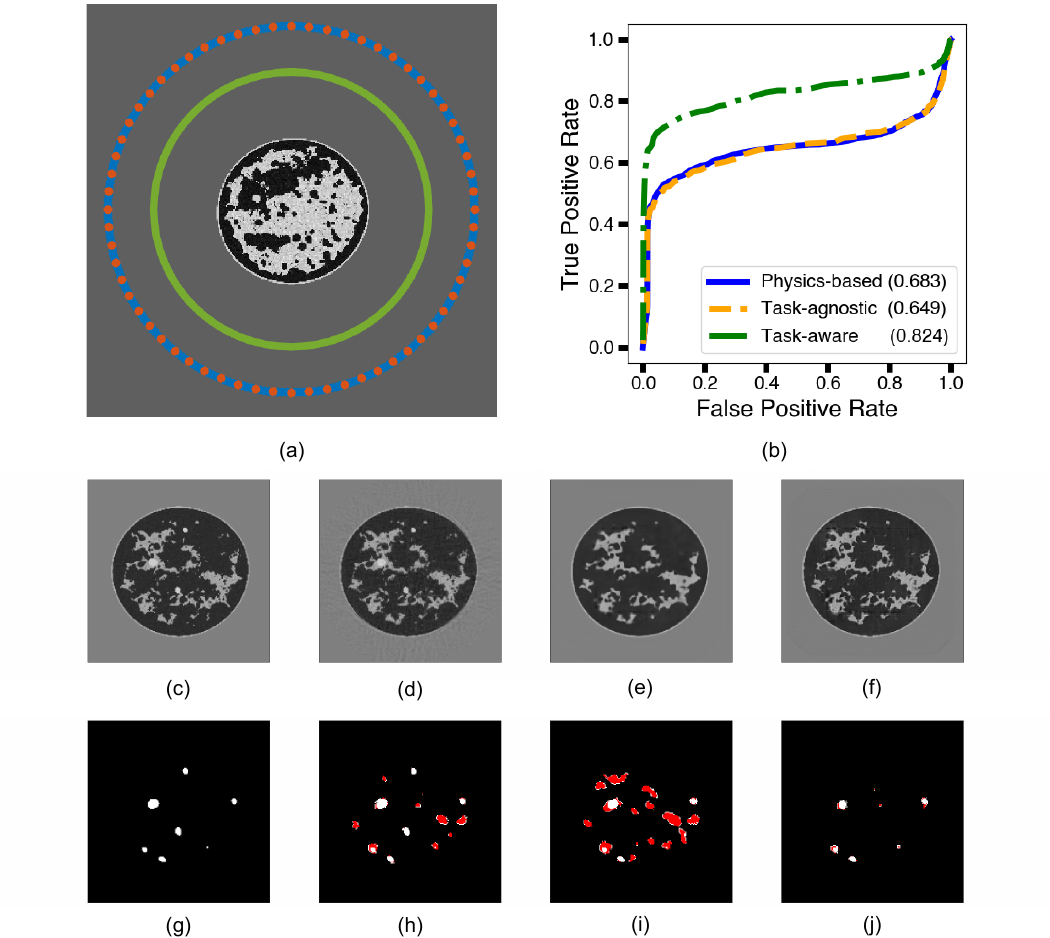}
\caption{(a) Geometry of the idealized 2D USCT imaging system in \cite{Learned-2023-Lozenski} with blue dots indicating emitter locations, red dots receivers, and green circle the field of view. (b) Receiver operating curve (ROC) for a tumor detection and localization task; area under the curve (AUC) values reported in the legend. The true speed of sound map (c) and estimates obtained by physics-based USCT (d), task-agnostic learned USCT (e), and the task-informed learned USCT method (f) in \cite{Learned-2023-Lozenski}. Tumor segmentation maps of the object (g), and segmentation maps estimated from reconstructed images with the three methods (h,i,j). True detection/localization is indicated in white and false detection in red. Speed of sound images produced by the task-informed method have similar visual or superior appearance (second rows) and provide more reliable lesion localization and detection compared to that achieved with physics-based or task-agnostic learned USCT. In particular,  The ROC curve plot shows that the task-informed method can achieve over 20\% higher AUC than physics-based USCT or other task-agnostic learned methods.}
\label{fig:task-based-learning}
\end{figure}

Here we highlight the work of \cite{Learned-2023-Lozenski} on enhancing USCT~(as shown in Fig.~\ref{fig:task-based-learning}). They address the training and assessment of learned USCT methods using task-based measurements of image quality. In this work, a fully supervised physics-aware approach is proposed where a CNN mapping measurement data to the corresponding velocity is trained to minimize a specifically designed loss function that incorporates information specific to the clinical task for which the image was taken. To incorporate task-specific information, a weighted sum of the mean square error loss (MSE) in the image domain and a novel task-informed objective was proposed based on features extracted by a CNN-based numerical observer that was pre-trained on a tumor detection-localization task using the reference velocity models and tumor segmentation maps. Computer simulation studies using an idealized 2D USCT imaging system demonstrate that the proposed learning method can achieve a similar visual appearance to the velocity model estimated by USCT but also provide more reliable detection and localization of tumors. 

\section{Lessons Learned}

Traditional physics-based CWI methods face three challenges: (1) non-unique solutions, (2) multiple local minima, and (3) high computational demands. In this section, we aim to provide lessons learned and analyses concerning these crucial issues.

\subsection{Ill-posedness and Non-uniqueness Solution}

The challenge of ill-posedness, primarily caused by incomplete acquisition coverage and the vast dimensionality of the solution space, is a fundamental aspect of CWI problems. Traditional methods, such as regularization or preconditioning, can be effective, but usually require careful design of the penalty function to encode the user's preferences for which of the non-unique solutions to the unregularized problem to favor. Thus, selecting and optimizing the right regularization function to balance data fidelity and regularization terms, as in Eq.~(\ref{eq:MisFit}), can be challenging. In contrast, emerging ML-based methods offer alternative strategies to mitigate ill-posedness. Among the opportunities enabled by ML-based methods, we highlight the following approaches.

\begin{itemize}
    \item \textbf{Learning the Null-Space Component of the Image from Labeled Data.~}To address the challenge of non-uniqueness in CWI solutions, fully supervised models trained on carefully curated datasets are effective. A typical example of this is InversionNet~\cite{wu-2019-inversionnet}, which uses training datasets constructed in a pairwise way to ensure a direct one-to-one correspondence between velocity maps and waveform data. This strategy confines the model's predictive capabilities to the data distribution of the training set, thereby alleviating the ill-posedness of CWI problems. However, a potential drawback is the model’s tendency to overfit to its training dataset, which can limit its generalization ability. These issues, along with computational cost and generalization capability, will be further discussed in subsequent sections. 
    
    \item \textbf{Learning a Low-Dimensional Embedding of the Model Parameter Space.~}An alternative strategy for reducing the null space involves transitioning the learning process to a lower-dimensional space. \cite{Feng-2022-Intriguing} provides a compelling illustration of this approach. Instead of performing inversion within the original, high-dimensional solution space, their method employs a sophisticated encoder-decoder framework to navigate a significantly reduced solution space (\ie, the latent space). This projection technique not only mitigates the inherent ill-posedness associated with CWI but also uncovers intriguing relationships between data and labels, such as an unexpected near-linearity~\cite{Feng-2022-Intriguing, AutoLinear-2024-Feng}.
        
    \item \textbf{Plug and Play Priors.~}This particular regularization approach, rooted in the traditional CWI framework, is elaborated upon in the section on \textbf{Plug and Play Methods}. Its key distinction lies in its ability to derive the regularization term directly from data, moving away from the conventional practice of employing a predefined analytical regularization term. This data-driven approach substantially amplifies the efficacy of the regularization, offering a notable improvement and flexibility in addressing the challenges inherent in CWI.

    \item \textbf{Implicit Regularization through Neural Network Parameterization.~}As discussed in the earlier section on \textbf{Parameterization Solution Methods}, the choice of a neural network for parameterizing solutions has a significant impact on the structure and exploration of the solution space. Following a similar idea, in the context of CWI, \cite{Integrating-2021-Zhu} finds that reparameterizing the velocity map using neural networks implicitly introduces a form of regularization. This regularization can improve CWI's effectiveness by shaping the solution space in a way that is informed by the neural network's architecture, leading to more robust imaging results.
    
\end{itemize}

\subsection{Nonlinearity and Nonconvexity: Conventional vs. Data-driven CWI}

Computational wave imaging is inherently nonlinear and nonconvex due to the physics of wave propagation and the limited information available in measured data. Nonlinearity arises from the sensitivity of wavefields to changes in material properties, particularly in heterogeneous or high-contrast media. Nonconvexity is further exacerbated by incomplete frequency content, which introduces ambiguity into the inversion and increases the likelihood of convergence to suboptimal solutions.

These challenges manifest differently across application domains. In geophysical FWI, the absence of low-frequency components is a primary limitation. These frequencies are essential for recovering large-scale background structures and for guiding the inversion toward global minima. Their absence results in a phenomenon known as \textit{stratigraphic filtering}, where the recorded high-frequency wavefields predominantly reflect small-scale structures, suppressing information about the smooth, large-scale features. This effect contributes significantly to the cycle-skipping problem. In contrast, in USCT and NDT, the primary difficulties stem from strong acoustic contrasts and the need for high spatial resolution. Here, the lack of high-frequency data limits the ability to resolve fine structural details. In both settings, incomplete spectral information reduces gradient accuracy and increases instability in optimization.

These issues make optimization difficult for conventional CWI methods, such as the one shown in Eq.~(\ref{eq:MisFit}), which rely on first-order local optimization techniques like accelerated gradient or nonlinear conjugate gradient methods. Such methods often struggle to escape local minima and are particularly vulnerable to the cycle-skipping phenomenon, especially when reconstructing smooth, long-wavelength features. To address this, several strategies have been developed. Frequency continuation \cite{pratt1999seismic} is a multiscale approach that leverages the low-frequency information content in the waveform data to build an initial guess of the sought-after material properties and then sequentially refine them using higher-frequency information. Adaptive FWI~\cite{WarnerGuasch16} defines a new data fidelity term that, by introducing a learnable convolutional filter mapping the predicted data into the measured data, can overcome the cycle-skipping phenomenon. Optimal transport-based FWI \cite{yang2018application} defines a new data fidelity term based on the quadratic Wasserstein metric to correctly correct for cycle-skipping. 

Despite these advances, most ``Data Consistent'' learning-based methods (see Fig.~\ref{fig:trend_breakdown}) continue to adopt standard $\ell_2$-based losses and remain susceptible to similar limitations, particularly in the presence of missing low-frequency content. Some recent studies (\eg,\cite{FangshuJianwei23}) have explored alternative loss formulations or implicit regularization schemes, but such methods remain relatively limited. More recently, a set of emerging methods~\cite{Self-2024-Cheng, Progressive-2021-Hu, Efficient-2021-Jin, Extrapolated-2020-Sun} has proposed learning-based strategies to reconstruct missing low-frequency components directly from high-frequency measurements. This approach uses a neural network to synthesize low-frequency wavefields, which are then used to augment the inversion process. By recovering this missing information, the method helps reduce cycle-skipping and improve convergence in bandwidth-limited scenarios. Such strategies highlight promising directions for integrating physics-informed modeling with data-driven learning to address fundamental limitations in practical imaging systems.



\renewcommand{\arraystretch}{1}

\begin{table*}
\centering
\setlength{\extrarowheight}{0pt}
\addtolength{\extrarowheight}{\aboverulesep}
\addtolength{\extrarowheight}{\belowrulesep}
\setlength{\aboverulesep}{0pt}
\setlength{\belowrulesep}{0pt}
\arrayrulecolor{black}
\caption{Computational and memory cost comparison of Data-driven Inversion Methods. $\dagger$: For Augmentation method~\cite{Renan-2022-Physics}, we amortize the cost of pretraining InversionNet on the non-augmented dataset. Subsequent fine-tuning costs ($\delta$) vary with the augmented dataset size. }
\begin{adjustbox}{width=1\textwidth}
\begin{tabular}{|c|c|c|c|c|c|}
\hline
\rowcolor[rgb]{0.753,0.753,0.753} \multicolumn{2}{|c|}{{\cellcolor[rgb]{0.753,0.753,0.753}}}                                         & \multicolumn{2}{c|}{\textbf{Computational Cost}}                                                                                   & \multicolumn{2}{c|}{\textbf{Memory Cost }}                                                                                                                                                                                                                                                                                            \\ 
\hhline{{\arrayrulecolor{black}}|>{\arrayrulecolor[rgb]{0.753,0.753,0.753}}-->{\arrayrulecolor{black}} |>{\arrayrulecolor{black}}---->{\arrayrulecolor{black}}|}
\rowcolor[rgb]{0.753,0.753,0.753} \multicolumn{2}{|c|}{\multirow{-2}{*}{{\cellcolor[rgb]{0.753,0.753,0.753}}\textbf{Techniques }}}   & \textbf{FLOPs\,(M)}                                                                            & \textbf{Time\,(hr)} & \multicolumn{1}{c|}{\begin{tabular}[c]{@{}>{\cellcolor[rgb]{0.753,0.753,0.753}}c@{}}\textbf{Parameter Number\,(M)}\end{tabular}} & \multicolumn{1}{c|}{\begin{tabular}[c]{@{}>{\cellcolor[rgb]{0.753,0.753,0.753}}c@{}}\textbf{Max Memory\,(G)}\end{tabular}}                \\ 
\hhline{|------|}
\multirow{6}{*}{\begin{sideways}{\begin{tabular}[c]{@{}c@{}}Data-driven Inversion\\ Methods\end{tabular}}\end{sideways}}                                                     & {InversionNet~\cite{wu-2019-inversionnet}}  &                                         {$2,908$}     &                                                                                               {4}                                   &     {24.5}                                                &  {11.4}                                                                                                                                 \\ 
\cline{2-6}         & {Augmentation~\cite{Renan-2022-Physics}}      &      {$(2,908+\delta)^{\dagger}$ }                                   &   {$(4+\delta)^{\dagger}$}                                                                                                                              &       {$24.5$}                                                                                                                            &           {$(11.4+\delta)^{\dagger}$ }                                                                                                                            \\ 
\cline{2-6} & {Auto-Linear~\cite{AutoLinear-2024-Feng}}&                    {  305.2}                                &                                         {8}                                                                                        &                                                      {12.2}                                                                            &           {2.2}                                                                                                                         \\ 
\cline{2-6} & {UPFWI~\cite{Jin-2021-Unsupervised}}&                     { 3,229}                             &                                                                                             {60}                                   &     {24.5}                                                                                                                              &             {27.9}                                                                                                                       \\  
\cline{2-6} & Fourier-DeepONet~\cite{Zhu-2023-DeepONet}  &    {3,664} &    {15.3}                                                                                                                         &    {15.5}                                                                                                                              &     {29.2}                                                                                                                               \\
\cline{2-6}
   & Invertible Network~\cite{Solving-2023-Gupta} &  4,180   &                                                              33.8                                                              &        26.1                                                                                                                          &      37.9                                                                                               \\
\hhline{|------|}

\end{tabular}
\end{adjustbox}

\label{tab:Cost_NeuralNet}
\end{table*}

\renewcommand{\arraystretch}{1}

\begin{table*}
\centering
\arrayrulecolor{black}
\caption{Computational and memory cost comparison of a conventional physics method~\cite{Bunks-1995-Multiscale} and InversionNet~\cite{wu-2019-inversionnet}. This comparison is based on varying the training-to-test ratio by incrementally enlarging the training set while keeping the test set size constant as one sample. Here the computational cost for InversionNet includes both the training and inference in total, while the physics method only involves inference.}
\begin{adjustbox}{width=1\textwidth}
\begin{tabular}{|c|c|c|c|c|} 
\hline
\rowcolor[rgb]{0.753,0.753,0.753} {\cellcolor[rgb]{0.753,0.753,0.753}}                                                & \multicolumn{2}{c|}{\textbf{Computational Cost}}             & \multicolumn{2}{c|}{\textbf{Memory Cost }}                    \\ 
\hhline{|>{\arrayrulecolor[rgb]{0.753,0.753,0.753}}->{\arrayrulecolor{black}}|>{\arrayrulecolor{black}}----|}
\rowcolor[rgb]{0.753,0.753,0.753} \multirow{-2}{*}{{\cellcolor[rgb]{0.753,0.753,0.753}}\textbf{Training/Test Ratio }} & \textbf{Conventional Physics Method} & \textbf{InversionNet} & \textbf{Conventional Physics Method} & \textbf{InversionNet}  \\ 
\hline
10                                                                                                                    & \multirow{4}{*}{113~s}              & 0.3~s                 & \multirow{4}{*}{0.4~G}                    &          5.2~G                \\ 
\cline{1-1}\cline{3-3}\cline{5-5}
100                                                                                                                   &                                      & 3~s                   &                                      &              12.1~G             \\ 
\cline{1-1}\cline{3-3}\cline{5-5}
1,000                                                                                                                  &                                      & 30~s                  &                                      &             80.3~G             \\ 
\cline{1-1}\cline{3-3}\cline{5-5}
10,000                                                                                                                 &                                      & 300~s                 &                                      &             762.4~G            \\
\hline
\end{tabular}
\end{adjustbox}

\label{tab:Cost_Physics}
\end{table*}

\subsection{Computational Cost Analysis}

Computational efficiency plays a critical role in the practical adoption of CWI methods. While algorithmic accuracy and generalization are essential, their real-world utility is ultimately constrained by available computing resources, problem scale, and application requirements. In this section, we outline key factors that influence computational costs and compare representative methods across a common benchmark to provide actionable insight.

\begin{itemize}
    \item \textbf{Training vs. Inference Trade-off.~}ML-based CWI methods generally separate computation into an intensive \textit{training} phase and a typically lightweight \textit{inference} phase. Once trained, these models can produce reconstructions rapidly, making them attractive for high-throughput or real-time applications. In contrast, physics-based methods, such as traditional adjoint-state methods, do not involve training but incur substantial cost during \textit{inference}, as each reconstruction requires iterative PDE solves. Consequently, ML approaches are often advantageous in monitoring settings (\eg, CO$_2$ storage or seismic hazard warning), where a single trained model supports repeated inferences across time or space.

    \item \textbf{Problem Size and Application Demands.~}The feasibility of deploying a given method also depends on the scale and fidelity required by the application. For example, large-scale 3D seismic inversion demands high-resolution imaging over complex domains, often requiring terabyte-scale memory and extended runtime. Conversely, safety-critical systems such as tsunami warning or industrial inspection prioritize low-latency response, imposing strict time and compute budgets. While ML models offer efficient inference, they may struggle with generalization in out-of-distribution settings. Physics-based solvers offer more interpretability and robustness but are often prohibitively slow for time-sensitive tasks.

    \item \textbf{Hardware and Software Considerations.~}Modern ML models are typically optimized for GPUs and can benefit from data-parallel processing on multi-GPU systems or HPC clusters. However, inference performance on CPU-only environments or edge devices can degrade significantly. Physics-based solvers, while more general in terms of platform compatibility, scale poorly with increasing problem complexity unless deployed on distributed-memory systems. Emerging architectures (\eg, TPUs or hybrid CPU-GPU clusters) and frameworks (\eg, JAX, TensorFlow, PyTorch) continue to shape how different methods perform across platforms.

\end{itemize}
To concretely compare computational profiles, we benchmarked several representative models on the OpenFWI Fault Family dataset~\cite{OpenFWI-2022-Deng}. Each model transforms input waveform data of size $5 \times 1,000 \times 70$ into a $70 \times 70$ velocity map. Performance metrics include floating point operations (FLOPs), training and inference runtime, memory usage, and parameter counts. Measurements were collected using the \texttt{ptflops} package on an NVIDIA Tesla V100 GPU, with all models trained over 120 epochs using mini-batch parallelism.

Table~\ref{tab:Cost_NeuralNet} summarizes results for ML-based approaches. InversionNet achieves rapid inference but exhibits increasing training cost as dataset size grows. Physics-aware methods like UPFWI~\cite{Jin-2021-Unsupervised} and invertible networks~\cite{Solving-2023-Gupta} incur additional computational overhead due to repeated forward modeling. For comparison, Table~\ref{tab:Cost_Physics} reports the computational demands of a time-domain FWI solver~\cite{Bunks-1995-Multiscale}, highlighting its high per-inference cost even with multiscale acceleration strategies.

While ML-based methods can reduce per-inference cost by orders of magnitude, they often require large volumes of high-quality training data and may not generalize beyond synthetic domains without careful domain adaptation. Physics-based methods, although more robust to unseen configurations, are less scalable in large or real-time scenarios. Hybrid frameworks, such as physics-informed networks and plug-and-play priors, offer promising compromises by combining inductive biases from wave physics with learning-based flexibility.


\section{Limitations, Challenges, and Outlook}

We summarize the aforementioned methods in Table~\ref{tab:summary}, outlining their advantages, limitations, and optimal use cases. 
As research in this field progresses, new challenges emerge. Below, we identify key challenges and suggest directions for future work that could significantly advance CWI.

\textbf{Challenges in Generalization of Learning-Based Methods.~}A fundamental impediment to the deployment of learning-based CWI methods lies in their limited generalization capacity across heterogeneous acquisition conditions, physical media, and experimental setups. This challenge originates from the domain shift between the training environments—typically synthetic and idealized—and real-world application scenarios, which are far more complex and variable.

Unlike in computer vision or natural language processing, where large-scale datasets are acquired from naturally occurring sources, training datasets in CWI are almost exclusively generated via numerical forward modeling on procedurally designed velocity models. These datasets, although grounded in physical principles, often fail to capture the full complexity of measurement noise, structural heterogeneity, and wavefield dynamics observed in practice. To understand the origins of this generalization breakdown, we decompose the problem into three interdependent components:

\paragraph{Input Discrepancy -- Measurement Domain Gap} 
Wavefield measurements used for training are generated through forward simulations under idealized conditions—complete acquisition coverage, deterministic source functions, and noise-free environments. In contrast, empirical measurements are corrupted by aleatoric noise, temporal jitter, sensor-specific distortions, and limited aperture effects. These discrepancies alter the spectral and statistical structure of the data, resulting in covariate shift at inference time and degraded model performance.

\paragraph{Output Discrepancy -- Velocity Model Prior Gap} 
Ground-truth velocity maps in supervised training regimes are typically drawn from procedurally defined priors—\eg, Gaussian random fields, rule-based geological models, or anatomical templates—that provide limited coverage of real-world variability. Such priors often lack high-order spatial statistics, anisotropy, or multiscale heterogeneity that characterize actual subsurface or biological structures. Consequently, the trained inverse operator becomes biased toward synthetic regularities and may fail to recover critical structural features in out-of-distribution domains.

\paragraph{Forward Model Discrepancy -- Simulator Fidelity Gap} 
A third and subtler source of generalization error stems from discrepancies between the physical processes simulated during training and those present during deployment. Training-time simulations often assume simplified physics—such as constant density acoustics, perfect boundary conditions, or linear propagation. These assumptions may not hold in real-world scenarios involving viscoelastic attenuation, scattering, mode conversion, or nonlinear effects. As a result, the learned mappings do not generalize well when faced with data governed by richer, unmodeled dynamics.

Together, these three sources form a framework for understanding generalization limits in CWI. They emphasize that robustness cannot be achieved by improving data priors or network architectures alone: the fidelity of the simulated environment and its alignment with real-world physics is equally critical.

Looking forward, advancing the generalization and applicability of learning-based CWI methods will require the development of tightly coupled frameworks that integrate data-driven models with high-fidelity physics solvers. Recent progress in differentiable programming and operator learning presents an opportunity to embed wave physics more deeply into learning pipelines, enabling inversion models to account for complex propagation phenomena such as anisotropy, viscoelasticity, and mode conversion. Parallel to this, semi-supervised and self-supervised learning strategies offer a path toward harnessing large volumes of unlabeled field data, thereby improving model robustness while reducing reliance on synthetic priors. Another important direction lies in the co-design of acquisition and inference---jointly optimizing transducer placement, source design, and learning objectives to improve imaging resolution under practical constraints. Finally, there is a growing need to incorporate principled uncertainty quantification into learning-based frameworks to assess model confidence and guide decision-making in safety-critical applications such as medical imaging and subsurface monitoring.

 
\textbf{Model Interpretability.~}The topic of model interpretability is of particular significance for deep neural networks known for their ``black box'' nature. In CWI, understanding the rationale behind a model's predictions can be as helpful as the predictions themselves. Interpretability enables us to grasp why certain predictions are made, making it an invaluable tool for debugging, improving model performance, and identifying biases. While tools like UQ, as mentioned earlier, can aid in interpreting deep models, extensive research on this aspect has been conducted in the broader ML/AI community~\cite{Interpretability-2017-Chakraborty, Interpretable-2021-Li}. Surprisingly, the CWI field has seen relatively little exploration in this area, pointing to a promising direction for future research and the potential for significant advancements in CWI through enhanced interpretability.

\textbf{Imaging Quality Assessment.~}Implementing effective quality control plays a crucial role in evaluating successful applications of CWI. Ultimately, image quality should be assessed based on the utility of the image to perform a specific task. For example, in the context of carbon sequestration, CWI may allow the detection of an underground gas plume; in medical USCT, CWI can support breast cancer screening and diagnostics tasks; and, in industrial ultrasonic CWI, can allow the non-destructive detection of defects. However, a growing body of evidence is showing that, while ML-based approaches can improve conventional metrics of image accuracy (such as mean square error, MSE; mean absolute error, MAE; or structural similarity, SSIM), they can result in diminished task performance\cite{zhang21, li21,LiVillaLiEtAl24}. In fact, these metrics do not always  correlate with the usefulness of a reconstructed image for performing a task of interest ~\cite{barrett93, christianson15}.  Due to this discrepancy, there is a growing interest in assessing the task performance of proposed image reconstruction methods by use of statistical signal detection theory and numerical observer methods \cite{zhang21, adler22, he2013model,Learned-2023-Lozenski}. 

\begin{table*}
\large
\centering
\setlength{\extrarowheight}{0pt}
\addtolength{\extrarowheight}{\aboverulesep}
\addtolength{\extrarowheight}{\belowrulesep}
\setlength{\aboverulesep}{0pt}
\setlength{\belowrulesep}{0pt}
\arrayrulecolor{black}
\caption{A Summary of CWI Methods. Those aforementioned methods are categorized into data-driven inversion methods and physics-based methods. Their respective advantages, limitations, and preferable applications are provided in the table.}
\begin{adjustbox}{width=1\textwidth}
\begin{tabular}{|c|c|p{9cm}|p{9cm}|p{9cm}|} 
\hline
\rowcolor[rgb]{0.753,0.753,0.753} \multicolumn{2}{|c|}{\multirow{-1}{*}{{\cellcolor[rgb]{0.753,0.753,0.753}}\textbf{Techniques }}}    & \multicolumn{1}{c|}{\multirow{-1}{*}{\textbf{Advantage }}}                        & \multicolumn{1}{c|}{\multirow{-1}{*}{{\cellcolor[rgb]{0.753,0.753,0.753}}\textbf{Limitations }}}                               & \multicolumn{1}{c|}{\multirow{-1}{*}{{\cellcolor[rgb]{0.753,0.753,0.753}}\textbf{Preferable Applications }}}                \\ 
\hhline{|-----|}
\multirow{35}{*}{\rotatebox{90}{Data-driven Inversion Methods}}                                                      & \multirow{5.5}{*}{{\begin{sideways}InversionNet~\cite{wu-2019-inversionnet}\end{sideways}}}   & \begin{itemize} 
  
    \item Directly extract physical principles from labeled data.
    \item Achieve high accuracy in estimating in-distribution data.
\end{itemize}  & \begin{itemize} 
    \item Depends significantly on the availability of ample labeled data.
    \item Limited ability to generalize to out-of-distribution data.
\end{itemize} & \begin{itemize} \item Optimal for problems with abundant pre-labeled data and target data similar to the training set, like in monitoring tasks. \end{itemize} \\ [15ex]
\cline{2-5}
          & \multirow{5.5}{*}{\begin{sideways}Augmentation~\cite{Renan-2022-Physics}\end{sideways}}      & \begin{itemize} 
    \item Decrease the model's dependence on labels.
    \item Enhance the training set's representativeness by integrating underlying physics and incorporating unlabeled data.
\end{itemize}  & \begin{itemize} 
    \item Increases computational costs for model fine-tuning.
\end{itemize} & \begin{itemize} \item Ideal for tasks requiring a mix of pre-labeled and additional data, like unlabeled data, exemplified by multi-physics imaging. \end{itemize}\\ [15ex]
\cline{2-5} & \multirow{5.5}{*}{\begin{sideways}Auto-Linear~\cite{AutoLinear-2024-Feng}\end{sideways}}& \begin{itemize} 
    \item Train the encoder and decoder independently using self-supervised learning.
    \item Boost model efficacy by utilizing unpaired data and labels in various applications.
\end{itemize}  & \begin{itemize} 
    \item More samples would be needed to train the model.
    \item Additional computational cost due to pre-training of the encoder and decoder.
\end{itemize} & \begin{itemize} \item Adaptable to imaging tasks under diverse scenarios, including variable testing conditions.\end{itemize}\\ [14ex]
\cline{2-5} & \multirow{5.5}{*}{\begin{sideways}UPFWI~\cite{Jin-2021-Unsupervised}\end{sideways}}& \begin{itemize} 
    \item Eliminates the need for label information. 
    \item Enhanced generalization capability through the integration of physics.
\end{itemize}  & \begin{itemize} 
    \item Requires more samples than supervised learning.
    \item Increased computational costs due to extensive forward modeling.
\end{itemize} & \begin{itemize} \item Ideal for label-free problems with ample waveform data, such as Distributed Acoustic Sensing (DAS) imaging.\end{itemize}\\ [14ex] 
\cline{2-5} & \multirow{5.5}{*}{\begin{sideways}\begin{tabular}[c]{@{}c@{}}Fourier-\\DeepONet~\cite{Zhu-2023-DeepONet}\end{tabular}\end{sideways}}  & \begin{itemize} 
    \item Enhances the model's generalization capability for source and receiver locations. 
    \item Boosts the model's robustness regarding source frequency.
    \end{itemize}  & \begin{itemize} 
    \item Requires additional samples to encompass a broader range of source/receiver locations and source frequencies. 
    \item Results in higher computational costs.
\end{itemize} & \begin{itemize} \item Optimal for applications involving diverse or uncertain source/receiver characteristics, such as central frequency and locations. \end{itemize}\\[14ex]
\cline{2-5}
   & \multirow{5.5}{*}{\begin{sideways}\begin{tabular}[c]{@{}c@{}}Invertible\\Network~\cite{Solving-2023-Gupta}\end{tabular}\end{sideways}} & \begin{itemize} 
    \item Boosts inversion performance through the integration of forward modeling.
    \item Enables dual functionality for both inversion and forward modeling in the model.
    \end{itemize}  & \begin{itemize} 
    \item Significantly increased computational costs. 
    \item Necessitates a larger sample size.
\end{itemize} & \begin{itemize} \item Ideal for applications lacking forward modeling, such as blind inversion scenarios.\end{itemize} \\[14ex]
\hhline{|-----|}
\multirow{18}{*}{\begin{sideways}{Physics-Based Inversion}\end{sideways}}                                                                & \multirow{6}{*}{\begin{sideways}\begin{tabular}[c]{@{}c@{}}Conventional \\Physics-Method~\cite{Virieux-2009-Overview}\end{tabular}\end{sideways}} & \begin{itemize} 
    \item Upon optimization, the inversion reliably conforms to the measurements.
    \item Several computational techniques, such as advanced regularization and preconditioning, exist to alleviate the issue of ill-posedness.
    \end{itemize}  & \begin{itemize} 
    \item Computationally demanding when inverting multiple datasets.
    \item Results in high sensitivity to initial guesses and susceptibility to cycle-skipping.
\end{itemize} & \begin{itemize} \item Optimal for inverting baseline data during time-lapse monitoring.\end{itemize} \\[17ex]
\hhline{|~----|}
& \multirow{6}{*}{\begin{sideways}Parameterization~\cite{Integrating-2021-Zhu}\end{sideways}} & \begin{itemize} 
    \item Inherits similar advantages as those found in conventional physics methods. 
    \item Further mitigates the effects of local minima and is more robust against noise in data. 
    \end{itemize}  & \begin{itemize} 
    \item Requires retraining for new data adaptation. 
    \item Retains the limitations of physical inversion techniques, including sensitivity to initial guesses and cycle-skipping.
\end{itemize} & \begin{itemize} \item Ideal for scenarios lacking training samples.\end{itemize} \\[17ex]
\hhline{|~----|}
& \multirow{6}{*}{\begin{sideways}Plug and Play~\cite{Fu-2023-Efficient}\end{sideways}}   & \begin{itemize} 
    \item Analogous to the conventional physics methods.
    \item Offers flexibility in eliminating artifacts and noise during inversion.
    \end{itemize}  & \begin{itemize} 
    \item Comparable to the aforementioned approach. 
    \item Choosing the denoising operator is challenging and demands domain expertise.
\end{itemize} & \begin{itemize} \item Ideal for scenarios lacking training samples with known noise distribution.\end{itemize} \\[17ex]
\hhline{|-----|}
\end{tabular}
\end{adjustbox}
\label{tab:summary}
\end{table*}

\section{Acknowledgements}
Y. Lin acknowledges support from the University of North Carolina at Chapel Hill School of Data Science and Society through a faculty start-up grant and from the U.S. National Science Foundation (Award No. 2504439). Y. Lin and S. Feng also acknowledge support from the U.S. Department of Energy (DOE), including the Office of Science’s Advanced Scientific Computing Research (ASCR) program (Award No. DE-SC0025377) and the Office of Fossil Energy’s Carbon Storage Research Program, under the Science-Informed Machine Learning to Accelerate Real-Time Decision Making for Carbon Storage (SMART-CS) Initiative. 
U. Villa and M. Anastasio acknowledge support from the National Institute of Biomedical Imaging and Bioengineering (NIBIB) of the U.S. National Institutes of Health (NIH) under awards EB028652, EB031585, and EB034261. U. Villa would also like to acknowledge Dr. Luke Lozenski for assisting with the generation of Fig.~\ref{fig:task-based-learning} and Mr. Evan Scope Crafts for the insightful discussion and suggestion regarding the Generative AI Section. Computational resources were provided by the University of North Carolina at Chapel Hill Information Technology Services Research Computing.
\appendices

\section{Wave Equations}
\label{sec:Appendix-A}

This supplementary section provides detailed formulations for wave equations that extend beyond the acoustic models discussed in the main text. These include elastic, viscoacoustic, and anisotropic equations used to model more complex wave phenomena.

\subsection{Elastic Wave Equation}
Within a solid medium, as the material undergoes deformation caused by stress, this results in the transmission of both P-waves and secondary (S) waves. The elastic wave equation follows~\cite{levander1988fourth}:
\begin{eqnarray}
\rho\frac{\partial{{\mathbf{u} }}}{\partial{t}}-\nabla \cdot \underline{\underline{\boldsymbol{\sigma}}}&=&0 \,,  \nonumber\\
\frac{\partial \underline{\underline{\boldsymbol{\sigma}}}}{\partial t} - \mu \left( \nabla \mathbf{u} + (\nabla \mathbf{u})^T \right) - \lambda (\nabla \cdot \mathbf{u}) \mathbf{I} &=& \mathbf{S},
\label{Elastic}
\end{eqnarray}
where $\mathbf{u} \in \mathbb{R}^d$ is the particle velocity, $\underline{\underline{\boldsymbol{\sigma}}} \in \mathbb{R}^{d\times d}$ is the stress tensor, $\mathbf{I}$ is the identity matrix, $\lambda$ and $\mu$ are the Lamé parameters, and $\mathbf{S} \in \mathbb{R}^{d\times d}$ represents external sources. The P- and S-wave speeds are $V_P=\sqrt{(\lambda+2\mu)/\rho}$ and $V_S=\sqrt{\mu/\rho}$.

\subsection{Viscoacoustic Wave Equation}
The viscoacoustic wave equation models attenuation effects via a memory term~\cite{blanch1995efficient}:
\begin{eqnarray}
\frac{\partial{p}}{\partial{t}}+K\frac{\tau_\epsilon}{\tau_\sigma}\left(\nabla\cdot\mathbf{u}+r_{p}\right)&=&\mathbf{S} \,,  \nonumber\\
\frac{\partial{\mathbf{u}}}{\partial{t}}+\frac{1}{\rho}\nabla{p}&=&0 \,,  \nonumber\\
\frac{\partial{{r_p}}}{\partial{t}}+\frac{1}{\tau_\sigma}\left(r_p+K\left(\frac{\tau_\epsilon}{\tau_\sigma}-1\right)\left(\nabla\cdot\mathbf{u}\right)\right)&=&0 \,.
\label{Viscoacoustic}
\end{eqnarray}
Here, $K$ is the bulk modulus, $\tau_\epsilon$ and $\tau_\sigma$ are relaxation times, and $r_p$ is a memory variable.

\subsection{Anisotropic Wave Equation}
The pseudo-acoustic approximation for vertically transverse isotropic (VTI) media is expressed as~\cite{alkhalifah2000acoustic,duveneck2008acoustic}:
\begin{eqnarray}
\frac{\partial{u_x}}{\partial{t}}-\frac{1}{\rho}\frac{\partial{\sigma_{xx}}}{\partial{x}}&=&0 \,,  \nonumber\\
\frac{\partial{u_z}}{\partial{t}}-\frac{1}{\rho}\frac{\partial{\sigma_{zz}}}{\partial{z}}&=&0 \,, \nonumber\\
\frac{\partial{\sigma_{xx}}}{\partial{t}}-{\rho}v_p^2\left[(1+2\epsilon)\frac{\partial{u_x}}{\partial{x}}+\sqrt{1+2\delta}\frac{\partial{u_z}}{\partial{z}}\right]&=&S_{{xx}} \,,  \nonumber\\
\frac{\partial{\sigma_{zz}}}{\partial{t}}-{\rho}v_p^2\left[\sqrt{1+2\delta}\frac{\partial{u_x}}{\partial{x}}+\frac{\partial{u_z}}{\partial{z}}\right]&=&S_{{zz}} \,,
\label{Anisotropy}
\end{eqnarray}
where $u_x$, $u_z$ are particle velocities, $\sigma_{xx}$, $\sigma_{zz}$ are stress components, $\epsilon$, $\delta$ are Thomsen anisotropy parameters, and $v_p$ is vertical P-wave velocity.

\begin{figure*}[t]
    \centering
    \includegraphics[width=\textwidth]{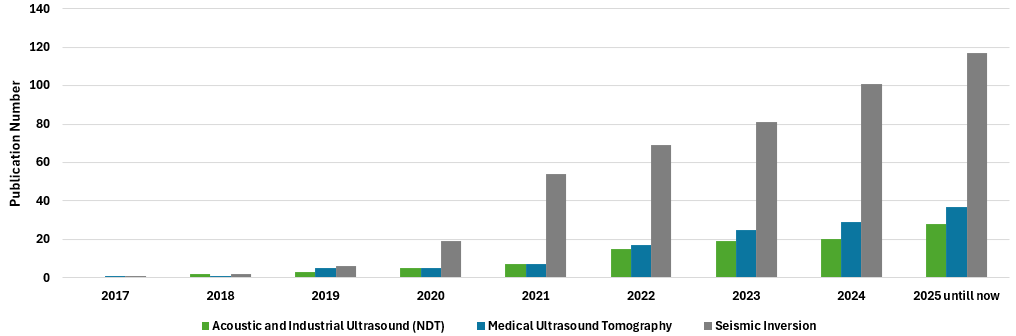}
    \caption{The overall trends of the ML methods developed in computational wave imaging. The figures are plotted based on a collection of over 600 papers.}
    \label{fig:counts}
\end{figure*}

\begin{figure*}[ht]
    \centering
    \includegraphics[width=1\textwidth]{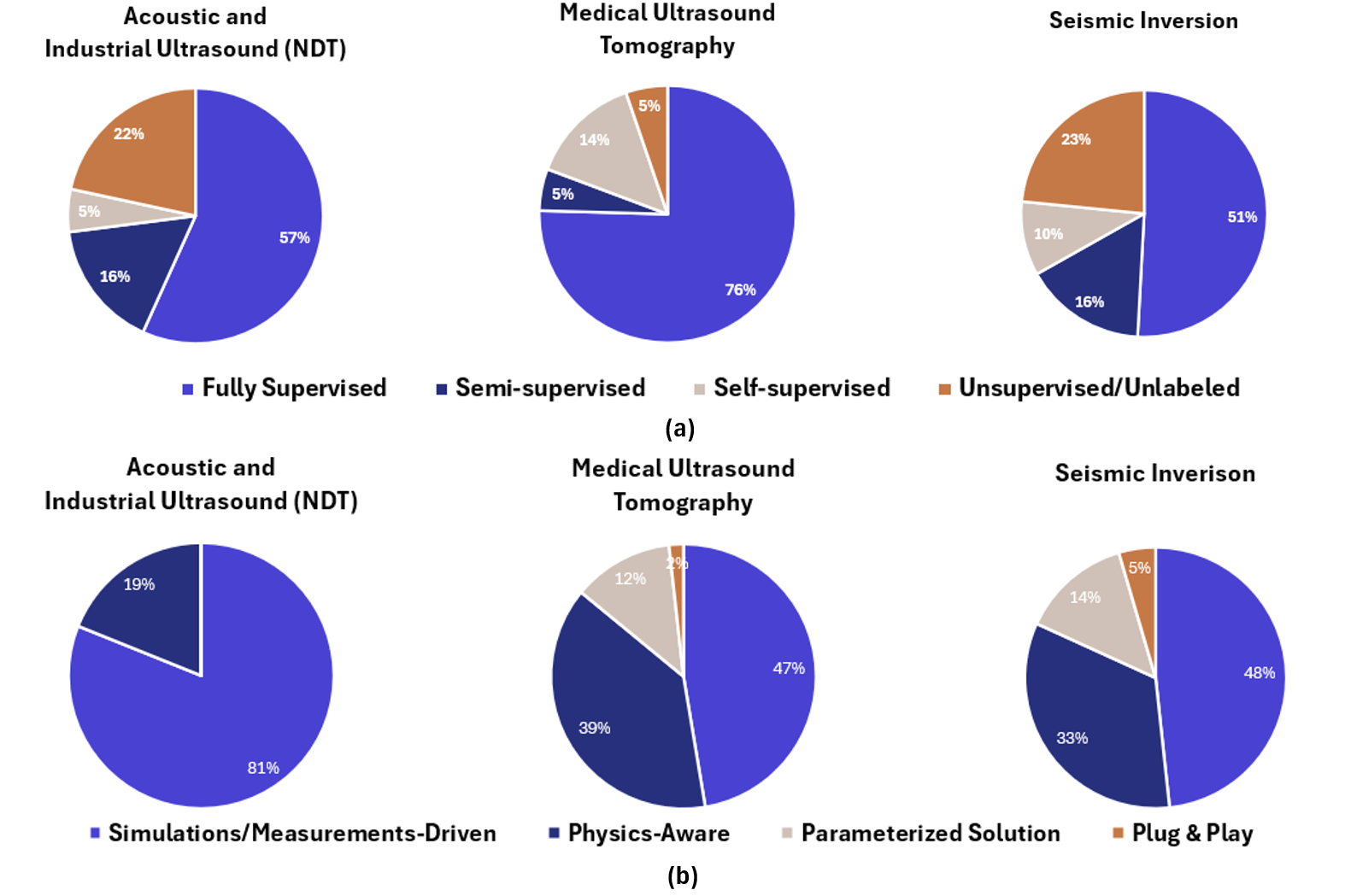}
\caption{Illustrations of (a) a breakdown of all the collected papers in four groups of methods under supervision strategy for acoustic imaging, ultrasound tomography, and seismic imaging, and (b) a breakdown of all the collected papers in four groups of methods under learning strategy for acoustic imaging, ultrasound tomography, and seismic imaging. }
\label{fig:paper_breakdown}
\end{figure*}

\section{Research Trends and Existing Review Papers}
\label{sec:Appendix-B}

Our review of the literature spans over 200 publications from diverse branches of the CWI, where a notable uptick in machine learning-centric studies is evident, as illustrated in Fig.~\ref{fig:counts}. These ML-focused methodologies have been systematically categorized in two ways: one is based on the nature of their training sets (whether samples are labeled, whether samples and labels are associated, etc); the other is based on the learning strategy (physics-aware, parameterized, etc.). The proportional representation of each category, delineated by the method of supervision and learning strategy, is graphically summarized in Figs.~\ref{fig:paper_breakdown}(a) and \ref{fig:paper_breakdown}(b), respectively.

Table~\ref{tab:summary2} summarizes a selection of high-quality review papers, categorized into three primary groups: physics-based methods, machine-learning approaches, and hybrid methods combining data-driven techniques with fundamental physics principles. The descriptions of each category are provided below.  Additionally, Table~\ref{tab:paperCollection} presents a curated collection of papers encompassing the full spectrum of categories and groups identified in our analysis.

\begin{itemize}
    \item \textbf{Physics.} Seismic FWI, acoustic NDT, and USCT have extensively utilized physics-based methods. In seismic FWI, comprehensive insights into techniques for exploration geophysics can be found in the work of \cite{Virieux-2009-Overview}. Additionally, \cite{Tromp-2020-Seismic} presents a distinct perspective tailored to global seismology challenges. \cite{Agudo-2018-Acoustic} provides a valuable summary of physics-based methods applied across both seismic and medical imaging domains. \cite{Sizing-2018-Felice} discusses the applications of acoustic imaging techniques, particularly in the realm of non-destructive flaw detection, localization, and sizing within materials.

    \item \textbf{Machine Learning.}~\cite{Deep-2020-Gregory} overviews the general topic of ML for computational imaging problems, mainly concentrating on general methodology development rather than a specific imaging problem. \cite{Deep-2021-Adler} provides an in-depth understanding of ML techniques tailored for seismic imaging challenges, with a specific focus on ``pure'' ML methods. \cite{Deep-2021-Yu} examines deep neural networks, exploring their diverse applications in geophysics, which encompass seismic imaging among others. \cite{Deep-2023-Wang} offers an extensive review of pure ML approaches, particularly for biomedical imaging, including USCT.

    \item \textbf{Hybrid.}~There are several reviews on hybrid models for general inverse problems~\cite{Integrating-2020-Willard, Physics-2021-Karniadakis, Scientific-2022-Cuomo}. Both \cite{Physics-2023-Banerjee} and \cite{kamilov-2023-plug} explore ML methods for solving imaging inverse problems. Particularly, \cite{Physics-2023-Banerjee}  explores methods and applications based on hybrid ML models, while \cite{kamilov-2023-plug} reviews the plug-and-play method and its applications. In the seismic FWI domain, \cite{Physics-2023-Lin} discusses hybrid ML methods with a focus on model development. 
    
\end{itemize}


\begin{table*}[t]
\begin{center}
\def\arraystretch{1.5}
\caption{Summary of representative physics-based, data-driven, and hybrid approaches for inverse problems across seismic inversion, medical ultrasound, and computational imaging. Checkmarks indicate whether a method relies primarily on physics-based modeling (Physics), machine learning (ML), or a hybrid combination of both.}
\label{tab:summary2} 
\begin{tabular}{c c c | c | c c c | c |c | c }
    \hline
    \multicolumn{3}{c|}{\textbf{Reference}} & \multicolumn{1}{c|}{\textbf{Year}} & \multicolumn{3}{c|}{\textbf{Application Domain}} & \multicolumn{1}{c|}{\textbf{Physics}} &\multicolumn{1}{c|}{\textbf{ML}} & \multicolumn{1}{c}{\textbf{Hybrid}}   \\ 
    \hline
    \hline
    \multicolumn{3}{c|}{Virieux and Operto~\cite{Virieux-2009-Overview}} & 2009 & \multicolumn{3}{c|}{Seismic Inversion} & \checkmark  &   & \\
    \hline
    \multicolumn{3}{c|}{Agudo~\cite{Agudo-2018-Acoustic}} & 2018 & \multicolumn{3}{c|}{Medical Ultrasound} & \checkmark  &   & \\
    \hline
    \multicolumn{3}{c|}{Felice and Fan~\cite{Sizing-2018-Felice}} & 2018 & \multicolumn{3}{c|}{Acoustic \& Industrial Ultrasound} & \checkmark  &   & \\
    \hline
    \multicolumn{3}{c|}{Khairi~\etal~\cite{Ultrasound-2019-Mohd}} & 2019 & \multicolumn{3}{c|}{Acoustic \& Industrial Ultrasound} & \checkmark  &   & \\
    \hline
    \multicolumn{3}{c|}{Tromp~\cite{Tromp-2020-Seismic}} & 2020 & \multicolumn{3}{c|}{Seismic Inversion} & \checkmark &   & \\
    \hline
    \multicolumn{3}{c|}{Operto~\etal~\cite{Extending-2023-Operto}} & 2023 & \multicolumn{3}{c|}{Seismic Inversion} & \checkmark  &  &  \\
    \hline
    \multicolumn{3}{c|}{Ongie~\etal~\cite{Deep-2020-Gregory}} & 2020 & \multicolumn{3}{c|}{Computational Imaging}  & & \checkmark  & \\
    \hline
    \multicolumn{3}{c|}{Wang~\etal~\cite{Deep-2023-Wang}} & 2020 & \multicolumn{3}{c|}{Medical Ultrasound} & & \checkmark  &\\  
    \hline
    \multicolumn{3}{c|}{Adler~\etal~\cite{Deep-2021-Adler}} & 2021 & \multicolumn{3}{c|}{Seismic Inversion} & & \checkmark  &\\
    \hline
    \multicolumn{3}{c|}{Yu and Ma~\cite{Deep-2021-Yu}} & 2021 & \multicolumn{3}{c|}{Seismic Inversion} &  & \checkmark  & \\  
    \hline
    \multicolumn{3}{c|}{Willard~\etal~\cite{Integrating-2020-Willard}} & 2020 & \multicolumn{3}{c|}{Inverse Problems}  &  &  & \checkmark \\
    \hline
    \multicolumn{3}{c|}{Karniadakis~\etal~\cite{Physics-2021-Karniadakis}} & 2021 & \multicolumn{3}{c|}{Inverse Problems} & &  & \checkmark  \\
    \hline
    \multicolumn{3}{c|}{Cuomo~\etal~\cite{Scientific-2022-Cuomo}} & 2022 & \multicolumn{3}{c|}{Inverse Problems} & &  & \checkmark  \\
    \hline
    \multicolumn{3}{c|}{Yu~\etal~\cite{yu_temperature_2022}} & 2022 & \multicolumn{3}{c|}{Inverse Problems} & &  & \checkmark  \\
    \hline
    \multicolumn{3}{c|}{Banerjee~\etal~\cite{Physics-2023-Banerjee}} & 2023 & \multicolumn{3}{c|}{Computational Imaging} & &  & \checkmark  \\
    \hline
    \multicolumn{3}{c|}{Kamilov~\etal~\cite{kamilov-2023-plug}} & 2023 & \multicolumn{3}{c|}{Computational Imaging} & &  & \checkmark  \\
    \hline
    \multicolumn{3}{c|}{Lin~\etal~\cite{Physics-2023-Lin}} & 2023 & \multicolumn{3}{c|}{Seismic Inversion} & &  & \checkmark  \\
    \hline
    \multicolumn{3}{c|}{Schuster~\etal~\cite{Review-2024-Schuster}} & 2024 & \multicolumn{3}{c|}{Seismic Inversion} & &  & \checkmark  \\
    \hline
    \multicolumn{3}{c|}{Yan~\etal~\cite{Review-2025-Yan}} & 2025 & \multicolumn{3}{c|}{Medical Ultrasound} & \checkmark  &  \checkmark  & \checkmark  \\
    \hline
  \end{tabular}
\end{center}  

\end{table*}



\section{Adaptive Data Augmentation -- A Semi-Supervised Method}
\label{sec:Appendix-D}


This appendix describes an adaptive data augmentation strategy for CWI, originally proposed by \cite{Renan-2022-Physics}, which is based on semi-supervised learning. The core idea is to iteratively refine the training dataset by generating new samples that are statistically closer to a given observed waveform $\bm{d}_{\text{obs}}$.

Conventional data augmentation techniques commonly used in image processing~(\eg, rotations, flips, scaling) are generally not applicable in wave-based imaging due to the underlying physical constraints. In contrast, this adaptive strategy creates additional training samples that are both physically consistent and specific to the target inversion task. The method consists of the following steps (as illustrated in Fig.~\ref{fig:Semisupervised}):

\begin{enumerate}
\item Estimate approximate solver $\mathcal{G}(\boldsymbol{\hat{\theta}}, \bm{d}_{\text{obs}})$;

\item Generate approximate velocity maps from unlabeled data $\hat{\bm{m}}_{r}=\mathcal{G}(\boldsymbol{\hat{\theta}}, \bm{d}_{\text{obs},r})$;

\item Create synthetic seismic data using forward model $\hat{\bm{d}}_{\text{obs}, r}=f(\hat{\bm{m}}_{r})$;

\item Add new pairs to the original training set.
\end{enumerate}

This semi-supervised augmentation scheme enhances model accuracy by injecting physically meaningful and geologically relevant examples into the training data. The process can be repeated iteratively, with each round retraining the operator $\mathcal{G}$ using the expanded dataset. This iterative refinement helps improve the quality of the reconstructed velocity maps for the observed data $\bm{d}_{\text{obs}}$.

The first step is to learn a reconstruction operator $g_0$ from the waveform-data manifold to velocity-map manifold using the original pairwise data. From the measured data, use $g_0(\bm{d}+\eta)$ to create a new batch of models $\bm{m}$, and then apply $f$ to those models to create a new batch of labeled samples.  The training set is augmented with the new labeled samples, and a new regression $g_1$ is fit.  With each iteration, more labeled samples are in the regime of the measured sample, and the performance of the resulting $g$ is improved for the waveform $\bm{d}$ of interest.

\section{Open-Source Software and Data Availability}
\label{sec:Appendix-E}

Accessible open data resources facilitate research on wave imaging challenges. Below, we provide a few open datasets for different wave imaging problems. 

\begin{itemize}

    \item \textbf{Seismic Full-Waveform Inversion.~}We find two open accessible datasets: \textsc{OpenFWI}~\cite{OpenFWI-2022-Deng} and $\mathbf{\mathbb{E}^{FWI}}$~\cite{EFWI-2023-Feng}, of which \textsc{OpenFWI} was designed for seismic imaging through the application of acoustic wave equations, and $\mathbf{\mathbb{E}^{FWI}}$ was tailored for seismic imaging employing elastic wave equations. Particularly, \textsc{OpenFWI} comprises 12 datasets that have been synthesized from a variety of sources. These datasets span a wide range of geophysical domains, including interfaces, faults, CO$_2$ reservoirs, and more. They also encompass different geological subsurface structures, such as flat and curved formations, and offer varying amounts of data samples. Within the \textsc{OpenFWI} collection, you can find both P-wave velocity models and seismic data. In contrast, $\mathbf{\mathbb{E}^{FWI}}$ encompasses 8 distinct datasets. Notably, the seismic dataset within $\mathbf{\mathbb{E}^{FWI}}$ includes both vertical and horizontal components, providing a more comprehensive view of the seismic information. Additionally, the velocity models in $\mathbf{\mathbb{E}^{FWI}}$ incorporate both P-wave and S-wave velocities, offering a richer dataset for waveform inversion studies. Both of these valuable datasets can be accessed through the following URLs:~\url{https://openfwi-lanl.github.io/} for \textsc{OpenFWI} and \url{https://efwi-lanl.github.io/} for $\mathbf{\mathbb{E}^{FWI}}$. The source code for both InversionNet and VelocityGAN can be accessed on the GitHub repository at ~\url{https://github.com/lanl/OpenFWI}, and to facilitate researchers in their learning process, a comprehensive Colab Tutorial is additionally made available for convenient access via this link: \url{https://colab.research.google.com/drive/17s5JmVs9ABl8MpmFlhWMSslj9_d5Atfx?usp=sharing}.

    \item \textbf{Acoustic and Industrial Ultrasound.} In industrial settings, events such as crack propagation, shocks, impacts, or other events generate Acoustic Emission (AE) signals, which can be measured via passive listening. In Ref. [\citenum{verdin_harvard_2021}], acoustic measurements were collected on bolted joints to detect events where the bolts were untightened, which resulted in AE signals. The data set, ORION-AE, containing the time-series measurements can be accessed via \url{https://github.com/emmanuelramasso/ORION_AE_acoustic_emission_multisensor_datasets_bolts_loosening}. Stiller et al. utilizes impact based AE measurements to identify damage modes in composites under compression. Analysis software and datasets can be accessed via \url{https://github.com/DominikStiller/tudelft-damage-identification}.
    Active interrogation is also a popular method for detecting defects or damage in industrial structures. In Ref. [\citenum{virkkunen_arxiv_2019}], a CNN is used to detect cracks based on acoustic phased-array measurements. The source code and dataset can be accessed via \url{https://github.com/iikka-v/ML-NDT}. Ref. [\citenum{moleroarmenta_compphyscomm_2014}] provides a finite element method simulation package to simulate acoustic NDT measurements based on Pulse-echo, Through-Transmission, and Linear Scanning. The source code to generate synthetic data can be accessed via \url{https://github.com/mmolero/SimNDT}.
    In addition to industrial measurements, there has been a push to classify sounds found commonly in everyday life, such as cars, animal calls, construction, music, etc. Google has extracted 10-second audio clips from YouTube videos to compile AudioSet, a labeled data set with 632 distinct event classes that is accessible via \url{http://research.google.com/audioset/index.html}. In Ref. [\citenum{salamon_icm_2014}] New York University presents the UrbanSound8k dataset containing 8732 4-second audio clips with labels of events common in urban environments. This dataset can be accessed via \url{https://urbansounddataset.weebly.com/urbansound8k.html}.   
    
    \item \textbf{Medical Ultrasound Tomography.} In Refs.~[\citenum{Li20213Dstochastic,Li2023forward}], the authors develop a three-dimensional anatomically realistic numerical phantom of the female breast and a high-fidelity three-dimensional imaging model of ring-array USCT breast ultrasound. The goal of the generated numerical breast phantoms (NBPs) and corresponding simulated measurements is to provide a dataset to 1) support the development and assessment (using objective measurement of image quality) of model-based and learned image reconstruction methods and 2) conduct virtual imaging trials to guide the design of new USCT breast imaging systems. Compared to other acoustic NBPs in the literature, the ones in Ref.~[\citenum{Li20213Dstochastic}] are particularly relevant to the development of new learning-based USCT image reconstruction method since they 1) comprise realistic structures and acoustic properties; 2) include lesions and/or other pathologies; and 3) are representative of the stochastic variability in breast size, shape, composition, anatomy, and tissue properties observed in a specified cohort of to-be-imaged subjects.
    Leveraging tools developed and released by the Virtual Imaging Clinical Trials for Regulatory Evaluation (VICTRE) project~\cite{badano2018evaluation,badano2021silico} of the Food and Drug Administration (FDA), the NBPs in Ref.~[\citenum{Li20213Dstochastic}] correspond to the four different levels of breast density defined according to the American College of Radiology's (ACR) Breast Imaging Reporting and Data System (BI-RADS) \cite{american2013acr}: A) Breast is almost entirely fat, B) Breast has scattered areas of fibroglandular density, C) Breast is heterogeneously dense, and D) Breast is extremely dense. Each NBP is a 3D voxelized map consisting of ten tissue types: fat,  skin,  glandular,  nipple, ligament (connective tissue), muscle, terminal duct lobular unit, duct, artery, and vein. In the framework of the authors, large ensembles of stochastic NBPs with realistic variability in breast volume, shape, fraction of glandular tissue, ligament orientation, tissue anatomy, can be generated by controlling input parameters and selecting the random seed number. A variety of lesions  (e.g.,  circumscribed or spiculated masses) can also be inserted at physiologically plausible locations. Tissue-specific acoustic properties (density, SOS, and AA) are then stochastically assigned to each tissue structure within physiological ranges and acoustic heterogeneity within fatty and glandular tissues is also modeled as spatially correlated random fields. Realization of these numerical phantoms and corresponding simulated measurements can be downloaded from Refs.~[\citenum{li2021NBPs3D, li2021NBPs2D, Li2023NBS3D}], and additional NBPs can be generated using the open-source software in Ref.~[\citenum{Fu2021code}]. \cite{zeng2025openwaves} presents a large-scale, anatomically realistic ultrasound dataset designed to benchmark neural solvers for wave equations. Featuring over 16 million simulations across diverse breast phantoms, it enables evaluation of neural operators in both forward and inverse tasks under clinically relevant conditions. The dataset addresses scalability, realism, and generalization gaps, and is publicly available to support reproducible research in neural wave-based imaging. Last but not least, \cite{Openpros-2025-Wang} introduces the first large-scale open dataset specifically designed for limited-view prostate USCT. It provides over 280,000 paired samples of realistic 2D speed-of-sound phantoms and corresponding full-waveform ultrasound data, derived from anatomically accurate 3D prostate models based on clinical MRI/CT and expert annotations. This dataset addresses a critical gap in prostate cancer imaging research, enabling development and benchmarking of neural inversion methods under realistic limited-angle conditions. Baseline experiments show that while deep learning significantly outperforms traditional physics-based methods in both speed and accuracy, there remains room for improvement before achieving clinical-grade resolution.

\end{itemize}

\begin{sidewaystable*}[t]
\centering
\caption{Paper Summary}
\label{tab:paperCollection}
\begin{adjustbox}{width=1\textwidth}
\begin{tabular}{c|c|c|c|c|c|c|c|c|c} 
\hline
\multicolumn{2}{c|}{\multirow{3}{*}{\textbf{Reference}}} & \multicolumn{4}{c|}{\textbf{Pure ML Methods}}                                                              & \multicolumn{4}{c}{\textbf{Hybrid Methods}}                                                                           \\ 
\cline{3-10}
\multicolumn{2}{c|}{}                                    & \textbf{Supervised} & \textbf{Semi-Supervised} & \textbf{Self-Supervised} & \begin{tabular}[c]{@{}c@{}}\textbf{Unsupervised/}\\\textbf{Unlabeled}\end{tabular} &\begin{tabular}[c]{@{}c@{}}\textbf{Simulations/}\\\textbf{Measurements-Driven}\end{tabular} & \textbf{\textbf{Physics Aware}} &\begin{tabular}[c]{@{}c@{}}\textbf{Parameterized}\\\textbf{Solution}\end{tabular}& \textbf{Plug \& Play}  \\ 
\hline
\multirow{14}{*}{\textbf{Seismic}}    & Wu and Lin~\cite{wu-2019-inversionnet}        & \checkmark                   &                          &                          &                                & \checkmark                           &                                 &                                 &                     \\
                                     & Araya-Polo et al.~\cite{araya2018deep} & \checkmark                   &                          &                          &                                & \checkmark                           &                                 &                                 &                     \\                                     & Renan et al.~\cite{Renan-2022-Physics}         &\checkmark                     &                         &                          &                                &  \checkmark                           &                                &                                 &                     \\                 & Wang et al.~\cite{wang2023seismic}         &\checkmark                      &                        &                          &                                &  \checkmark                           &                              &                                 &                     \\                  & Kazei et al.~\cite{kazei2021mapping}         &\checkmark                      &                        &                          &                                &  \checkmark                           &                              &                                 &                     \\                 
                                     & Wu et al.~\cite{wu2021semi}         &                     & \checkmark                        &                          &                                &                             & \checkmark                               &                                 &                     \\                                 
                                     & Sun et al.~\cite{Learning-2023-Sun}        &                     & \checkmark                        &                          &                                &                             & \checkmark                               &                                 &                     \\
                                     & Hu et al.~\cite{hu2020physics}          &                     &                          & \checkmark                        &                                &                             & \checkmark                               &                                 &                     \\
                                     & Zhu et al.~\cite{Integrating-2021-Zhu}         &                     &                          &                          & \checkmark                              &                             &                                 & \checkmark                               &                     \\
                                     & Saad et al.~\cite{SiameseFWI-2024-Saad}         &                     &                          &                          & \checkmark                              &                             &                                 & \checkmark                               &                     \\
                                     & Wu et al.~\cite{wu2021cnn}          &                   &                          &  \checkmark                         &                                &                             &                                 & \checkmark                               &                 \\                                     & Fu et al.~\cite{Fu-2023-Efficient}          &                     &                          &                          & \checkmark                                &                             &                                 &                                 & \checkmark                   \\ 
                                     
                                     & Vamaraju et al.~\cite{vamaraju2019unsupervised}          &                     &                          &                          & \checkmark                                &                             &   \checkmark                              &                                 &                    \\ 
                                     & Rasht-Behesh et al.~\cite{Physics-2022-Behesht}          &                     &                          &                          & \checkmark                                &                             &   \checkmark                              &                                 &                    \\ 
                                     & Izzatullah et al.~\cite{Izzatullah-2023-Plug}          &                     &                          &                          & \checkmark                                &                             &                                 &                                 & \checkmark                   \\
                                     & Zhang and Wang~\cite{Progressive-2025-Zhang}          &                     &                          &                          & \checkmark                                &                             &                                 &                                 & \checkmark                   \\
                                     
\hline
\multirow{10}{*}{\begin{tabular}[c]{@{}c@{}}\textbf{Medical}\\\textbf{Ultrasound}\end{tabular}}   
& Lozenski et al.~\cite{Learned-2023-Lozenski}          & \checkmark                   &                          &                          &                                &      \checkmark                        &                                   &                                 &                     \\
& Vedula et al.~\cite{vedula2017towards}          & \checkmark                   &                          &                          &                                &      \checkmark                        &                                   &                                 &                     \\
                                     & Feigin et al.~\cite{feigin2019deep}          & \checkmark                    &                         &                          &                                &  \checkmark                           &                                 &                                 &                     \\ 
                                     & Zhao et al.~\cite{zhao2020ultrasound}          & \checkmark                    &                         &                          &                                &  \checkmark                           &                                 &                                 &                     \\  
                                     & Tong et al.~\cite{tong2022deep}          & \checkmark                    &                         &                          &                                &  \checkmark                           &                                 &                                 &                     \\  
                                     & Liu et al.~\cite{liu2021deep}          &                     &\checkmark                          &                         &                                &                             &  \checkmark                               &                                 &                     \\                                & Zhang et al.~\cite{zhang2021general}          &                   &                         & \checkmark                           &                                &  \checkmark                           &                                 &                                 &                     \\
                                     & Zhang et al.~\cite{zhang2020self}          &                     &                          &    \checkmark                       &                                & \checkmark                           &                                 &                                 &                     \\
                                     & Dai et al.~\cite{dai2021self}          &                     &                          &      \checkmark                      &                                &                              & \checkmark                                 &                                 &                     \\                & Dai et al.~\cite{liu2021ultrasound}          &                     &                          &                         &   \checkmark                                &                              & \checkmark                                 &                                 &                     \\         & Fan et al.~\cite{fan2022model}          &  \checkmark                    &                          &                         &                                  &                              &                                 &  \checkmark                                &                     \\ 
                                     
\hline

\multirow{19}{*}{\begin{tabular}[c]{@{}c@{}}\textbf{Acoustic/}\\\textbf{Ultrasonic NDT}\end{tabular}} &Lahivaara et al.~\cite{lahivaara2018deep}          & \checkmark                    &                          &                          &                                &\checkmark                             &                                 &                                 &                     \\
&Wang et al.~\cite{wang2022ultrasonic}          & \checkmark                    &                          &                          &                                &\checkmark                             &                                 &                                 &                     \\
&Rao et al.~\cite{Quantitative-2023-Rao}           & \checkmark                    &                          &                          &                                &\checkmark                             &                                 &                                 &                     \\
&Hong et al.~\cite{Hong2021MSE}           & \checkmark                    &                          &                          &                                & \checkmark                           &                                 &                              &      \\
&Prakash et al.~\cite{Prakash2023}           & \checkmark                    &                          &                          &                                &  \checkmark                           &                                 &                                 &                \\
&Song et al.~\cite{Song2020}           & \checkmark                    &                          &                          &                                &  \checkmark                 &  \checkmark                   &  \checkmark                               &                  \\
&Siljama et al.~\cite{Siljama2021JNE}           & \checkmark                    &                          &                          &                                & \checkmark                    & \checkmark                   &                                 &               \\
&Rudolph et al.~\cite{Rudolph2021}           &                     & \checkmark                         &                          &                                & \checkmark                            &                                 &                                 &                   \\
&Sen et al.~\cite{Sen2019MSSP}           &                     & \checkmark                         &                          &                                &  \checkmark                   &  \checkmark             &  \checkmark                               &               \\

&Jiang et al.~\cite{Jiang2023RESS}           &                     & \checkmark                         &                          &                                & \checkmark                             &                                 &                                 &                  \\
&Bouzenad et al.~\cite{Bouzenad2019}           &                     & \checkmark                         &                          &                                &  \checkmark                  & \checkmark                           &                                 &                     \\
&Akcay et al.~\cite{Akcay2018}           &                     & \checkmark                         &                          &                                &  \checkmark                           &                                 &                                 &                     \\
&Luiken et al.~\cite{Luiken2023}           &                     &                          &\checkmark                          &                                & \checkmark                   & \checkmark                   & \checkmark                                &                     \\
&Shukla et al.~\cite{Shukla2020JNE}           &                     &                          &                         &\checkmark                                 & \checkmark                  &  \checkmark                  &   \checkmark                              &                     \\
&Virupakshappa et al.~\cite{Virupakshappa2019}           &                     &                          &                         &\checkmark                                 & \checkmark                   &  \checkmark              &                                 &                     \\
&Kraljevski et al.~\cite{Kraljevski2021IEEE}           &                     &                          &                         &\checkmark                                 & \checkmark                  &  \checkmark                  &                                 &                     \\
&Wang et al.~\cite{Wang2023NETWORK}           &                     &                          &                         &\checkmark                                 &  \checkmark                  &   \checkmark                                &                                 &                     \\
&Rautela et al.~\cite{Rautela2022CS}           &                     &                          &                         &\checkmark                                 & \checkmark                    &  \checkmark                    & \checkmark                                 &                     \\
&Sawant et al.~\cite{Sawant2023ultrasonics}           &                     &                          &                         &\checkmark                          & \checkmark                    &  \checkmark                             &                                 &                     \\
\hline
\end{tabular}
\end{adjustbox}
\end{sidewaystable*}

\section{List of Acronyms}
For the convenience of the reader, this appendix provides a list of acronyms used throughout the manuscript. Each term is defined below to ensure clarity and consistency.

\begin{description}
  \item[ADMM] \hspace{2em} Alternating Direction Method of Multipliers
  \item[AE] \hspace{2em} Autoencoder
  \item[AI] \hspace{2em} Artificial Intelligence
  \item[CNN] \hspace{2em} Convolutional Neural Network
  \item[CT] \hspace{2em} Computed Tomography
  \item[CWI] \hspace{2em} Computational Wave Imaging
  \item[EFWI] \hspace{2em} Elastic Full Waveform Inversion
  \item[FCN] \hspace{2em} Fully Convolutional Network
  \item[FNO] \hspace{2em} Fourier Neural Operator
  \item[FWI] \hspace{2em} Full Waveform Inversion
  \item[FWIGAN] \hspace{2em} Full Waveform Inversion Generative Adversarial Network
  \item[GAN] \hspace{2em} Generative Adversarial Network
  \item[KL] \hspace{2em} Kullback-Leibler
  \item[MAE] \hspace{2em} Mean Absolute Error
  \item[MAP] \hspace{2em} Maximum A Posteriori
  \item[MCMC] \hspace{2em} Markov Chain Monte Carlo
  \item[ML] \hspace{2em} Machine Learning
  \item[ML-NDT] \hspace{2em} Machine Learning for Nondestructive Testing
  \item[MRI] \hspace{2em} Magnetic Resonance Imaging
  \item[MSE] \hspace{2em} Mean Squared Error
  \item[NDT] \hspace{2em} Nondestructive Testing
  \item[OOD] \hspace{2em} Out of Distribution
  \item[PDE] \hspace{2em} Partial Differential Equation
  \item[PINN] \hspace{2em} Physics-Informed Neural Network
  \item[PPP] \hspace{2em} Plug-and-Play Prior
  \item[PPP-CWI] \hspace{2em} Plug-and-Play Prior for Computational Wave Imaging
  \item[SDE] \hspace{2em} Stochastic Differential Equation
  \item[SNR] \hspace{2em} Signal-to-Noise Ratio
  \item[SOS] \hspace{2em} Speed of Sound
  \item[SSIM] \hspace{2em} Structural Similarity Index
  \item[TV] \hspace{2em} Total Variation
  \item[UPFWI] \hspace{2em} Unsupervised Physics-Informed Full Waveform Inversion
  \item[UQ] \hspace{2em} Uncertainty Quantification
  \item[USCT] \hspace{2em} Ultrasound Computed Tomography
  \item[VAE] \hspace{2em} Variational Autoencoder
\end{description}

\bibliography{main}

@misc{BrainPuzzle-2025-Chen,
      title={{BrainPuzzle:} Hybrid Physics and Data-Driven Reconstruction for Transcranial Ultrasound Tomography}, 
      author={Chen, Shengyu and Feng, Shihang  and Luo, Yi and Jia, Xiaowei  and Lin, Youzuo},
      year={2025},
      eprint={2510.20029},
      archivePrefix={arXiv},
      url={https://arxiv.org/abs/2510.20029}
}

@ARTICLE{Robust-2025-Chen,
  author={Chen, Haotian and Han, Aiguo},
  journal={IEEE Transactions on Medical Imaging}, 
  title={Robust Deep Learning for Pulse-Echo Speed of Sound Imaging via Time-Shift Maps}, 
  year={2026},
  volume={45},
  number={2},
  pages={463-476},
  doi={10.1109/TMI.2025.3602000}
  }

@article{Ultrasound-2019-Mohd,
author = {Taufiq Mohd Khairi, Mohd and Ibrahim, Sallehuddin  and Amri Md Yunus, Mohd and Faramarzi, Mahdi and Pei Sean, Goh and Pusppanathan, Jaysuman and Abid, Azwad},
title = {Ultrasound computed tomography for material inspection: Principles, design and applications},
journal = {Measurement},
volume = {146},
pages = {490-523},
year = {2019},
issn = {0263-2241},
doi = {https://doi.org/10.1016/j.measurement.2019.06.053}
}

@article{Review-2025-Yan,
      title={Review of Current Advances in Ultrasound Computed Tomography for Medical Imaging}, 
      author={Yan, Weicheng and He, Lei and Zhang, Hui  and Wang, Zesong and Zeng, Xiaolu and Tan, Hongrui  and Zhou, Xiang and Wu, Yun and Liu, Zhaohui and Fan, Liren and Zhang, Pengcheng and Guo, Zhaoyuan and Xie, Gaoxiang and Cai, Chao and Ding, Mingyue and Yuchi, Ming and Qiu, Wu },
      year={2025},
      archivePrefix={TechRxiv},
      primaryClass={physics.med-ph},
      url={https://10.36227/techrxiv.176472780.09188058/v1}
}

@inproceedings{Openpros-2025-Wang,
      title={{OpenPros:} A Large-Scale Dataset for Limited View Prostate Ultrasound Computed Tomography}, 
      author={Wang, Hanchen and Wu, Yixuan and Feng, Yinan  and Jin, Peng  and Feng, Shihang  and Zhang, Luoyuan and Wiskin, James  and Turkbey, Baris  and Pinto, Peter A. and Wood, Bradford J.  and Luo, Songting  and Chen, Yinpeng  and Boctor, Emad  and Lin, Youzuo},
      year={2026},
    booktitle =	 {The International Conference on Learning Representations~(ICLR)},
      url={https://arxiv.org/abs/2505.12261}
}

@misc{zeng2025openwaves,
title={OpenWaves: A Large-Scale Anatomically Realistic Ultrasound-{CT} Dataset for Benchmarking Neural Wave Equation Solvers},
author={Zeng, Zhijun and Zheng, Youjia  and Hu, Hao and Dong, Zeyuan  and Zheng, Yihang  and Liu, Xinliang and Wang, Jinzhuo  and Shi, Zuoqiang  and Zhang, Linfeng  and Li, Yubing and Sun, He},
year={2025},
url={https://openreview.net/forum?id=u14Y236LwX}
}

@article{Extending-2023-Operto,
author = {Operto, Stéphane and Gholami, Ali and Aghamiry,  Hossein and Guo, Gaoshan and Beller, Stephen and Aghazade, Kamal and Mamfoumbi,  Frichnel and Combe, Laure and Ribodetti, Alessandra},
title = {Extending the search space of full-waveform inversion beyond the single-scattering Born approximation: A tutorial review},
journal = {GEOPHYSICS},
volume = {88},
number = {6},
pages = {R671-R702},
year = {2023},
doi = {10.1190/geo2022-0758.1}
}

@article{Quantifying-2015-Lin,
    author = {Lin, Youzuo and Huang, Lianjie},
    title = {Quantifying subsurface geophysical properties changes using double-difference seismic-waveform inversion with a modified total-variation regularization scheme},
    journal = {Geophysical Journal International},
    volume = {203},
    number = {3},
    pages = {2125-2149},
    year = {2015},
    month = {11},
    doi = {10.1093/gji/ggv429}
}

@article{Robust-2025-Aghazade,
      title={Robust acoustic and elastic full waveform inversion by adaptive Tikhonov-TV regularization}, 
      author={Aghazade, Kamal and Gholami, Ali},
      year={2025},
      eprint={2505.04022},
      archivePrefix={arXiv},
      primaryClass={physics.geo-ph},
      url={https://arxiv.org/abs/2505.04022}, 
  doi       = {https://doi.org/10.48550/arXiv.2505.04022}
}

@article{kalita2019Regularized,
  title     = {Regularized full‐waveform inversion with automated salt flooding},
  author    = {Kalita, Mahesh and Kazei, Vladimir and Choi, Yunseok and Alkhalifah, Tariq},
  journal   = {Geophysics},
  volume    = {84},
  number    = {4},
  pages     = {R569-R582},
  year      = {2019},
  doi       = {https://doi.org/10.1190/geo2018-0146.1}
}

@article{aghamiry2020hybrid,
    author = {Aghamiry, Hamed and Gholami, Amir and Operto, Sergio},
    title = {Full waveform inversion by proximal Newton method using adaptive regularization},
    journal = {Geophysical Journal International},
    volume = {224},
    issue = {1},
    pages = {169-180},
    year = {2020},
    doi = {https://doi.org/10.1093/gji/ggaa434}
}

@article{Elastic-2023-Dhara,
      title={Elastic Full-Waveform Inversion Using a Physics-Guided Deep Convolutional Encoder–Decoder}, 
      author={Dhara, Arnab and Sen, Mrinal},
      journal={IEEE Transactions on Geoscience and Remote Sensing},
      volume={61},
      year={2023},
doi={https://10.1109/TGRS.2023.3294427}
}

@article{Progressive-2025-Zhang,
author = {Zhang, Benwen and Wang, Linrong},
title = {Progressive plug and play full waveform inversion with multitask learning},
journal = {Scientific Reports},
volume = {15},
number = {19805},
doi = {https://doi.org/10.1038/s41598-025-04506-2},
year = {2025}
}

@article{One-2025-Jiang,
author = {Jiang, Yiran and Ma, Jianwei and Ning, Jieyuan and Li, Jiaqi and Wu, Han and Bao, Tiezhao},
title = {One-Fit-All Transformer for Multimodal Geophysical Inversion: Method and Application},
journal = {Journal of Geophysical Research: Machine Learning and Computation},
volume = {2},
number = {1},
pages = {e2024JH000432},
doi = {https://doi.org/10.1029/2024JH000432},
year = {2025}
}

@ARTICLE{Seismic-2025-Jia,
  author={Jia, Anqi and Sun, Jian and Du, Bo and Lin, Yuzhao},
  journal={IEEE Transactions on Geoscience and Remote Sensing}, 
  title={Seismic Full Waveform Inversion With Uncertainty Analysis Using Unsupervised Variational Deep Learning}, 
  year={2025},
  volume={63},
  number={},
  pages={1-16},
  doi={10.1109/TGRS.2025.3564647}}

@inproceedings{Self-2020-Wang,
  title={Self-supervised learning for low frequency extension of seismic data},
  author={Wang, Meixia and Xu, Sheng and Zhou, Hongbo},
  booktitle={SEG International Exposition and Annual Meeting},
  year={2020},
doi={https://doi.org/10.1190/segam2020-3427086.1}
}

@article{SiameseFWI-2024-Saad,
author = {Saad, Omar M. and Harsuko, Randy and Alkhalifah, Tariq},
title = {{SiameseFWI:} A Deep Learning Network for Enhanced Full Waveform Inversion},
journal = {Journal of Geophysical Research: Machine Learning and Computation},
volume = {1},
number = {3},
pages = {e2024JH000227},
doi = {https://doi.org/10.1029/2024JH000227},
year = {2024}
}

@article{Self-2024-Cheng,
author = {Cheng, Shijun and Wang, Yi and Zhang, Qingchen and Harsuko, Randy and Alkhalifah, Tariq},
title = {A Self-Supervised Learning Framework for Seismic Low-Frequency Extrapolation},
journal = {Journal of Geophysical Research: Machine Learning and Computation},
volume = {1},
number = {3},
pages = {e2024JH000157},
doi = {https://doi.org/10.1029/2024JH000157},
year = {2024}
}

@Article{Review-2024-Schuster,
  Title                    = {Review of Physics-Informed Machine Learning Inversion of Geophysical Data},
  Author                   = {Schuster, Gerard and Chen, Yuqing and Feng, Shihang},
  Journal                  = {Geophysics},
  Year                     = {2024},
  Volume                   = {89},
  Issue                    = {6},
  doi={https://doi.org/10.1190/geo2023-0615.1}
}

@inproceedings{Parameterizing-2020-Rizzuti,
  title={Parameterizing uncertainty by deep invertible networks: An application to reservoir characterization},
  author={Rizzuti, Gabrio and Siahkoohi, Ali and Witte, Philipp and Herrmann, Felix},
  booktitle={SEG International Exposition and Annual Meeting},
  year={2020},
doi={https://doi.org/10.1190/segam2020-3428150.1}
}

@Article{Deep-2022-Siahkoohi,
  Title                    = {Deep {Bayesian} inference for seismic imaging with tasks},
  Author                   = {Siahkoohi, Ali and Rizzuti, Gabrio and Herrmann, Felix},
  Journal                  = {Geophysics},
  Year                     = {2022},
  Pages                    = {1SO--V558},
  Volume                   = {87},
  Issue                    = {5},
  doi={https://doi.org/10.1190/geo2021-0666.1}
}

@Article{Ultrasound-2017-Sigrist,
  Title                    = {{Ultrasound Elastography:} Review of Techniques and Clinical Applications},
  Author                   = {Sigrist, Rosa and Liau, Joy and Kaffas, Ahmed El and Chammas, Maria and Willmann, Juergen},
  Journal                  = {Theranostics},
  Year                     = {2017},
  Pages                    = {1303--1329},
  Volume                   = {7},
  Issue                    = {5},
  doi={https://doi.org/10.7150/thno.18650}
}

@book{Quantitative-2023-Ruiter,
author= {Ruiter, Nicole and Zapf, Michael and Hopp, Torsten
and Gemmeke, Hartmut},
editor= {Mamou, Jonathan and Oelze, Michael},
title=  {Ultrasound Tomography},
bookTitle= {Quantitative Ultrasound in Soft Tissues},
year= {2023},
pages= {171--200},
publisher={Springer},
doi={https://doi.org/10.1007/978-3-031-21987-0_9}
}

@article{Deep-2017-Jin,
  title={{Deep Convolutional Neural Network for Inverse Problems in Imaging}},
  author={Jin, Kyong Hwan and McCann,  Michael and Froustey, Emmanuel and Unser, Michael},
  journal={IEEE Transactions on Image Processing},
  volume={26},
  number={9},
  pages={4509-4522},
  year={2017},
doi={https://doi.org/10.1109/TIP.2017.2713099}
}

@inproceedings{AutoLinear-2024-Feng,
  author =	 {Feng, Yinan and Chen, Yinpeng and Jin, Peng and Feng, Shihang and Lin, Youzuo},
  title =	 {Auto-Linear Phenomenon in Subsurface Imaging},
  booktitle =	 {The Forty-first International Conference on Machine Learning~(ICML)},
  year =	 {2024},
doi={
https://doi.org/10.48550/arXiv.2305.13314}
}

@article{Cai-2022-Semi,
  title={Semi-Supervised Surface Wave Tomography With Wasserstein Cycle-Consistent {GAN:} Method and Application to Southern California Plate Boundary Region},
  author={Cai, Ao and Qiu, Hongrui and Niu, Fenglin},
  journal={Journal of Geophysical Research: Solid Earth},
  volume={127},
  number={3},
  pages={e2021JB023598},
  year={2022},
  publisher={Wiley Online Library},
doi={ https://doi.org/10.1029/2021JB023598}
}

@inproceedings{Ronneberger-2015-UNet,
  title={U-net: Convolutional networks for biomedical image segmentation},
  author={Ronneberger, Olaf and Fischer, Philipp and Brox, Thomas},
  booktitle={Medical image computing and computer-assisted intervention--MICCAI 2015},
  pages={234--241},
  year={2015},
doi={https://doi.org/10.1007/978-3-319-24574-4_28}
}

@inproceedings{Ho-2020-Denoising,
  author =	 {Ho, Jonathan and Jain, Ajay and Abbeel, Pieter},
  title =	 {Denoising diffusion probabilistic models},
  booktitle =	 {The Advances in neural information processing systems~(NeurIPS)},
  year =	 {2020},
doi={
https://doi.org/10.48550/arXiv.2006.11239}
}

@inproceedings{Yang-2024-EdGeo,
  author =	 {Yang, Junhuan and Wang, Hanchen and Sheng, Yi and Lin, Youzuo and Yang, Lei},
  title =	 {EdGeo: A Physics-guided Generative AI Toolkit for Geophysical Monitoring on Edge Devices},
  booktitle =	 {The Design Automation Conference~(DAC)},
  year =	 {2024},
doi={
https://doi.org/10.48550/arXiv.2401.03131}
}

@inproceedings{Sohl-2015-Diffusion,
  author =	 {Sohl-Dickstein, Jascha and Weiss, Eric and Maheswaranathan, Niru and Ganguli, Surya},
  title =	 {Deep unsupervised learning using nonequilibrium thermodynamics},
  booktitle =	 {The International Conference on Machine Learning~(ICML)},
  year =	 {2015},
doi={
https://doi.org/10.48550/arXiv.1503.03585
}
}

@inproceedings{Goodfellow-2014-Generative,
  author =	 {Goodfellow, Ian and Pouget-Abadie, Jean and Mirza, Mehdi and Xu, Bing and Warde-Farley, David and Ozair, Sherjil and Courville, Aaron and Bengio, Yoshua},
  title =	 {Generative Adversarial Networks},
  booktitle =	 {The Advances in neural information processing systems~(NeurIPS)},
  year =	 {2014},
doi={
https://doi.org/10.48550/arXiv.1406.2661}
}

@inproceedings{Kingma-2014-Auto,
  author =	 {Kingma, Diederik and Welling, Max},
  title =	 {Auto-encoding variational bayes},
  booktitle =	 {The International Conference on Learning Representations~(ICLR)},
  year =	 {2014},
doi={https://doi.org/10.48550/arXiv.1312.6114}
}

@inproceedings{Song-2022-Solving,
  author =	 {Song, Yang and Shen, Liyue and Xing, Lei and Ermon, Stefano},
  title =	 {Solving Inverse Problems in Medical Imaging with Score-Based Generative Models},
  booktitle =	 {The Tenth International Conference on Learning Representations~(ICLR)},
  year =	 {2022},

doi={https://doi.org/10.48550/arXiv.2111.08005}

}

@article{DiGiT-2024-Yang,
  title={{DiGiT:} A Diffusion-based Modular Geophysical Toolkit Platform — A Case Study on On-Device Paired Geophysical Training Data Generation},
  author={Yang, Junhuan and Lin, Youzuo and Jiang, Weiwen},
  journal={arXiv preprint},
  year={2024},

}

@article{Prior-2023-Wang,
      title={A Prior Regularized Full Waveform Inversion Using Generative Diffusion Models}, 
      author={Wang, Fu and Huang, Xinquan and Alkhalifah, Tariq},
      JOURNAL    = {IEEE Transactions on Geoscience and Remote Sensing},
      YEAR       = {2023},
      VOLUME     = {61},
      ISSUE      = {},
      PAGES      = {1-11},
doi={
https://doi.org/10.1109/TGRS.2023.3337014}
}

@inproceedings{Interpretability-2017-Chakraborty,
  author =	 {Chakraborty, Supriyo and Tomsett, Richard and Raghavendra, Ramya and Harborne, Daniel and Alzantot, Moustafa and Cerutti, Federico and Srivastava, Mani and Preece, Alun and Julier, Simon and Rao, Raghuveer M. and Kelley, Troy D. and Braines, Dave and Sensoy, Murat and Willis, Christopher J. and Gurram, Prudhvi},
  title =	 {Interpretability of deep learning models: A survey of results},
  booktitle =	 {2017 IEEE SmartWorld, Ubiquitous Intelligence \& Computing, Advanced \& Trusted Computed, Scalable Computing \& Communications, Cloud \& Big Data Computing, Internet of People and Smart City Innovation},
  year =	 {2017},
doi={https://doi.org/10.1109/UIC-ATC.2017.8397411}
}

@article{Interpretable-2021-Li,
      title={Interpretable Deep Learning: Interpretation, Interpretability, Trustworthiness, and Beyond}, 
      author={Li, Xuhong and Xiong, Haoyi and Li, Xingjian and Wu, Xuanyu and Zhang, Xiao and Liu, Ji and Bian, Jiang and Dou, Dejing},
      journal={arXiv preprint},
      year={2021},

doi={https://doi.org/10.48550/arXiv.2103.10689}

}

@article{Uncertainty-2023-Yablokov,
  title={Uncertainty quantification of multimodal surface wave inversion using artificial neural networks},
  author={Yablokov, Alexandr and Lugovtsova, Yevgeniya and Serdyukov, Aleksander},
  journal={Geophysics},
  volume={88},
  number={2},
  pages={1MA--Y5},
  year={2023},
doi={https://doi.org/10.1190/geo2022-0261.1}
}

@inproceedings{Single-2019-Tagasovska,
  author =	 {Tagasovska, Natasa and Lopez-Paz, David},
  title =	 {Single-Model Uncertainties for Deep Learning},
  booktitle =	 {Advances in Neural Information Processing Systems 33 },
  year =	 {2019},
doi={
https://doi.org/10.48550/arXiv.1811.00908}
}

@article{Enhanced-2023-Liu,
      title={Enhanced prediction accuracy with uncertainty quantification in monitoring CO2 sequestration using convolutional neural networks}, 
      author={Liu, Yanhua and Zhang, Xitong and Tsvankin, Ilya and Lin, Youzuo},
      journal={arXiv preprint arXiv:2212.04567},
      year={2023},
doi={
https://doi.org/10.48550/arXiv.2212.04567
}
}

@book{Li-2021-Distributed,
  title={Distributed acoustic sensing in geophysics: Methods and applications},
  author={Li, Yingping and Karrenbach, Martin and Ajo-Franklin, Jonathan},
  volume={268},
  year={2022},
  publisher={John Wiley \& Sons},
  doi         = {https://doi.org/10.1002/9781119521808}

}

@article{Tackling-2019-Yao,
  title={Tackling cycle skipping in full-waveform inversion with intermediate data},
  author={Yao, Gang and da Silva, Nuno and Warner, Michael and Wu, Di and Yang, Chenhao},
  journal={Geophysics},
  volume={84},
  number={3},
  pages={R411--R427},
  year={2019},
  publisher={Society of Exploration Geophysicists},
  doi={https://doi.org/10.1190/geo2018-0096.1}
}

@article{firouzi2012first,
  title={A first-order k-space model for elastic wave propagation in heterogeneous media},
  author={Firouzi, Kamyar and Cox, BT and Treeby, BE and Saffari, N},
  journal={The Journal of the Acoustical Society of America},
  volume={132},
  number={3},
  pages={1271--1283},
  year={2012},
  publisher={AIP Publishing},
  doi={https://doi.org/10.1121/1.4730897}
}

@article{cox2007k,
  title={k-space propagation models for acoustically heterogeneous media: Application to biomedical photoacoustics},
  author={Cox, Benjamin T and Kara, S and Arridge, Simon R and Beard, Paul C},
  journal={The Journal of the Acoustical Society of America},
  volume={121},
  number={6},
  pages={3453--3464},
  year={2007},
  publisher={Acoustical Society of America},
  doi={https://doi.org/10.1121/1.2717409}
}

@article{Assessment-2019-Yang,
  author =	 {Yang, Xianjin and Buscheck, Thomas and Mansoor,  Kayyum and Wang, Zan and Gao, Kai and Huang, Lianjie and Wainwright, Haruko and Carroll, Susan},
  title =	 {Assessment of geophysical monitoring methods for detection of brine and {CO}$_2$ leakage in drinking water aquifers},
  journal =	 {International Journal Greenhouse Gas Control},
  volume =	 {90},
  pages = {102803},
  year = {2019},
doi={https://doi.org/10.1016/j.ijggc.2019.102803}
}

@article{Downhole-2019-Buscheck,
  author =	 {Buscheck, Thomas and Mansoor,  Kayyum and Yang, Xianjin and Wainwright, Haruko and Carroll, Susan},
  title =	 {Downhole pressure and chemical monitoring for {CO}$_2$ and brine leak detection in aquifers above a {CO}$_2$ storage reservoir},
  journal =	 {International Journal Greenhouse Gas Control},
  volume =	 {91},
  pages = {102812},
  year = {2019},
doi={https://doi.org/10.1016/j.ijggc.2019.102812}
}

@article{Data-2019-Zhou,
      title={A Data-Driven {CO$_2$} Leakage Detection Using Seismic Data and Spatial-Temporal Densely Connected Convolutional Neural Networks}, 
      author={Zhou, Zheng and Lin, Youzuo and Zhang, Zhongping and Wu, Yue and Wang, Zan and Dilmore, Robert and Guthrie, George},
  journal={International Journal of Greenhouse Gas Control},
  volume={90},
  pages={102790},
  year={2019},
doi={https://doi.org/10.1016/j.ijggc.2019.102790}
}

@inproceedings{Izzatullah-2023-Plug,
  author =	 {Izzatullah, M and Alkhalifah, T and Romero, J and Corrales, M and Luiken,  N and Ravasi,  M },
  title =	 {{Plug-and-Play} Stein variational gradient descent for Bayesian post-stack seismic inversion },
  booktitle =	 {84th EAGE Annual Conference \& Exhibition},
  pages ={1–5},
  year =	 {2023},
doi={https://doi.org/10.3997/2214-4609.202310177}
}

@article{Fu-2023-Efficient,
  title={An efficient plug-and-play regularization method for full waveform inversion},
  author={Fu, Hongsun and Yang, Lu and Miao, Xinyue},
  journal={Journal of Geophysics and Engineering},
  volume={20},
  number={6},
  pages={1140--1149},
  year={2023},
  publisher={Oxford University Press},
doi={https://doi.org/10.1093/jge/gxad073}
}

@article{Yang-2023-FWIGAN,
  title={{FWIGAN}: Full-Waveform Inversion via a Physics-Informed Generative Adversarial Network},
  author={Yang, Fangshu and Ma, Jianwei},
  journal={Journal of Geophysical Research: Solid Earth},
  volume={128},
  number={4},
  pages={e2022JB025493},
  year={2023},
  publisher={Wiley Online Library},
doi={ https://doi.org/10.1029/2022JB025493}
}

@ARTICLE{Wen-2022-UFNO,
  TITLE      = {{U-FNO}—An enhanced Fourier neural operator-based deep-learning model for multiphase flow},
  AUTHOR     = {Wen, Gege and Li, Zongyi and Azizzadenesheli, Kamyar and Anandkumar, Anima and Benson, Sally M },
  JOURNAL    = {Advances in Water Resources},
  YEAR       = {2022},
  VOLUME     = {163},
  ISSUE      = {},
  PAGES      = {104180},
doi={https://doi.org/10.1016/j.advwatres.2022.104180}
}

@ARTICLE{Lu-2021-DeepONet,
  TITLE      = {Learning nonlinear operators via {DeepONet} based on the universal approximation theorem of operators},
  AUTHOR     = {Lu, Lu and Jin, Pengzhan and Pang, Guofei and Zhang, Zhongqiang and Karniadakis, George Em},
  JOURNAL    = {Nature Machine Intelligence},
  YEAR       = {2021},
  VOLUME     = {3},
  ISSUE      = {},
  PAGES      = {218-229},
doi={https://doi.org/10.1038/s42256-021-00302-5}
}

@ARTICLE{Zhu-2023-DeepONet,
  TITLE      = {{Fourier-DeepONet:} {Fourier-enhanced} deep operator networks for full waveform inversion with improved accuracy, generalizability, and robustness},
  AUTHOR     = {Zhu, Min and Feng, Shihang and Lin, Youzuo and Lu, Lu},
  JOURNAL    = {Computer Methods in Applied Mechanics and Engineering},
  YEAR       = {2023},
  VOLUME     = {416},
  ISSUE      = {},
  PAGES      = {116300},
doi={https://doi.org/10.1016/j.cma.2023.116300}
}

@ARTICLE{Liu-2023-Physics,
  TITLE      = {Physics-driven self-supervised learning system for seismic velocity inversion},
  AUTHOR     = {Liu, Bin and Jiang, Peng and Wang, Qingyang and Ren, Yuxiao and Yang, Senlin and Cohn, Anthony},
  JOURNAL    = {Geophysics},
 volume={88},
  number={2},
  pages={R145--R161},
  year={2023},
  publisher={Society of Exploration Geophysicists},
doi={https://doi.org/10.1190/geo2021-0302.1}
}

@article{Learning-2023-Sun,
      title={Learning with real data without real labels: a strategy for extrapolated full-waveform inversion with field data}, 
      author={Sun, Hongyu and Sun, Yen and Nammourand, Rami and Rivera, Christian and Williamson, Paul and Demanet, Laurent},
      journal={Geophysical Journal International},
      volume={235},
      issue={2},
      pages={1761–1777},
      year={2023},
doi={https://doi.org/10.1093/gji/ggad330}
}

@techreport{Solving-2023-Gupta,
  author  = {Gupta, Naveen},
  title   = {Solving Forward and Inverse Problems for Seismic Imaging using Invertible Neural Networks},
  institution = {Virginia Tech},
  year    = {2023},
  doi={http://hdl.handle.net/10919/115742}
}

@inproceedings{Lin-2014-Ultrasound,
  author =	 {Lin, Youzuo and Huang, Lianjie},
  title =	 {Ultrasound waveform tomography with a spatially variant regularization scheme},
  booktitle={Medical Imaging 2014: Ultrasonic Imaging and Tomography},
  volume={9040},
  pages={421--427},
  year={2014},
doi={https://doi.org/10.1117/12.2043110}
}

@inproceedings{Lin-2013-Ultrasound,
  author =	 {Lin, Youzuo and Huang, Lianjie},
  title =	 {Ultrasound waveform tomography with a modified total-variation regularization scheme},
  booktitle={Medical Imaging 2013: Ultrasonic Imaging, Tomography, and Therapy},
  volume={8675},
  pages={379--387},
  year={2013},
doi={https://doi.org/10.1117/12.2007650}
}

@article{Implicit-2023-Sun,
      title={Implicit Seismic Full Waveform Inversion With Deep Neural Representation}, 
      author={Sun, Jian and Innanen, Kristopher and Zhang, Tianze and Trad, Daniel},
      JOURNAL    = {Journal of Geophysical Research: Solid Earth},
      YEAR       = {2023},
      VOLUME     = {128},
      ISSUE      = {3},
      PAGES      = {e2022JB025964},
doi={https://doi.org/10.1029/2022JB025964}
}

@article{Quantitative-2023-Rao,
      title={Quantitative reconstruction of defects in multi-layered bonded composites using fully convolutional network-based ultrasonic inversion}, 
      author={Rao, Jing and Yang, Fangshu and Mo, Huadong and Kollmannsberger, Stefan and Rank, Ernst},
      journal    = {Journal of Sound and Vibration},
      volume     = {542},
      number      ={},
      pages      = {117418},
      year       = {2023},
      doi        = {https://doi.org/10.1016/j.jsv.2022.117418}
}

@article{Learned-2023-Lozenski,
      title={Learned Full Waveform Inversion Incorporating Task Information  for Ultrasound Computed Tomography}, 
      author={Lozenski, Luke and Wang, Hanchen and Li, Fu and Anastasio, Mark and Wohlberg, Brendt and Lin, Youzuo and Villa, Umberto},
      journal    = {IEEE Transactions on Computational Imaging},
      volume={10},
      number={},
      pages={69-82},
      year={2024},
      doi={https://doi.org/10.1109/TCI.2024.3351529}
}

@article{Sizing-2018-Felice,
      title={Sizing of flaws using ultrasonic bulk wave testing: A review}, 
      author={Felice, Maria and Fan, Zheng},
      JOURNAL    = {Ultrasonics},
      YEAR       = {2018},
      VOLUME     = {88},
      ISSUE      = {},
      PAGES      = {26--42},
doi={https://doi.org/10.1016/j.ultras.2018.03.003}
}

@article{Deep-2023-Wang,
      title={Deep learning for tomographic image reconstruction}, 
      author={Wang, Ge and Ye, Jong Chul and De Man, Bruno},
      JOURNAL    = {Nature Machine Intelligence},
      YEAR       = {2020},
      VOLUME     = {2},
      ISSUE      = {},
      PAGES      = {737--748},
doi={https://doi.org/10.1038/s42256-020-00273-z}
}

@article{Physics-2023-Lin,
      title={Physics-Guided Data-Driven Seismic Inversion: Recent Progress and Future Opportunities in Full Waveform Inversion}, 
      author={Lin, Youzuo and Theiler, James and Wohlberg, Brendt},
      JOURNAL    = {IEEE Signal Processing Magazine},
      YEAR       = {2023},
      VOLUME     = {40},
      ISSUE      = {1},
      PAGES      = {115--133},
      DOI        = {https://doi.org/10.1109/MSP.2022.3217658}

}

@article{Physics-2023-Banerjee,
      title={Physics-Informed Computer Vision: A Review and Perspectives}, 
      author={Banerjee, Chayan and Nguyen, Kien and Fookes, Clinton and Karniadakis, George},
      JOURNAL    = {arXiv preprint arXiv:2305.18035},
      YEAR       = {2023},
      VOLUME     = {},
      ISSUE      = {},
      PAGES      = {},
doi={
https://doi.org/10.48550/arXiv.2305.18035
}
}

@article{Image-2023-Chen,
      title={Image as First-Order Norm+Linear Autoregression: Unveiling Mathematical Invariance}, 
      author={Chen, Yinpeng and Dai, Xiyang and Chen, Dongdong and Liu, Mengchen and Yuan, Lu and Liu, Zicheng and Lin, Youzuo},
      JOURNAL    = {arXiv preprint},
      YEAR       = {2023},
      VOLUME     = {},
      ISSUE      = {},
      PAGES      = {},
      DOI        = {https://doi.org/10.48550/arXiv.2305.16319}
}

@ARTICLE{Azevedo-2022-Model,
  TITLE      = {Model reduction in geostatistical seismic inversion with functional data analysis},
  author={Azevedo, Leonardo},
  journal={Geophysics},
  volume={87},
  number={1},
  pages={M1--M11},
  year={2022},
doi={https://doi.org/10.1190/geo2021-0096.1}
}

@article{InversionNet3D-2022-Zeng,
      title={{InversionNet3D}: Efficient and Scalable Learning for {3-D} Full-Waveform Inversion}, 
      author={Zeng, Qili and Feng, Shihang and Wohlberg, Brendt and Lin, Youzuo},
      journal={IEEE Transactions on Geoscience and Remote Sensing},
      volume={60},
      issue={},
      pages={1-16},
      year={2022},
doi={https://doi.org/10.1109/TGRS.2021.3135354}

}

@article{Scientific-2022-Cuomo,
      title={Scientific Machine Learning Through Physics–Informed Neural Networks: Where we are and What’s Next}, 
      author={Cuomo, Salvatore and Schiano Di Cola, Vincenzo and Giampaolo, Fabio  and Rozza, Gianluigi and Raissi, Maziar  and Piccialli, Francesco},
      journal={Journal of Scientific Computing},
      volume={92},
      issue={88},
      pages={},
      year={2022},
doi={https://doi.org/10.1007/s10915-022-01939-z}
}

@article{Physics-2022-Dhara,
      title={Physics-guided deep autoencoder to overcome the need for a starting model in full-waveform inversion}, 
      author={Dhara, Arnab and Sen, Mrinal},
      journal={The Leading Edge},
      volume={41},
      issue={6},
      pages={375-381},
      year={2022},
doi={https://doi.org/10.1190/tle41060375.1}
}

@article{Physics-2022-Behesht,
      title={Physics-Informed Neural Networks ({PINNs}) for Wave Propagation and Full Waveform Inversions}, 
      author={Rasht-Behesht, Majid and Huber, Christian  and Shukla, Khemraj and Em Karniadakis, George},
      journal={Journal of Geophysical Research: Solid Earth},
      volume={127},
      issue={5},
      pages={e2021JB023120},
      year={2022},
doi={https://doi.org/10.1029/2021JB023120}
}

@inproceedings{Feng-2022-Exploring,
  title={Extremely weak supervision inversion of multiphysical properties},
  author={Feng, Shihang and Jin, Peng and Zhang, Xitong and Chen, Yinpeng and Alumbaugh, David and Commer, Michael and Lin, Youzuo},
  booktitle={Second International Meeting for Applied Geoscience \& Energy},
  pages={1785--1789},
  year={2022},
  organization={Society of Exploration Geophysicists and American Association of Petroleum},
doi={https://doi.org/10.1190/image2022-3746487.1}
}

@ARTICLE{Um-2022-Deep,
  TITLE      = {Deep-Learning Multiphysics Network for Imaging {CO}$_2$ Saturation and Estimating Uncertainty in Geological Carbon Storage},
  AUTHOR     = {Schankee Um, Evan and Alumbaugh, David and Commer, Michael and Feng, Shihang and Gasperikova, Erika and Li, Yaoguo and Lin, Youzuo  and Samarasinghe, Savini},
  JOURNAL    = {Geophysical Prospecting},
volume={72},
  number={Machine learning applications in geophysical exploration and monitoring},
  pages={183--198},
  year={2023},
doi={https://doi.org/10.1111/1365-2478.13257}
}

@inproceedings{Feng-2022-Intriguing,
  author =	 {Feng, Yinan and Chen, Yinpeng and Feng, Shihang and Jin, Peng and Liu, Zicheng and Lin, Youzuo},
  title =	 {An Intriguing Property of Geophysics Inversion},
  booktitle =	 {The Thirty-ninth International Conference on Machine Learning~(ICML)},
  year =	 {2022},

doi={https://doi.org/10.48550/arXiv.2204.13731}
}

@ARTICLE{Um-2022-Real,
  TITLE      = {Real-time deep-learning inversion of seismic full waveform data for {CO}$_2$ saturation and uncertainty in geological carbon storage monitoring},
  AUTHOR     = {Schankee Um, Evan and Alumbaugh, David and Lin, Youzuo  and Feng, Shihang},
journal={Geophysical Prospecting},
  volume={72},
  number={Machine learning applications in geophysical exploration and monitoring},
  pages={199--212},
  year={2023},
  publisher={European Association of Geoscientists \& Engineers},
doi={ https://doi.org/10.1111/1365-2478.13197}
}

@ARTICLE{Osher-1992-Nonlinear,
  TITLE      = {Nonlinear {Total Variation} Based Noise Removal Algorithms},
  AUTHOR     = {Rudin, L. and Osher, S. and Fatemi, E.},
  JOURNAL    = {Physica D.},
  YEAR       = {1992},
  VOLUME     = {60},
  PAGES      = {259-268},
 doi={https://doi.org/10.1016/0167-2789(92)90242-F}
}

@phdthesis{Agudo-2018-Acoustic,
  author  = {Agudo, \`{O}scar},
  title   = {Acoustic full-waveform inversion in geophysical and medical imaging},
  school  = {Imperial College London},
  year    = {2018},
doi={https://doi.org/10.1093/gji/ggad158}
}

@article{Efficient-2021-Jin,
      title={Efficient Progressive Transfer Learning for Full-Waveform Inversion With Extrapolated Low-Frequency Reflection Seismic Data}, 
      author={Jin, Yuchen and Hu, Wenyi and Wang, Shirui and Zi, Yuan and Wu, Xuqing and Chen, Jiefu },
      journal={IEEE Transactions on Geoscience and Remote Sensing},
      volume={80},
      year={2021}
}

@article{Progressive-2021-Hu,
      title={Progressive transfer learning for low-frequency data prediction in full-waveform inversion}, 
      author={Hu, Wenyi and Jin, Yuchen and Wu, Xuqing and Chen, Jiefu },
      journal={Geophysics},
      volume={86},
      issue={4},
      pages={1JA -- X4},
      year={2021}
}

@article{Extrapolated-2020-Sun,
      title={Extrapolated full-waveform inversion with deep learning}, 
      author={Sun, Hongyu and Demanet, Laurent},
      journal={Geophysics},
      volume={85},
      issue={3},
      pages={1MJ -- Z13},
      year={2020}
}

@inproceedings{EFWI-2023-Feng,
  author =	 {Feng, Shihang and Wang, Hanchen and Deng, Chengyuan and Feng, Yinan  and Liu, Yanhua and Zhu, Min and Jin, Peng and Chen, Yinpeng and Lin, Youzuo},
  title =	 {$\mathbf{E^{FWI}}$: Multi-parameter Benchmark Datasets for Elastic Full Waveform Inversion of Geophysical Properties},
  booktitle =	 {Advances in Neural Information Processing Systems 35 (NeurIPS 2022) Datasets and Benchmarks Track},
  howpublished = {\url{https://efwi-lanl.github.io/}},
  year =	 {2023},
doi={
https://doi.org/10.48550/arXiv.2306.12386}
}

@inproceedings{OpenFWI-2022-Deng,
  author =	 {Deng, Chengyuan and Feng, Yinan and Feng, Shihang and Jin, Peng and Zhang, Xitong and Zeng, Qili and Lin, Youzuo},
  title =	 {{$\mathbf{OpenFWI}$}: Benchmark Seismic Datasets for Machine Learning-Based Full Waveform Inversion},
  booktitle =	 {Advances in Neural Information Processing Systems 35 (NeurIPS 2022) Datasets and Benchmarks Track},
  howpublished = {\url{https://openfwi-lanl.github.io/}},
  year =	 {2022},
doi={
https://doi.org/10.48550/arXiv.2111.02926
}
}

@inproceedings{Deep-2020-Ulyanov,
  title={Deep image prior},
  author={Ulyanov, Dmitry and Vedaldi, Andrea and Lempitsky, Victor},
  booktitle={Proceedings of the IEEE conference on computer vision and pattern recognition},
  pages={9446--9454},
  year={2018},
doi={
https://doi.org/10.1007/s11263-020-01303-4}

}

@article{mosser2020stochastic,
  title={Stochastic seismic waveform inversion using generative adversarial networks as a geological prior},
  author={Mosser, Lukas and Dubrule, Olivier and Blunt, Martin J},
  journal={Mathematical Geosciences},
  volume={52},
  number={1},
  pages={53--79},
  year={2020},
doi={https://doi.org/10.1007/s11004-019-09832-6}
}

@article{wu2019parametric,
  title={Parametric convolutional neural network-domain full-waveform inversion},
  author={Wu, Yulang and McMechan, George A},
  journal={Geophysics},
  volume={84},
  number={6},
  pages={R881--R896},
  year={2019},
  publisher={Society of Exploration Geophysicists},
doi={https://doi.org/10.1190/geo2018-0224.1}
}

@article{he2021reparameterized,
  title={Reparameterized full-waveform inversion using deep neural networks},
  author={He, Qinglong and Wang, Yanfei},
  journal={Geophysics},
  volume={86},
  number={1},
  pages={V1--V13},
  year={2021},
  publisher={Society of Exploration Geophysicists},
doi={https://doi.org/10.1190/geo2019-0382.1}
}

@article{Physics-2020-Ren,
 title={A Physics-Based Neural-Network Way to Perform Seismic Full Waveform Inversion}, 
 author={Ren, Yuxiao and Xu, Xinji and Yang, Senlin and Nie, Lichao and Chen, Yangkang},
 journal={IEEE Access},
 year = {2020},
 volume = {8},
 pages = {112266 -- 112277},
doi={https://doi.org/10.1109/ACCESS.2020.2997921}
}

@article{Integrating-2021-Zhu,
 title={Integrating deep neural networks with full-waveform inversion: Reparameterization, regularization, and uncertainty quantification}, 
 author={Zhu, Weiqiang and Xu, Kailai and Darve, Eric and Biondi, Biondo and Beroza, Gregory},
  journal={Geophysics},
  volume={87},
  number={1},
  pages={R93--R109},
  year={2022},
  publisher={Society of Exploration Geophysicists},
doi={https://doi.org/10.1190/geo2020-0933.1}
}

@article{Physics-2019-Raissi,
 title={Physics-informed neural networks: A deep learning framework for solving forward and inverse problems involving nonlinear partial differential equations}, author={Raissi, Maziar and Perdikaris, Paris and Karniadakis, George},
 journal={Journal of Computational Physics},
 year = {2019},
 volume = {378},
 pages = {686 -- 707},
 issue = {1},
doi={https://doi.org/10.1016/j.jcp.2018.10.045}
}

@article{Tromp-2020-Seismic,
  title={Seismic wavefield imaging of Earth’s interior across scales},
  author={Tromp, Jeroen},
  journal={Nature Review Earth and Environment},
  volume={1},
  pages={40--53},
  year={2020},
doi={https://doi.org/10.1038/s43017-019-0003-8}
}

@article{zhang2020data,
  title={Data-driven seismic waveform inversion: A study on the robustness and generalization},
  author={Zhang, Zhongping and Lin, Youzuo},
  journal={IEEE Transactions on Geoscience and Remote sensing},
  volume={58},
  number={10},
  pages={6900--6913},
  year={2020},
  publisher={IEEE},
doi={https://doi.org/10.1109/TGRS.2020.2977635}

}

@ARTICLE{Bunks-1995-Multiscale,
  TITLE      = {Multiscale Seismic Waveform Inversion},
  AUTHOR     = {Bunks, C. and Saleck, F. and Zaleski, S. and Chavent, G.},
  JOURNAL    = {Geophysics},
  YEAR       = {1995},
  VOLUME     = {60},
  NUMBER     = {5},
  PAGES      = {1457--1473},
doi={https://doi.org/10.1190/1.1443880}
}

@inproceedings{Jin-2021-Unsupervised,
  author =	 {Jin, Peng and Zhang, Xitong and Chen, Yinpeng  and Huang, Sharon and Liu, Zicheng and Lin, Youzuo},
  title =	 {Unsupervised Learning of Full-Waveform Inversion: Connecting {CNN} and Partial Differential Equation in a Loop},
  booktitle =	 {The Tenth International Conference on Learning Representations~(ICLR)},
  year =	 {2022},
doi={
https://doi.org/10.48550/arXiv.2110.07584}
}

@ARTICLE{feng2021multiscale,
  author={Feng, Shihang and Lin, Youzuo and Wohlberg, Brendt},
  journal={IEEE Transactions on Geoscience and Remote Sensing}, 
  title={Multiscale Data-Driven Seismic Full-Waveform Inversion With Field Data Study}, 
  year={2021},
  volume={60},
  number={},
  pages={1-14},
doi={https://doi.org/10.1109/TGRS.2021.3114101}

}

@article{Yang-2022-Making,
author= {Yang, Yuxin and Zhang, Xitong and Guan, Qiang  and Lin, Youzuo },
title = {Making Invisible Visible: Data-Driven Seismic Inversion with Spatio-temporally Constrained Data Augmentation},
journal   = {IEEE Transactions on Geoscience and Remote Sensing},
volume    = {60},
year      = {2022},
doi={https://doi.org/10.1109/TGRS.2022.3144636}

}

@article{Renan-2022-Physics,
author={Rojas-G\'omez, Ren\'an and Yang, Jihyun and Lin, Youzuo  and Theiler, James and Wohlberg, Brendt},
journal={IEEE Geoscience and Remote Sensing Letters}, 
title={Physics-Consistent Data-Driven Waveform Inversion With Adaptive Data Augmentation}, 
year={2022},
volume={19},
number={},
pages={1-5},
doi={https://doi.org/10.1109/LGRS.2020.3022021}
}

@article{araya2018deep,
  title={Deep-learning tomography},
  author={Araya-Polo, Mauricio and Jennings, Joseph and Adler, Amir and Dahlke, Taylor},
  journal={The Leading Edge},
  volume={37},
  number={1},
  pages={58--66},
  year={2018},
  publisher={Society of Exploration Geophysicists},
doi={https://doi.org/10.1190/tle37010058.1}
}

@article{wu-2019-inversionnet,
  title={{InversionNet}: An Efficient and Accurate Data-Driven Full Waveform Inversion},
  author={Wu, Yue and Lin, Youzuo},
  journal={IEEE Transactions on Computational Imaging},
  volume={6},
  pages={419--433},
  year={2019},
  publisher={IEEE},
  doi={https://doi.org/10.1109/TCI.2019.2956866}
}

@Article{Physics-2021-Karniadakis,
  Title                    = {Physics-informed machine learning},
  Author                   = {Karniadakis, George and Kevrekidis, Ioannis and Lu, Lu and Perdikaris, Paris and Wang, Sifan and Yang, Liu},
  Journal                  = {Nature Reviews Physics},
  Year                     = {2021},
  Pages                    = {422-440},
  Volume                   = {3},
  Issue                    = {},
doi={https://doi.org/10.1038/s42254-021-00314-5}
}

@Article{Integrating-2020-Willard,
  Title                    = {Integrating Scientific Knowledge with Machine Learning for Engineering and Environmental Systems},
  Author                   = {Willard, Jared and Jia, Xiaowei and Xu, Shaoming and Steinbach, Michael and Kumar, Vipin},
  Journal                  = {arXiv preprint},
  Year                     = {2020},
  Pages                    = {},
  Issue                    = {},
doi={
https://doi.org/10.48550/arXiv.2003.04919
}
}

@Article{Deep-2021-Adler,
  Title                    = {Deep Learning for Seismic Inverse Problems: Toward the Acceleration of Geophysical Analysis Workflows},
  Author                   = {Adler, Amir and Araya-Polo, Mauricio and Poggio, Tomaso},
  Journal                  = {IEEE Signal Processing Magazine},
  Year                     = {2021},
  Volume                   = {38},
  Issue                    = {2},
  Pages                    = {89-119},
doi={https://doi.org/10.1109/MSP.2020.3037429}
}

@Article{Deep-2021-Yu,
  Title                    = {Deep Learning for Geophysics: Current and Future Trends},
  Author                   = {Yu, Xiwei and Ma, Jianwei},
  Journal                  = {Reviews of Geophysics},
  Year                     = {2021},
  Pages                    = {e2021RG000742},
  Volume                   = {59},
  Issue                    = {3},
doi={https://doi.org/10.1029/2021RG000742}
}

@Article{Deep-2020-Gregory,
  Title                    = {Deep Learning Techniques for Inverse Problems in Imaging},
  Author                   = {Ongie, Gregory and Jalal, Ajil and Metzler, Christopher and Baraniuk, Richard and Dimakis, Alexandros and Willett, Rebecca},
  Journal                  = {IEEE Journal on Selected Areas in Information Theory},
  Year                     = {2020},
  Pages                    = {39 - 56},
  Volume                   = {1},
  Issue                    = {1},
doi={https://doi.org/10.1093/gji/ggad158}
}

@Article{Acoustic-2015-Lin,
  Title                    = {Acoustic- and elastic-waveform inversion using a modified {Total-Variation} regularization scheme},
  Author                   = {Lin, Youzuo and Huang, Lianjie},
  Journal                  = {Geophysical Journal International},
  Year                     = {2015},
  Pages                    = {489-502},
  Volume                   = {200},
  Issue                    = {1},
doi={https://doi.org/10.1093/gji/ggu393}
}

@ARTICLE{Virieux-2009-Overview,
  TITLE      = {An Overview of Full-waveform Inversion in Exploration Geophysics},
  AUTHOR     = {Virieux, Jean and Operto, Stephane},
  JOURNAL    = {Geophysics},
  YEAR       = {2009},
  VOLUME     = {74},
  NUMBER     = {6},
  PAGES      = {WCC1--WCC26},
doi={https://doi.org/10.1190/1.3238367}
}

@article{yang2019deep,
  title={Deep-learning inversion: A next-generation seismic velocity model building method},
  author={Yang, Fangshu and Ma, Jianwei},
  journal={Geophysics},
  volume={84},
  number={4},
  pages={R583--R599},
  year={2019},
  publisher={Society of Exploration Geophysicists},
doi={https://doi.org/10.1190/geo2018-0249.1}
}

@article{vedula2017towards,
  title={Towards {CT}-quality ultrasound imaging using deep learning},
  author={Vedula, Sanketh and Senouf, Ortal and Bronstein, Alex M and Michailovich, Oleg V and Zibulevsky, Michael},
  journal={arXiv preprint},
  year={2017},
doi={
https://doi.org/10.48550/arXiv.1710.06304
}
}

@article{dai2021self,
  title={Self-supervised learning for accelerated {3D} high-resolution ultrasound imaging},
  author={Dai, Xianjin and Lei, Yang and Wang, Tonghe and Axente, Marian and Xu, Dong and Patel, Pretesh and Jani, Ashesh B and Curran, Walter J and Liu, Tian and Yang, Xiaofeng},
  journal={Medical Physics},
  volume={48},
  number={7},
  pages={3916--3926},
  year={2021},
  publisher={Wiley Online Library},
doi={ https://doi.org/10.1002/mp.14946}
}

@article{fan2022model,
  title={Model-data-driven image reconstruction with neural networks for ultrasound computed tomography breast imaging},
  author={Fan, Yuling and Wang, Hongjian and Gemmeke, Hartmut and Hopp, Torsten and Hesser, Juergen},
  journal={Neurocomputing},
  volume={467},
  pages={10--21},
  year={2022},
  publisher={Elsevier},
doi={https://doi.org/10.1016/j.neucom.2021.09.035}
}

@inproceedings{leong2022estimating,
  title={Estimating {CO}$_2$ saturation maps from seismic data using deep convolutional neural networks},
  author={Leong, Zi Xian and Zhu, Tieyuan and Sun, Alexander Y},
  booktitle={Second International Meeting for Applied Geoscience \& Energy},
  pages={510--514},
  year={2022},
  organization={Society of Exploration Geophysicists and American Association of Petroleum},
doi={https://doi.org/10.1190/image2022-3746727.1}
}

@article{zhong2020inversion,
  title={Inversion of time-lapse seismic reservoir monitoring data using {CycleGAN}: A deep learning-based approach for estimating dynamic reservoir property changes},
  author={Zhong, Zhi and Sun, Alexander Y and Wu, Xinming},
  journal={Journal of Geophysical Research: Solid Earth},
  volume={125},
  number={3},
  pages={e2019JB018408},
  year={2020},
  publisher={Wiley Online Library},
doi={https://doi.org/10.1029/2019JB018408}
}

@article{moczo2007finite,
  title={The finite-difference time-domain method for modeling of seismic wave propagation},
  author={Moczo, Peter and Robertsson, Johan OA and Eisner, Leo},
  journal={Advances in geophysics},
  volume={48},
  pages={421--516},
  year={2007},
  publisher={Elsevier},
  doi={https://doi.org/10.1016/S0065-2687(06)48008-0}
}

@article{graves1996simulating,
  title={Simulating seismic wave propagation in {3D} elastic media using staggered-grid finite differences},
  author={Graves, Robert W},
  journal={Bulletin of the seismological society of America},
  volume={86},
  number={4},
  pages={1091--1106},
  year={1996},
  publisher={The Seismological Society of America},
  doi={https://doi.org/10.1785/BSSA0860041091}
}

@article{fornberg1988pseudospectral,
  title={The pseudospectral method; accurate representation of interfaces in elastic wave calculations},
  author={Fornberg, Bengt},
  journal={Geophysics},
  volume={53},
  number={5},
  pages={625--637},
  year={1988},
  publisher={Society of Exploration Geophysicists},
  doi={https://doi.org/10.1190/1.1442497}
}

@article{kuhlemeyer1973finite,
  title={Finite element method accuracy for wave propagation problems},
  author={Kuhlemeyer, Roger L and Lysmer, John},
  journal={Journal of the Soil Mechanics and Foundations Division},
  volume={99},
  number={5},
  pages={421--427},
  year={1973},
  publisher={American Society of Civil Engineers},
  doi={https://doi.org/10.1061/JSFEAQ.000188}
}

@article{kelly1976synthetic,
  title={Synthetic seismograms: A finite-difference approach},
  author={Kelly, Kenneth R and Ward, Ronald W and Treitel, Sven and Alford, Richard M},
  journal={Geophysics},
  volume={41},
  number={1},
  pages={2--27},
  year={1976},
  publisher={Society of Exploration Geophysicists},
  doi={https://doi.org/10.1190/1.1440605}
}

@article{virieux1984sh,
  title={S{H}-wave propagation in heterogeneous media: Velocity-stress finite-difference method},
  author={Virieux, Jean},
  journal={Geophysics},
  volume={49},
  number={11},
  pages={1933--1942},
  year={1984},
  doi={https://doi.org/10.1190/1.1441605},
  publisher={Society of Exploration Geophysicists}
}

@article{virieux1986p,
  title={P-{SV} wave propagation in heterogeneous media: Velocity-stress finite-difference method},
  author={Virieux, Jean},
  journal={Geophysics},
  volume={51},
  number={4},
  pages={889--901},
  year={1986},
  doi={https://doi.org/10.1190/1.1442147},
  publisher={Society of Exploration Geophysicists}
}

@article{kosloff1982forward,
  title={Forward modeling by a Fourier method},
  author={Kosloff, Dan D and Baysal, Edip},
  journal={Geophysics},
  volume={47},
  number={10},
  pages={1402--1412},
  year={1982},
  publisher={Society of Exploration Geophysicists},
  doi={https://doi.org/10.1190/1.1441288}
}

@article{komatitsch2005spectral,
  title={The spectral-element method in seismology},
  author={Komatitsch, Dimitri and Tsuboi, Seiji and Tromp, Jeroen and Levander, A and Nolet, G},
  journal={Geophysical Monograph-American Geophysical Union},
  volume={157},
  pages={205},
  year={2005},
  publisher={AGU AMERICAN GEOPHYSICAL UNION},
  doi={https://doi.org/10.1029/157GM13}
}

@article{kudela2007modelling,
  title={Modelling of wave propagation in composite plates using the time domain spectral element method},
  author={Kudela, Pawe{\l} and {\.Z}ak, Arkadiusz and Krawczuk, Marek and Ostachowicz, Wies{\l}aw},
  journal={Journal of Sound and Vibration},
  volume={302},
  number={4-5},
  pages={728--745},
  year={2007},
  publisher={Elsevier},
  doi={https://doi.org/10.1016/j.jsv.2006.12.016}
}

@article{Chan2017,
  author =	 {S. H. Chan and X. Wang and O. A. Elgendy},
  journal =	 {IEEE Trans. Comp. Imag.},
  title =	 {Plug-and-Play {ADMM} for Image Restoration:
                  Fixed-Point Convergence and Applications},
  year =	 2017,
  volume =	 3,
  number =	 1,
  pages =	 {84-98},
  doi =		 {https://doi.org/10.1109/TCI.2016.2629286},
  month =	 Mar,
}

@ARTICLE{DabovBM3D07,
  author =	 {Dabov, K. and Foi, A. and Katkovnik, V. and
                  Egiazarian, K.},
  journal =	 {IEEE Transactions on Image Processing},
  title =	 {Image Denoising by Sparse {3-D} Transform-Domain
                  Collaborative Filtering},
  year =	 2007,
  volume =	 16,
  number =	 8,
  pages =	 {2080--2095},
  doi =		 {https://doi.org/10.1109/TIP.2007.901238},
}

@article{boyd2011distributed,
  title =	 {Distributed optimization and statistical learning
                  via the alternating direction method of multipliers},
  author =	 {Boyd, Stephen and Parikh, Neal and Chu, Eric and
                  Peleato, Borja and Eckstein, Jonathan},
  journal =	 {Foundations and Trends in Machine Learning},
  volume =	 3,
  number =	 1,
  year =	 2011,
  doi={http://doi.org/10.1561/2200000016}
}

@article{gabay1976dual,
  title =	 {A dual algorithm for the solution of nonlinear
                  variational problems via finite element
                  approximation},
  author =	 {Gabay, Daniel and Mercier, Bertrand},
  journal =	 {Computers \& Mathematics with Applications},
  volume =	 2,
  number =	 1,
  pages =	 {17--40},
  year =	 1976,
  publisher =	 {Elsevier},
doi={https://doi.org/10.1016/0898-1221(76)90003-1}
}

@article{glowinski1975approximation,
  title =	 {Sur {l'}approximation, par elements finis {d'}ordre
                  un, et la resolution, par penalisation-dualite
                  {d'}une classe de problemes de Dirichlet non
                  lineaires},
  author =	 {Glowinski, Roland and Marroco, A},
  journal =	 {ESAIM: Mathematical Modelling and Numerical
                  Analysis-Mod{\'e}lisation Math{\'e}matique et
                  Analyse Num{\'e}rique},
  volume =	 9,
  number =	 {R2},
  pages =	 {41--76},
  year =	 1975,

}

@ARTICLE{sreehari2016TCI,
  author =	 {S. Sreehari and S. V. Venkatakrishnan and
                  B. Wohlberg and G. T. Buzzard and L. F. Drummy and
                  J. P. Simmons and C. A. Bouman},
  journal =	 {IEEE Trans. Comp. Imag.},
  title =	 {Plug-and-Play Priors for Bright Field Electron
                  Tomography and Sparse Interpolation},
  year =	 2016,
  volume =	 2,
  number =	 4,
  pages =	 {408--423},
  month =	 Dec,
      doi        = {https://doi.org/10.1109/TCI.2016.2599778}

}

@inproceedings{venkatakrishnan2013,
  title =	 {Plug-and-Play Priors for Model Based Reconstruction},
  author =	 {Venkatakrishnan, Singanallur V and Bouman, Charles A
                  and Wohlberg, Brendt},
  booktitle =	 {IEEE Global Conf.\ Signal Process.\ and Inf.\
                  Process. ({GlobalSIP})},
  pages =	 {945--948},
  year =	 2013,
      doi        = {https://doi.org/10.1109/GlobalSIP.2013.6737048}

}

@book{beck2017first,
	author = {Beck, Amir},
	publisher = {SIAM},
	series = {{MOS}-{SIAM} {S}eries on {O}ptimization},
	title = {First-Order Methods in Optimization},
	year = 2017,
    doi={https://doi.org/10.1137/1.9781611974997}
    }

@article{kamilov-2023-plug,
author = {Ulugbek Kamilov and Charles A. Bouman and Gregery T. Buzzard and Brendt Wohlberg},
title = {Plug-and-Play Methods for Integrating Physical and Learned Models in Computational Imaging},
journal = {IEEE Signal Processing Magazine},
year = 2023,
month = Jan,
volume = 40,
number = 1,
doi = {https://doi.org/10.1109/MSP.2022.3199595},
pages = {85--97}
}

@article{parikh2014proximal,
  title={Proximal algorithms},
  author={Parikh, Neal and Boyd, Stephen and others},
  journal={Foundations and trends in Optimization},
  volume={1},
  number={3},
  pages={127--239},
  year=2014,
doi = {http://doi.org/10.1561/2400000003}

}

@InProceedings{ryu2019plugandplay,
  title = 	 {Plug-and-Play Methods Provably Converge with Properly Trained Denoisers},
  author =       {Ryu, Ernest and Liu, Jialin and Wang, Sicheng and Chen, Xiaohan and Wang, Zhangyang and Yin, Wotao},
  booktitle = 	 {Proceedings of the 36th International Conference on Machine Learning},
  pages = 	 {5546--5557},
  year = 	 2019,
  editor = 	 {Chaudhuri, Kamalika and Salakhutdinov, Ruslan},
  volume = 	 {97},
  month = 	 Jun,
  publisher =    {PMLR},
  url = 	 {https://proceedings.mlr.press/v97/ryu19a.html},
doi = {http://doi.org/10.48550/arXiv.1905.05406}

}

@article{zhang2017beyond,
  title={Beyond a {G}aussian denoiser: Residual learning of deep {CNN} for image denoising},
  author={Zhang, Kai and Zuo, Wangmeng and Chen, Yunjin and Meng, Deyu and Zhang, Lei},
  journal={IEEE Transactions on Image Processing},
  volume={26},
  number={7},
  pages={3142--3155},
  year=2017,
      doi        = {https://doi.org/10.1109/TIP.2017.2662206}

}

@ARTICLE{Zhang2022Plug,
  author={Zhang, Kai and Li, Yawei and Zuo, Wangmeng and Zhang, Lei and Van Gool, Luc and Timofte, Radu},
  journal={IEEE Transactions on Pattern Analysis and Machine Intelligence}, 
  title={{Plug-and-Play} Image Restoration With Deep Denoiser Prior}, 
  year={2022},
  volume={44},
  number={10},
  pages={6360-6376},
  doi={10.1109/TPAMI.2021.3088914}
}

@article{wu2021semi,
  title={Semi-supervised learning for seismic impedance inversion using generative adversarial networks},
  author={Wu, Bangyu and Meng, Delin and Zhao, Haixia},
  journal={Remote Sensing},
  volume={13},
  number={5},
  pages={909},
  year={2021},
  publisher={MDPI},
doi={ https://doi.org/10.3390/rs13050909}
}

@incollection{hu2020physics,
  title={Physics-guided self-supervised learning for low frequency data prediction in FWI},
  author={Hu, Wenyi and Jin, Yuchen and Wu, Xuqing and Chen, Jiefu},
  booktitle={SEG Technical Program Expanded Abstracts 2020},
  pages={875--879},
  year={2020},
  publisher={Society of Exploration Geophysicists},
doi={https://doi.org/10.1190/segam2020-3423396.1}
}

@inproceedings{wu2021cnn,
  title={C{NN}-based gradient-free multiparameter reflection full-waveform inversion},
  author={Wu, Yulang and McMechan, George A and Wang, Yanfei},
  booktitle={First International Meeting for Applied Geoscience \& Energy},
  pages={1369--1373},
  year={2021},
  organization={Society of Exploration Geophysicists},
doi={https://doi.org/10.1190/segam2021-3593600.1}
}

@inproceedings{liu2021deep,
  title={Deep Learning Ultrasound Computed Tomography with Sparse Transmissions},
  author={Liu, Zhaohui and Wang, Jiameng and Ding, Mingyue and Yuchi, Ming},
  booktitle={2021 IEEE International Ultrasonics Symposium (IUS)},
  pages={1--4},
  year={2021},
  organization={IEEE},
doi={https://doi.org/10.1109/IUS52206.2021.9593459}
}

@inproceedings{zhang2020self,
  title={Self-supervised learning of a deep neural network for ultrafast ultrasound imaging as an inverse problem},
  author={Zhang, Jingke and He, Qiong and Xiao, Yang and Zheng, Hairong and Wang, Congzhi and Luo, Jianwen},
  booktitle={2020 IEEE International Ultrasonics Symposium (IUS)},
  pages={1--4},
  year={2020},
  organization={IEEE},
doi={https://doi.org/10.1109/IUS46767.2020.9251533}
}

@inproceedings{liu2021ultrasound,
  title={Ultrasound computed tomography using physical-informed neural network},
  author={Liu, Xilun and Almekkawy, Mohamed},
  booktitle={2021 IEEE International Ultrasonics Symposium (IUS)},
  pages={1--4},
  year={2021},
  organization={IEEE},
doi={https://doi.org/10.1109/IUS52206.2021.9593314}
}

@article{lahivaara2018deep,
  title={Deep convolutional neural networks for estimating porous material parameters with ultrasound tomography},
  author={L{\"a}hivaara, Timo and K{\"a}rkk{\"a}inen, Leo and Huttunen, Janne MJ and Hesthaven, Jan S},
  journal={The Journal of the Acoustical Society of America},
  volume={143},
  number={2},
  pages={1148--1158},
  year={2018},
  publisher={AIP Publishing},
doi={https://doi.org/10.1121/1.5024341}

}

@article{wang2022ultrasonic,
  title={Ultrasonic guided wave imaging with deep learning: Applications in corrosion mapping},
  author={Wang, Xiaocen and Lin, Min and Li, Jian and Tong, Junkai and Huang, Xinjing and Liang, Lin and Fan, Zheng and Liu, Yang},
  journal={Mechanical Systems and Signal Processing},
  volume={169},
  pages={108761},
  year={2022},
  publisher={Elsevier},
doi={https://doi.org/10.1016/j.ymssp.2021.108761}
}

@inproceedings{zhang2021general,
  title={A general framework for inverse problem solving using self-supervised deep learning: validations in ultrasound and photoacoustic image reconstruction},
  author={Zhang, Jingke and He, Qiong and Wang, Congzhi and Liao, Hongen and Luo, Jianwen},
  booktitle={2021 IEEE International Ultrasonics Symposium (IUS)},
  pages={1--4},
  year={2021},
  organization={IEEE},
doi={https://doi.org/10.1109/IUS52206.2021.9593902}

}

@article{vamaraju2019unsupervised,
  title={Unsupervised physics-based neural networks for seismic migration},
  author={Vamaraju, Janaki and Sen, Mrinal K},
  journal={Interpretation},
  volume={7},
  number={3},
  pages={SE189--SE200},
  year={2019},
  publisher={Society of Exploration Geophysicists and American Association of Petroleum~…},
doi={https://doi.org/10.1190/INT-2018-0230.1}
}

@article{kazei2021mapping,
  title={Mapping full seismic waveforms to vertical velocity profiles by deep learning},
  author={Kazei, Vladimir and Ovcharenko, Oleg and Plotnitskii, Pavel and Peter, Daniel and Zhang, Xiangliang and Alkhalifah, Tariq},
  journal={Geophysics},
  volume={86},
  number={5},
  pages={R711--R721},
  year={2021},
  publisher={Society of Exploration Geophysicists},
doi={https://doi.org/10.1190/geo2019-0473.1}

}

@article{wang2023seismic,
  title={Seismic velocity inversion transformer},
  author={Wang, Hongzhou and Lin, Jun and Dong, Xintong and Lu, Shaoping and Li, Yue and Yang, Baojun},
  journal={Geophysics},
  volume={88},
  number={4},
  pages={R513--R533},
  year={2023},
  publisher={Society of Exploration Geophysicists},
doi={https://doi.org/10.1190/geo2022-0283.1}
}

@article{zhao2020ultrasound,
  title={Ultrasound transmission tomography image reconstruction with a fully convolutional neural network},
  author={Zhao, Wenzhao and Wang, Hongjian and Gemmeke, Hartmut and Van Dongen, Koen WA and Hopp, Torsten and Hesser, J{\"u}rgen},
  journal={Physics in Medicine \& Biology},
  volume={65},
  number={23},
  pages={235021},
  year={2020},
  publisher={IOP Publishing},
doi={https://doi.org/10.1088/1361-6560/abb5c3}
}

@article{tong2022deep,
  title={Deep learning inversion with supervision: A rapid and cascaded imaging technique},
  author={Tong, Junkai and Lin, Min and Wang, Xiaocen and Li, Jian and Ren, Jiahao and Liang, Lin and Liu, Yang},
  journal={Ultrasonics},
  volume={122},
  pages={106686},
  year={2022},
  publisher={Elsevier},
doi={https://doi.org/10.1016/j.ultras.2022.106686}
}

@article{Rudolph2021,
  author =	 { Rudolph, Marco and Wandt,  Bastian and Rosenhahn, Bodo},
  title =	 { {Same same but DifferNet}: Semi-supervised defect detection with normalizing flows },
  journal =	 {Proceedings of the IEEE/CVF Winter Conference on Applications of Computer Vision},
  volume =	 {},
  pages = {1907–1916},
  year = {2021},
doi={
https://doi.org/10.48550/arXiv.2008.12577
}
}

@article{Akcay2018,
  author =	 {Akcay, Samet and Atapour-Abarghouei,  Amir and Breckon, Toby},
  title =	 { GANomaly: Semi-supervised Anomaly Detection via Adversarial Training },
  journal =	 { Asian Conference on Computer Vision },
  volume =	 {},
  pages = {622-637},
  year = {2018},
doi={ https://doi.org/10.1007/978-3-030-20893-6_39}
}

@article{Sen2019MSSP,
  author =	 {Sen, Debarshi and Aghazadeh,  Amirali and Mousavi, Ali and Nagarajaiah, Satish and Baraniuk, Richard  and Dabak, Anand },
  title =	 {Data-driven semi-supervised and supervised learning algorithms for health monitoring of pipes },
  journal =	 { Mechanical Systems and Signal Processing },
  volume =	 {131},
  pages = {524-537},
  year = {2019},
doi={https://doi.org/10.1016/j.ymssp.2019.06.003}
}

@article{Bouzenad2019,
  author =	 { Bouzenad, Abd, Ennour and Mountassir,  Mahjoub and Yaacoubi, Slah  and Dahmene, Fethi and Koabaz, Mahmoud  and Buchheit , Buchheit  and  Ke , Weina },
  title =	 { A semi-supervised based k-means algorithm for optimal guided waves structural health monitoring: A case study },
  journal =	 { Inventions },
  volume =	 {4},
  pages = {17},
  year = {2019},
doi={https://doi.org/10.3390/inventions4010017}
}

@article{Jiang2023RESS,
  author =	 { Jiang, Shengyu and He,  Rui and Chen, Guoming and Zhu, Yuan and Shi, Jiaming and  Liu, Kang and Chang, Yuanjiang },
  title =	 { Semi-supervised health assessment of pipeline systems based on optical fiber monitoring },
  journal =	 { Reliability Engineering \& System Safety },
  volume =	 {230},
  pages = {108932},
  year = {2023},
doi={https://doi.org/10.1016/j.ress.2022.108932}
}

@article{Luiken2023,
  author =	 { Luiken, Nick and Ravasi,  Matteo },
  title =	 { A deep learning-based approach to increase efficiency in the acquisition of ultrasonic non-destructive testing datasets},
  journal =	 { Proceedings of the IEEE/CVF Conference on Computer Vision and Pattern Recognition },
  volume =	 {},
  pages = {3094-3102},
  year = {2023},
doi={https://doi.org/10.1109/CVPRW59228.2023.00311}
}

@article{Ryu2023,
  author =	 { Ryu, Seongcheol and Park,  Seong-Hyun and Jhang
, Kyung-Young },
  title =	 { Plastic properties estimation of aluminum alloys using machine learning of ultrasonic and eddy current data },
  journal =	 { NDT \& E International },
  volume =	 {137},
  pages = {102857},
  year = {2023},
doi={https://doi.org/10.1016/j.ndteint.2023.102857}
}

@article{Song2020,
  author =	 { Song, Homin and Yang,  Yongchao },
  title =	 { Super-resolution visualization of subwavelength defects via deep learning-enhanced ultrasonic beamforming: A proof-of-principle study},
  journal =	 { NDT \& E International},
  volume =	 {116},
  pages = {102344},
  year = {2020},
doi={https://doi.org/10.1016/j.ndteint.2020.102344}
}

@article{Wang2022MSSP,
  author =	 { Wang, Xiaocen and Li,  Jian and Wang, Dingpeng  and Huang, Xinjing and Liang, Lin and Tang , Zhifeng and Fan , Zheng and Liu , Yang },
  title =	 { Sparse ultrasonic guided wave imaging with compressive sensing and deep learning },
  journal =	 { Mechanical Systems and Signal Processing },
  volume =	 {178},
  pages = {109346},
  year = {2022},
doi={https://doi.org/10.1016/j.ymssp.2022.109346}
}

@article{Siljama2021JNE,
  author = {Siljama, Oskar and Koskinen,Tuomas and Jessen-Juhler, Oskari and Virkkunen, Iikka},
  title =	 { Automated flaw detection in multi-channel phased array ultrasonic data using machine learning },
  journal =	 { Journal of Nondestructive Evaluation },
  volume =	 {40},
  pages = {67},
  year = {2021},
doi={ https://doi.org/10.1007/s10921-021-00796-4}
}

@article{Shukla2020JNE, 
author =	 { Shukla, Khemraj and Clark,  Patricio and Blackshire James and Sparkman, Daniel and Karniadakis, George},
  title =	 { Physics-informed neural network for ultrasound nondestructive quantification of surface breaking cracks },
  journal =	 { Journal of Nondestructive Evaluation },
  volume =	 {39},
  pages = {1-20},
  year = {2020},
doi={https://doi.org/10.1007/s10921-020-00705-1}
}

@article{Prakash2023,
  author =	 {Prakash, Navya and Nieberl,  Dorothea and Mayer, Monika and Schuster, Alfons },
  title =	 { Learning defects from aircraft NDT data },
  journal =	 { NDT \& E International },
  volume =	 {138},
  pages = {102885},
  year = {2023},
doi={https://doi.org/10.1016/j.ndteint.2023.102885}
}

@article{Rachman2021,
  author =	 {Rachman, Andika and Zhang,  Tieling and Ratnayake, Chandima },
  title =	 { Applications of machine learning in pipeline integrity management: A state-of-the-art review },
  journal =	 {International journal of pressure vessels and piping },
  volume =	 {193},
  pages = {104471},
  year = {2021},
doi={https://doi.org/10.1016/j.ijpvp.2021.104471}
}

@article{Hong2021MSE,
  author =	 {Hong, Xiaobin and Huang,  Liuwei and Gong, Shifeng and Xiao, Guoquan},
  title =	 { Shedding damage detection of metal underwater pipeline external anticorrosive coating by ultrasonic imaging based on {HOG} + {SVM}},
  journal =	 {Journal of Marine Science and Engineering},
  volume =	 {9(4)},
  pages = {364},
  year = {2021},
doi={https://doi.org/10.3390/jmse9040364}
}

@article{Virupakshappa2019,
  author =	 { Virupakshappa, Kushal and Oruklu,  Erdal },
  title =	 { Unsupervised machine learning for ultrasonic flaw detection using gaussian mixture modeling, k-means clustering and mean shift clustering },
  journal =	 { IEEE International Ultrasonics Symposium (IUS)},
  volume =	 {91},
  pages = {647-649},
  year = {2019},
doi={https://doi.org/10.1109/ULTSYM.2019.8926078}

}

@article{Kraljevski2021IEEE,
  author =	 { Kraljevski, Ivan and Duckhorn,  Frank and Barth, Martin and Tschoepe, Constanze and Schubert, Frank and Wolff, Matthias },
  title =	 { Autoencoder-based Ultrasonic {NDT} of Adhesive Bonds },
  journal =	 {2021 IEEE Sensors},
  volume =	 {978},
  pages = {4},
  year = {2021},
doi={https://doi.org/10.1109/SENSORS47087.2021.9639864}

}

@article{Sawant2023ultrasonics,
  author =	 { Sawant, Shruti and Sethi,  Amit and Banerjee, Sauvik and Tallur, Siddharth},
  title =	 { Unsupervised learning framework for temperature compensated damage identification and localization in ultrasonic guided wave $SHM$ with transfer learning},
  journal =	 { Ultrasonics},
  volume =	 {130},
  pages = {106931},
  year = {2023},
doi={https://doi.org/10.1016/j.ultras.2023.106931}
}

@article{Rautela2022CS,
  author =	 {Rautela, Mahindra and Senthilnath ,  J. and Monaco, Ernesto },
  title =	 { Delamination prediction in composite panels using unsupervised-feature learning methods with wavelet-enhanced guided wave representations },
  journal =	 { Compos. Struct },
  volume =	 {291},
  pages = {115579},
  year = {2022},
doi={https://doi.org/10.1016/j.compstruct.2022.115579}
}

@article{Wang2023NETWORK,
  author =	 { Wang, Boyang and Saniie,  Jafar },
  title =	 { Massive ultrasonic data compression using wavelet packet transformation optimized by convolutional autoencoders },
  journal =	 { IEEE Transactions on Neural Networks and Learning Systems },
  volume =	 {34},
  pages = {1395-1405},
  year = {2021},
doi={https://doi.org/10.1109/TNNLS.2021.3105367}

}

@article{Sergio2022,
  author =	 { Sergio, Cantero-Chinchilla and Wilcox,  Paul D and Croxford, Anthony J },
  title =	 {Deep learning in automated ultrasonic {NDE} – Developments, axioms and opportunities},
  journal =	 {NDT \& E International},
  volume =	 {131},
  pages = {102703},
  year = {2022},
doi={https://doi.org/10.1016/j.ndteint.2022.102703}
}

@article{Modlinski_Energy_2015,
	series = {Special {Issue} devoted to {The} 12th {International} {Conference} on {Boiler} {Technology} ({ICBT} 2014) - {Current} {Issues} of {Construction} and {Operation} of {Boilers}},
	title = {A validation of computational fluid dynamics temperature distribution prediction in a pulverized coal boiler with acoustic temperature measurement},
	volume = {92},
	issn = {0360-5442},
	url = {https://www.sciencedirect.com/science/article/pii/S036054421500746X},
	doi = {https://doi.org/10.1016/j.energy.2015.05.124},
	language = {en},
	urldate = {2023-02-07},
	journal = {Energy},
	author = {Modliński, Norbert and Madejski, Pawel and Janda, Tomasz and Szczepanek, Krzysztof and Kordylewski, Wlodzimierz},
	month = dec,
	year = {2015},
	pages = {77--86},
}

@article{Mizutani_JJAP_2006,
	title = {Measurement of Temperature Distribution Using Acoustic Reflector Array},
	volume = {45},
	issn = {0021-4922, 1347-4065},
	url = {https://iopscience.iop.org/article/10.1143/JJAP.45.4516},
	doi = {https://doi.org/10.1143/JJAP.45.4516},
	language = {en},
	number = {5B},
	urldate = {2023-02-15},
	journal = {Japanese Journal of Applied Physics},
	author = {Mizutani, Koichi and Kawabe, Satoshi and Saito, Ikumi and Masuyama, Hiroyuki},
	month = may,
	year = {2006},
	pages = {4516--4520},
}

@article{Kudo_AP_2003,
	title = {Temperature Distribution in a Rectangular Space Measured by a Small Number of Transducers and Reconstructed from Reflected Sounds},
	volume = {42},
	issn = {0021-4922, 1347-4065},
	url = {https://iopscience.iop.org/article/10.1143/JJAP.42.3189},
	doi = {https://doi.org/10.1143/JJAP.42.3189},
	language = {en},
	number = {Part 1, No. 5B},
	urldate = {2023-02-15},
	journal = {Japanese Journal of Applied Physics},
	author = {Kudo, Kousuke and Mizutani, Koichi and Akagami, Terumichi and Murayama, Riichi},
	month = may,
	year = {2003},
	pages = {3189--3193},
}

@article{Lu_MST_2000,
	title = {Acoustic computer tomographic pyrometry for two-dimensional measurement of gases taking into account the effect of refraction of sound wave paths},
	volume = {11},
	issn = {0957-0233, 1361-6501},
	url = {https://iopscience.iop.org/article/10.1088/0957-0233/11/6/312},
	doi = {https://doi.org/10.1088/0957-0233/11/6/312},
	language = {en},
	number = {6},
	urldate = {2023-03-21},
	journal = {Measurement Science and Technology},
	author = {Lu, J and Wakai, K and Takahashi, S and Shimizu, S},
	month = jun,
	year = {2000},
	pages = {692--697},
}

@article{Zhang_ATE_2015,
	title = {Online monitoring of the two-dimensional temperature field in a boiler furnace based on acoustic computed tomography},
	volume = {75},
	issn = {1359-4311},
	url = {https://www.sciencedirect.com/science/article/pii/S1359431114009673},
	doi = {https://doi.org/10.1016/j.applthermaleng.2014.10.085},
	urldate = {2023-08-30},
	journal = {Applied Thermal Engineering},
	author = {Zhang, Shiping and Shen, Guoqing and An, Liansuo and Niu, Yuguang},
	month = jan,
	year = {2015},
	pages = {958--966},
}

@article{Greenhall_Arxiv_2023,
	title = {Measuring {Thermal} {Profiles} in {High} {Explosives} using {Neural} {Networks}},
	url = {http://arxiv.org/abs/2310.12260},
	urldate = {2023-10-26},
	journal = {arXiv},
	author = {Greenhall, John and Zerkle, David K. and Davis, Eric S. and Broilo, Robert and Pantea, Cristian},
	month = oct,
	year = {2023},
	doi = {https://doi.org/10.48550/arXiv.2310.12260},
}

@book{Mcgee_temperature_1988,
	title = {Principles and {Methods} of {Temperature} {Measurement}},
	isbn = {978-0-471-62767-8},
	language = {en},
	publisher = {John Wiley \& Sons},
	author = {McGee, Thomas D.},
	month = may,
	year = {1988},
	note = {Google-Books-ID: qfmS7g4JzjwC},
}

@article{alkhalifah2000acoustic,
  title={An acoustic wave equation for anisotropic media},
  author={Alkhalifah, Tariq},
  journal={Geophysics},
  volume={65},
  number={4},
  pages={1239--1250},
  year={2000},
  publisher={Society of Exploration Geophysicists},
doi={https://doi.org/10.1190/1.1444815}
}

@article{blanch1995modeling,
  title={Modeling of a constant q; methodology and algorithm for an efficient and optimally inexpensive viscoelastic technique},
  author={Blanch, Joakim O and Robertsson, Johan OA and Symes, William W},
  journal={Geophysics},
  volume={60},
  number={1},
  pages={176--184},
  year={1995},
doi={https://doi.org/10.1190/1.1443744},
}

@article{mulder2004comparison,
  title={A comparison between one-way and two-way wave-equation migration},
  author={Mulder, Wim A and Plessix, R-E},
  journal={Geophysics},
  volume={69},
  number={6},
  pages={1491--1504},
  year={2004},
  publisher={Society of Exploration Geophysicists},
doi={https://doi.org/10.1190/1.1836822}
}

@book{meyers2008mechanical,
  title={Mechanical behavior of materials},
  author={Meyers, Marc Andr{\'e} and Chawla, Krishan Kumar},
  year={2008},
  publisher={Cambridge university press},
doi={https://doi.org/10.1007/978-3-030-84927-6}
}

@article{levander1988fourth,
  title={Fourth-order finite-difference P-SV seismograms},
  author={Levander, Alan R},
  journal={Geophysics},
  volume={53},
  number={11},
  pages={1425--1436},
  year={1988},
  publisher={Society of Exploration Geophysicists},
doi={https://doi.org/10.1190/1.1442422}
}

@incollection{duveneck2008acoustic,
  title={Acoustic VTI wave equations and their application for anisotropic reverse-time migration},
  author={Duveneck, Eric and Milcik, Paul and Bakker, Peter M and Perkins, Colin},
  booktitle={SEG technical program expanded abstracts 2008},
  pages={2186--2190},
  year={2008},
doi={https://doi.org/10.1190/1.3059320}
}

@article{komatitsch2002spectralI,
  title={Spectral-element simulations of global seismic wave propagation~{I}. Validation},
  author={Komatitsch, Dimitri and Tromp, Jeroen},
  journal={Geophysical Journal International},
  volume={149},
  number={2},
  pages={390--412},
  year={2002},
  publisher={Blackwell Publishing Ltd Oxford, UK},
  doi={https://doi.org/10.1046/j.1365-246X.2002.01653.x}
}

@article{komatitsch2002spectralII,
  title={Spectral-element simulations of global seismic wave propagation~{II}. Three-dimensional models, oceans, rotation and self-gravitation},
  author={Komatitsch, Dimitri and Tromp, Jeroen},
  journal={Geophysical Journal International},
  volume={150},
  number={1},
  pages={303--318},
  year={2002},
  publisher={Blackwell Publishing Ltd Oxford, UK},
doi={https://doi.org/10.1046/j.1365-246X.2002.01716.x}
}

@incollection{blanch1995efficient,
  title={Efficient iterative viscoacoustic linearized inversion},
  author={Blanch, Joakim O and Symes, William W},
  booktitle={SEG Technical Program Expanded Abstracts 1995},
  pages={627--630},
  year={1995},
doi={https://doi.org/10.1190/1.1887406}
}

@article{liu2009new,
  title={A new time--space domain high-order finite-difference method for the acoustic wave equation},
  author={Liu, Yang and Sen, Mrinal K},
  journal={Journal of computational Physics},
  volume={228},
  number={23},
  pages={8779--8806},
  year={2009},
  publisher={Elsevier},
doi={https://doi.org/10.1016/j.jcp.2009.08.027}
}

@article{yu_temperature_2022,
	title = {A Review on Acoustic Reconstruction of Temperature Profiles: From Time Measurement to Reconstruction Algorithm},
	volume = {71},
	issn = {0018-9456, 1557-9662},
	shorttitle = {A {Review} on {Acoustic} {Reconstruction} of {Temperature} {Profiles}},
	url = {https://ieeexplore.ieee.org/document/9872057/},
	doi = {https://doi.org/10.1109/TIM.2022.3203097},
	language = {en},
	urldate = {2023-11-21},
	journal = {IEEE Transactions on Instrumentation and Measurement},
	author = {Yu, Yang and Xiong, Qingyu and Ye, Zhi-Sheng and Liu, Xingchen and Li, Qiude and Wang, Kai},
	year = {2022},
	pages = {1--24},
}

@data{verdin_harvard_2021,
author = {Verdin, Benoit and Chevallier, Gaël and Ramasso, Emmanuel},
publisher = {Harvard Dataverse},
title = {{ORION-AE: Multisensor acoustic emission datasets reflecting supervised untightening of bolts in a jointed vibrating structure}},
year = {2021},
version = {DRAFT VERSION},
doi = {https://doi.org/10.7910/DVN/FBRDU0},
url = {https://doi.org/10.7910/DVN/FBRDU0}
}

@inproceedings{salamon_icm_2014,
author = {Salamon, Justin and Jacoby, Christopher and Bello, Juan Pablo},
title = {A Dataset and Taxonomy for Urban Sound Research},
year = {2014},
isbn = {9781450330633},
url = {https://doi.org/10.1145/2647868.2655045},
doi = {https://doi.org/10.1145/2647868.2655045},
booktitle = {Proceedings of the 22nd ACM International Conference on Multimedia},
pages = {1041–1044},
numpages = {4},
location = {Orlando, Florida, USA},
series = {MM '14}
}

@misc{virkkunen_arxiv_2019,
      title={Augmented Ultrasonic Data for Machine Learning}, 
      author={Iikka Virkkunen and Tuomas Koskinen and Oskari Jessen-Juhler and Jari Rinta-Aho},
      year={2019},
      archivePrefix={arXiv},
      primaryClass={eess.SP},
doi={https://doi.org/10.1007/s10921-020-00739-5}
}

@article{moleroarmenta_compphyscomm_2014,
title = {Optimized OpenCL implementation of the Elastodynamic Finite Integration Technique for viscoelastic media},
journal = {Computer Physics Communications},
volume = {185},
number = {10},
pages = {2683-2696},
year = {2014},
issn = {0010-4655},
doi = {https://doi.org/10.1016/j.cpc.2014.05.016},
url = {https://www.sciencedirect.com/science/article/pii/S0010465514001702},
author = {M. Molero-Armenta and Ursula Iturrarán-Viveros and S. Aparicio and M.G. Hernández},
}

@book{gaussorgues_infrared_1993,
	title = {Infrared {Thermography}},
	isbn = {978-0-412-47900-7},
	language = {en},
	publisher = {Springer Science \& Business Media},
	author = {Gaussorgues, G. and Chomet, S.},
	month = dec,
	year = {1993},
	note = {Google-Books-ID: FRQOiHeCXgMC},
}

@article{komatitsch2010high,
  title={High-order finite-element seismic wave propagation modeling with MPI on a large GPU cluster},
  author={Komatitsch, Dimitri and Erlebacher, Gordon and G{\"o}ddeke, Dominik and Mich{\'e}a, David},
  journal={Journal of computational physics},
  volume={229},
  number={20},
  pages={7692--7714},
  year={2010},
  publisher={Elsevier},
  doi={https://doi.org/10.1016/j.jcp.2010.06.024}
}

@article{de2009new,
  title={New developments in the finite-element method for seismic modeling},
  author={De Basabe, Jonas D and Sen, Mrinal K},
  journal={The Leading Edge},
  volume={28},
  number={5},
  pages={562--567},
  year={2009},
  publisher={Society of Exploration Geophysicists},
  doi={https://doi.org/10.1190/1.3124931}
}

@article{Li20213Dstochastic,
  author={Li, Fu and Villa, Umberto and Park, Seonyeong and Anastasio, Mark A.},
  journal={IEEE Transactions on Ultrasonics, Ferroelectrics, and Frequency Control}, 
  title={{3-D} Stochastic Numerical Breast Phantoms for Enabling Virtual Imaging Trials of Ultrasound Computed Tomography}, 
  year={2022},
  volume={69},
  number={1},
  pages={135-146},
  doi={https://doi.org/10.1109/TUFFC.2021.3112544}
}

@article{lucka21,
  title={High resolution {3D} ultrasonic breast imaging by time-domain full waveform inversion},
  author={Lucka, Felix and P{\'e}rez-Liva, Mailyn and Treeby, Bradley E and Cox, Ben T},
  journal={Inverse Problems},
  volume={38},
  number={2},
  pages={025008},
  year={2021},
  publisher={IOP Publishing},
 doi={https://doi.org/10.1088/1361-6420/ac3b64}
}

@inproceedings{zhang12,
  title={Efficient implementation of ultrasound waveform tomography using source encoding},
  author={Zhang, Zhigang and Huang, Lianjie and Lin, Youzuo},
  booktitle={Medical Imaging 2012: Ultrasonic Imaging, Tomography, and Therapy},
  volume={8320},
  pages={22--31},
  year={2012},
  organization={SPIE},
doi={https://doi.org/10.1117/12.910969}
}

@article{Li2023forward,
  author={Li, Fu and Villa, Umberto and Duric, Nebojsa and Anastasio, Mark A.},
  journal={IEEE Transactions on Ultrasonics, Ferroelectrics, and Frequency Control}, 
  title={A forward model incorporating elevation-focused transducer properties for 3D full-waveform inversion in ultrasound computed tomography}, 
  year={2023},
  volume={},
  number={},
  pages={1-1},
  doi={https://doi.org/10.1109/TUFFC.2023.3313549}
}

@data{li2021NBPs3D,
author = {Li, Fu and Villa, Umberto and Park, Seonyeong and Anastasio, Mark},
publisher = {Harvard Dataverse},
title = {{3D Acoustic Numerical Breast Phantoms}},
year = {2021},
version = {DRAFT VERSION},
doi = {https://doi.org/10.7910/DVN/KBYQQ7},
url = {https://doi.org/10.7910/DVN/KBYQQ7}
}

@data{li2021NBPs2D,
author = {Li, Fu and Villa, Umberto and Park, Seonyeong and Anastasio, Mark},
publisher = {Harvard Dataverse},
title = {{2D Acoustic Numerical Breast Phantoms and USCT Measurement Data}},
year = {2021},
version = {DRAFT VERSION},
doi = {https://doi.org/10.7910/DVN/CUFVKE},
url = {https://doi.org/10.7910/DVN/CUFVKE}
}

@misc{Fu2021code,
title={usct-breast-phantom: A python library to generate anatomically and physiologically realistic numerical breast phantoms for USCT virtual imaging studies},
author = {Li, Fu and Villa, Umberto and Park, Seonyeong and Anastasio, Mark},
year = {2021},
doi = {https://doi.org/10.5281/zenodo.5173070},
url = {https://doi.org/10.5281/zenodo.5173070}
}

@data{Li2023NBS3D,
author = {Li, Fu and Villa, Umberto},
publisher = {Harvard Dataverse},
title = {{3D Numerical Breast Phantoms and Ring-Array USCT measurements (3 rings)}},
year = {2023},
version = {V1},
doi = {https://doi.org/10.7910/DVN/8JVLAE},
url = {https://doi.org/10.7910/DVN/8JVLAE}
}

@inproceedings{Jeong2023deep,
author = {Jeong, Gangwon  and Li, Fu  and Villa, Umberto  and Anastasio, Mark A.},
title = {{A deep-learning-based image reconstruction method for USCT that employs multimodality inputs}},
volume = {12470},
booktitle = {Medical Imaging 2023: Ultrasonic Imaging and Tomography},
editor = {Christian Boehm and Nick Bottenus},
pages = {124700M},
year = {2023},
doi = {https://doi.org/10.1117/12.2654564},
}

@article{jeong2023investigating,
      title={Investigating the Use of Traveltime and Reflection Tomography for Deep Learning-Based Sound-Speed Estimation in Ultrasound Computed Tomography}, 
      author={Jeong, Gangwon and Li, Fu and Mitcham, Trevor and Duric,  Nebojsa and Villa, Umberto and Anastasio, Mark A.},
      year={2024},
      doi={10.1109/TUFFC.2024.3459391},
      journal={IEEE Transactions on Ultrasonics, Ferroelectrics, and Frequency Control},
      volume={early access}
}

@inproceedings{Poudel2019compensation,
author = {Joemini Poudel and Luca A. Forte and Mark A. Anastasio},
title = {{Compensation of 3D-2D model mismatch in ultrasound computed tomography with the aid of convolutional neural networks (Conference Presentation)}},
volume = {10955},
booktitle = {Medical Imaging 2019: Ultrasonic Imaging and Tomography},
editor = {Brett C. Byram and Nicole V. Ruiter},
pages = {1095507},
year = {2019},
doi = {https://doi.org/10.1117/12.2512966},
URL = {https://doi.org/10.1117/12.2512966}
}

@inproceedings{donaldson21,
  title={Instantaneous ultrasound computed tomography using deep convolutional neural networks},
  author={Donaldson, Robert W and He, Jiaze},
  booktitle={Health Monitoring of Structural and Biological Systems XV},
  volume={11593},
  pages={396--405},
  year={2021},
  organization={SPIE},
doi={https://doi.org/10.1117/12.2582630}
}

@article{stanziola2023learned,
  title={A learned {B}orn series for highly-scattering media},
  author={Stanziola, Antonio and Arridge, Simon and Cox, Ben T and Treeby, Bradley E},
  journal={JASA Express Letters},
  volume={3},
  number={5},
  year={2023},
  publisher={AIP Publishing},
doi={
https://doi.org/10.48550/arXiv.2212.04948
}
}

@book{american2013acr,
  title={ACR BI-RADS Atlas: Breast Imaging Reporting and Data System; Mammography, Ultrasound, Magnetic Resonance Imaging, Follow-up and Outcome Monitoring, Data Dictionary},
  author={American College of Radiology and D'Orsi, Carl J and others},
  year={2013},
  publisher={ACR, American College of Radiology}
}

@inproceedings{taskin20203d,
  title={{3D} redatuming for breast ultrasound},
  author={Taskin, Ulas and van Dongen, Koen WA},
  booktitle={Medical Imaging 2020: Physics of Medical Imaging},
  volume={11312},
  pages={113125H},
  year={2020},
  organization={International Society for Optics and Photonics},
doi={ 
https://doi.org/10.1117/12.2541131}
}

@article{javaherian2020refraction,
  title={Refraction-corrected ray-based inversion for three-dimensional ultrasound tomography of the breast},
  author={Javaherian, Ashkan and Lucka, Felix and Cox, Ben T},
  journal={Inverse Problems},
  volume={36},
  number={12},
  pages={125010},
  year={2020},
  publisher={IOP Publishing},
doi={https://doi.org/10.1088/1361-6420/abc0fc}
}

@article{wiskin2020full,
  title={Full wave {3D} inverse scattering transmission ultrasound tomography in the presence of high contrast},
  author={Wiskin, James and Malik, Bilal and Borup, David and Pirshafiey, Nasser and Klock, John},
  journal={Scientific Reports},
  volume={10},
  number={1},
  pages={1--14},
  year={2020},
  publisher={Nature Publishing Group},
doi={https://doi.org/10.1038/s41598-020-76754-3}
}

@article{duric2020using,
  title={Using whole breast ultrasound tomography to improve breast cancer risk assessment: a novel risk factor based on the quantitative tissue property of sound speed},
  author={Duric, Neb and Sak, Mark and Fan, Shaoqi and Pfeiffer, Ruth M and Littrup, Peter J and Simon, Michael S and Gorski, David H and Ali, Haythem and Purrington, Kristen S and Brem, Rachel F and others},
  journal={Journal of clinical medicine},
  volume={9},
  number={2},
  pages={367},
  year={2020},
  publisher={Multidisciplinary Digital Publishing Institute},
doi={https://doi.org/10.3390/jcm9020367}
}

@article{duric2018breast,
  title={Breast ultrasound tomography},
  author={Duric, Nebojsa and Littrup, Peter and Kuzmiak, CM},
  journal={Breast Imaging},
  volume={6},
  year={2018},
  publisher={IntechOpen},
doi={https://doi.org/10.5772/intechopen.69794}
}

@article{malik2016objective,
  title={Objective breast tissue image classification using Quantitative Transmission ultrasound tomography},
  author={Malik, Bilal and Klock, John and Wiskin, James and Lenox, Mark},
  journal={Scientific reports},
  volume={6},
  pages={38857},
  year={2016},
doi={https://doi.org/10.1038/srep38857},
  publisher={Nature Publishing Group}
}

@inproceedings{duric2013breast,
  title={Breast imaging with ultrasound tomography: Initial results with SoftVue},
  author={Duric, Neb and Littrup, Peter and Roy, Olivier and Schmidt, Steven and Li, Cuiping and Bey-Knight, Lisa and Chen, Xiaoyang},
  booktitle={2013 IEEE International Ultrasonics Symposium (IUS)},
  pages={382--385},
  year={2013},
doi={https://doi.org/10.1109/ULTSYM.2013.0099},
  organization={IEEE}
}

@article{ruiter20123d,
  title={{3D} ultrasound computer tomography of the breast: A new era?},
  author={Ruiter, Nicole V and Zapf, Michael and Hopp, Torsten and Dapp, Robin and Kretzek, Ernst and Birk, Matthias and Kohout, Benedikt and Gemmeke, Hartmut},
  journal={European Journal of Radiology},
  volume={81},
doi={https://doi.org/10.1016/S0720-048X(12)70055-4},
  pages={S133--S134},
  year={2012},
  publisher={Elsevier}
}

@article{badano2018evaluation,
  title={Evaluation of digital breast tomosynthesis as replacement of full-field digital mammography using an in silico imaging trial},
  author={Badano, Aldo and Graff, Christian G and Badal, Andreu and Sharma, Diksha and Zeng, Rongping and Samuelson, Frank W and Glick, Stephen J and Myers, Kyle J},
  journal={JAMA network open},
  volume={1},
  number={7},
  pages={e185474--e185474},
  year={2018},
  publisher={American Medical Association},
doi={https://doi.org/10.1001/jamanetworkopen.2018.5474}

}

@article{badano2021silico,
  title={In silico imaging clinical trials: cheaper, faster, better, safer, and more scalable},
  author={Badano, Aldo},
  journal={Trials},
  volume={22},
  number={1},
  pages={1--7},
  year={2021},
  publisher={Springer},
doi={https://doi.org/10.1186/s13063-020-05002-w}
}

@article{feigin2019deep,
  title={A deep learning framework for single-sided sound speed inversion in medical ultrasound},
  author={Feigin, Micha and Freedman, Daniel and Anthony, Brian W},
  journal={IEEE Transactions on Biomedical Engineering},
  volume={67},
  number={4},
  pages={1142--1151},
  year={2019},
  publisher={IEEE},
doi={https://doi.org/10.1109/TBME.2019.2931195}
}

@article{wang2015waveform,
  title={Waveform inversion with source encoding for breast sound speed reconstruction in ultrasound computed tomography},
  author={Wang, Kun and Matthews, Thomas and Anis, Fatima and Li, Cuiping and Duric, Neb and Anastasio, Mark A},
  journal={IEEE transactions on ultrasonics, ferroelectrics, and frequency control},
  volume={62},
  number={3},
  pages={475--493},
  year={2015},
  publisher={IEEE},
doi={https://doi.org/10.1109/TUFFC.2014.006788}
}

@inproceedings{pratt2007sound,
  title={Sound-speed and attenuation imaging of breast tissue using waveform tomography of transmission ultrasound data},
  author={Pratt, R Gerhard and Huang, Lianjie and Duric, Neb and Littrup, Peter},
  booktitle={Medical Imaging 2007: Physics of Medical Imaging},
  volume={6510},
  pages={65104S},
  year={2007},
  organization={International Society for Optics and Photonics},
doi={https://doi.org/10.1117/12.708789}
}

@article{tabei2002k,
  title={A k-space method for coupled first-order acoustic propagation equations},
  author={Tabei, Makoto and Mast, T Douglas and Waag, Robert C},
  journal={The Journal of the Acoustical Society of America},
  volume={111},
  number={1},
  pages={53--63},
  year={2002},
  publisher={Acoustical Society of America},
  doi={https://doi.org/10.1121/1.1421344}
}

@article{schreiman1984ultrasound,
  title={Ultrasound transmission computed tomography of the breast.},
  author={Schreiman, JS and Gisvold, JJ and Greenleaf, James F and Bahn, RC},
  journal={Radiology},
  volume={150},
  number={2},
  pages={523--530},
  year={1984},
doi={https://doi.org/10.1148/radiology.150.2.6691113}
}

@article{li2009vivo,
  title={In vivo breast sound-speed imaging with ultrasound tomography},
  author={Li, Cuiping and Duric, Nebojsa and Littrup, Peter and Huang, Lianjie},
  journal={Ultrasound in medicine \& biology},
  volume={35},
  number={10},
  pages={1615--1628},
  year={2009},
  publisher={Elsevier},
doi={https://doi.org/10.1016/j.ultrasmedbio.2009.05.011}
}

@article{duric2007detection,
  title={Detection of breast cancer with ultrasound tomography: First results with the Computed Ultrasound Risk Evaluation (CURE) prototype},
  author={Duric, Nebojsa and Littrup, Peter and Poulo, Lou and Babkin, Alex and Pevzner, Roman and Holsapple, Earle and Rama, Olsi and Glide, Carri},
  journal={Medical physics},
  volume={34},
  number={2},
  pages={773--785},
  year={2007},
  publisher={Wiley Online Library},
doi={https://doi.org/10.1118/1.2432161}
}

@article{sandhu2015frequency,
  title={Frequency domain ultrasound waveform tomography: breast imaging using a ring transducer},
  author={Sandhu, GY and Li, Cuiping and Roy, O and Schmidt, S and Duric, N},
  journal={Physics in Medicine \& Biology},
  volume={60},
  number={14},
  pages={5381},
  year={2015},
  publisher={IOP Publishing},
doi={https://doi.org/10.1088/0031-9155/60/14/5381}
}

@article{andre1997high,
  title={High-speed data acquisition in a diffraction tomography system employing large-scale toroidal arrays},
  author={Andr{\'e}, Michael P and Jan{\'e}e, Helmar S and Martin, Peter J and Otto, Gregory P and Spivey, Brett A and Palmer, Douglas A},
  journal={International Journal of Imaging Systems and Technology},
  volume={8},
  number={1},
  pages={137--147},
  year={1997},
  publisher={Wiley Online Library},
doi={https://doi.org/10.1002/(SICI)1098-1098(1997)8:1<137::AID-IMA15>3.0.CO;2-%23}
}

@article{carson1981breast,
  title={Breast imaging in coronal planes with simultaneous pulse echo and transmission ultrasound},
  author={Carson, Paul L and Meyer, Charles R and Scherzinger, Ann L and Oughton, Thomas V},
  journal={Science},
  volume={214},
  number={4525},
  pages={1141--1143},
  year={1981},
  publisher={American Association for the Advancement of Science},
doi={https://doi.org/10.1126/science.7302585}
}

@article{nam2013quantitative,
  title={Quantitative assessment of in vivo breast masses using ultrasound attenuation and backscatter},
  author={Nam, Kibo and Zagzebski, James A and Hall, Timothy J},
  journal={Ultrasonic imaging},
  volume={35},
  number={2},
  pages={146--161},
  year={2013},
  publisher={Sage Publications Sage CA: Los Angeles, CA},
doi={https://doi.org/10.1177/0161734613480281}
}

@inproceedings{greenleaf1977quantitative,
  title={QUANTITATIVE CROSS-SECTIONAL IMAGING OF ULTRASOUND PARAMETERS.},
  author={Greenleaf, James F and Johnson, SA and Bahn, Robert C and Rajagopalan, Balasubramanian},
  booktitle={Ultrason Symp Proc},
  year={1977},
doi={https://doi.org/10.1109/ULTSYM.1977.196985}
}

@ARTICLE{9424044,
  author={Bhadra, Sayantan and Kelkar, Varun A. and Brooks, Frank J. and Anastasio, Mark A.},
  journal={IEEE Transactions on Medical Imaging}, 
  title={On hallucinations in tomographic image reconstruction}, 
  year={2021},
  volume={},
  number={},
  pages={1-1},
  doi={https://doi.org/10.1109/TMI.2021.3077857}}

@inproceedings{li2024learning,
  title={A learning-based method for compensating 3D-2D model mismatch in ring-array ultrasound computed tomography},
  author={Li, Fu and Villa, Umberto and Anastasio, Mark A},
  booktitle={Medical Imaging 2024: Ultrasonic Imaging and Tomography},
  volume={12932},
  pages={328--334},
  year={2024},
  organization={SPIE},
doi={https://doi.org/10.1117/12.3006968}
}

@Article{Stuart10,
  Title                    = {Inverse problems: {A B}ayesian perspective},
  Author                   = {Andrew M. Stuart},
  Journal                  = {Acta Numerica},
  Year                     = {2010},
  Pages                    = {451-559},
  Volume                   = {19},
  Doi                      = {http://doi.org/10.1017/S0962492910000061}
}

@Book{KaipioSomersalo05,
  Title                    = {Statistical and Computational Inverse Problems},
  Author                   = {Kaipio, Jari and Somersalo, Erkki},
  Publisher                = {Springer-Verlag},
  Year                     = {2005},
  Address                  = {New York},
  Series                   = {Applied Mathematical Sciences},
  Volume                   = {160},
  Pages                    = {xvi+339},
doi={https://doi.org/10.1007/b138659}
}

@Book{Tarantola05,
  Title                    = {Inverse Problem Theory and Methods for Model Parameter Estimation},
  Author                   = {Tarantola, Albert},
  Publisher                = {SIAM},
  Year                     = {2005},
  Address                  = {Philadelphia, PA},
  Pages                    = {xii+342},
  doi={https://doi.org/10.1137/1.9780898717921}
}

@Book{BieglerBirosGhattasEtAl11,
  Title                    = {Large-scale Inverse Problems and Quantification of Uncertainty},
  Editor                   = {Larry Biegler and George Biros and Omar Ghattas and Youssef Marzouk and Matthias Heinkenschloss and David Keyes and Bani Mallick and Luis Tenorio and Bart van Bloemen Waanders and Karen Willcox},
  Publisher                = {Wiley},
  Year                     = {2011},
doi={https://doi.org/10.1002/9780470685853}
}

@Article{OdenMoserGhattas10,
  Title                    = {Computer Predictions with Quantified Uncertainty, {P}arts {I} \& {II}},
  Author                   = {J. Tinsley Oden and Robert M. Moser and Omar Ghattas},
  Journal                  = {SIAM News},
  Year                     = {2010},
  Number                   = {9\&10},
  Volume                   = {43},
doi={https://doi.org/10.1002/9781119176817.ecm2101}
}

@inproceedings{UlrichBoehmZuninoetal22,
author = {Ulrich, Ines E.  and  Boehm, Christian and Zunino, Andrea  and Fichtner, Andreas },
title = {{Analyzing resolution and model uncertainties for ultrasound computed tomography using Hessian information}},
volume = {12038},
booktitle = {Medical Imaging 2022: Ultrasonic Imaging and Tomography},
editor = {Nick Bottenus and Nicole V. Ruiter},
organization = {International Society for Optics and Photonics},
pages = {120380A},
year = {2022},
doi = {https://doi.org/10.1117/12.2608546},
URL = {https://doi.org/10.1117/12.2608546}
}

@article{bates2022probabilistic,
  title={A probabilistic approach to tomography and adjoint state methods, with an application to full waveform inversion in medical ultrasound},
  author={Bates, Oscar and Guasch, Lluis and Strong, George and Robins, Thomas Caradoc and Calderon-Agudo, Oscar and Cueto, Carlos and Cudeiro, Javier and Tang, Mengxing},
  journal={Inverse Problems},
  volume={38},
  number={4},
  pages={045008},
  year={2022},
  publisher={IOP Publishing},
doi = {https://doi.org/10.1088/1361-6420/ac55ee},


}

@article{BleiKucukelbirMcAuliffe17,
author = {Blei, David M. and  Kucukelbir, Alp and McAuliffe, Jon D. },
title = {Variational Inference: A Review for Statisticians},
journal = {Journal of the American Statistical Association},
volume = {112},
number = {518},
pages = {859--877},
year = {2017},
publisher = {Taylor \& Francis},
doi = {https://doi.org/10.1080/01621459.2017.1285773},




}

@article{ZhangCurtis20,
    author = {Zhang, Xin and Curtis, Andrew},
    title = "{Variational full-waveform inversion}",
    journal = {Geophysical Journal International},
    volume = {222},
    number = {1},
    pages = {406-411},
    year = {2020},
    month = {04},
    issn = {0956-540X},
    doi = {10.1093/gji/ggaa170},
    url = {https://doi.org/10.1093/gji/ggaa170},
}

@inproceedings{rezende2015variational,
  title={Variational inference with normalizing flows},
  author={Rezende, Danilo and Mohamed, Shakir},
  booktitle={International conference on machine learning},
  pages={1530--1538},
  year={2015},
doi={
https://doi.org/10.48550/arXiv.1505.05770
}
}

@article{papamakarios2021normalizing,
  title={Normalizing flows for probabilistic modeling and inference},
  author={Papamakarios, George and Nalisnick, Eric and Rezende, Danilo Jimenez and Mohamed, Shakir and Lakshminarayanan, Balaji},
  journal={Journal of Machine Learning Research},
  volume={22},
  number={57},
  pages={1--64},
  year={2021},
doi={
https://doi.org/10.48550/arXiv.1912.02762
}
}

@inproceedings{sun2021deep,
  title={Deep probabilistic imaging: Uncertainty quantification and multi-modal solution characterization for computational imaging},
  author={Sun, He and Bouman, Katherine L},
  booktitle={Proceedings of the AAAI Conference on Artificial Intelligence},
  volume={35},
  number={3},
  pages={2628--2637},
  year={2021},
doi={
https://doi.org/10.48550/arXiv.2010.14462
}
}

@Article{Bui-ThanhGhattasMartinEtAl13,
  Title                    = {A Computational Framework for Infinite-Dimensional {B}ayesian Inverse Problems {P}art {I}: {T}he Linearized Case, with Application to Global Seismic Inversion},
  Author                   = {Bui-Thanh, Tan and Ghattas, Omar and Martin, James and Stadler, Georg},
  Journal                  = {SIAM Journal on Scientific Computing},
  Year                     = {2013},
  Number                   = {6},
  Pages                    = {A2494-A2523},
  Volume                   = {35},

  Doi                      = {https://doi.org/10.1137/12089586X}
}

@Article{MartinWilcoxBursteddeEtAl12,
  Title                    = {A Stochastic {Newton MCMC} Method for Large-Scale Statistical Inverse Problems with Application to Seismic Inversion},
  Author                   = {Martin, James and Wilcox, Lucas C. and Burstedde, Carsten and Ghattas, Omar},
  Journal                  = {SIAM Journal on Scientific Computing},
  Year                     = {2012},
  Number                   = {3},
  Pages                    = {A1460-A1487},
  Volume                   = {34},

  Doi                      = {https://doi.org/10.1137/110845598}
}

@article{pratt1999seismic,
  title={Seismic waveform inversion in the frequency domain, Part 1: Theory and verification in a physical scale model},
  author={Pratt, R Gerhard},
  journal={Geophysics},
  volume={64},
  number={3},
  pages={888--901},
  year={1999},
  publisher={Society of Exploration Geophysicists},
doi={https://doi.org/10.1190/1.1444597}
}

@article{WarnerGuasch16,
author = {Warner, Michael  and  Guasch, Llu\'is},
title = {Adaptive waveform inversion: Theory},
journal = {GEOPHYSICS},
volume = {81},
number = {6},
pages = {R429-R445},
year = {2016},
doi = {https://doi.org/10.1190/geo2015-0387.1},



}

@article{yang2018application,
  title={Application of optimal transport and the quadratic Wasserstein metric to full-waveform inversion},
  author={Yang, Yunan and Engquist, Bj{\"o}rn and Sun, Junzhe and Hamfeldt, Brittany F},
  journal={Geophysics},
  volume={83},
  number={1},
  pages={R43--R62},
  year={2018},
  publisher={Society of Exploration Geophysicists},
doi={https://doi.org/10.1190/geo2016-0663.1}
}

@ARTICLE{FangshuJianwei23,

  author={Yang, Fangshu and Ma, Jianwei},

  journal={IEEE Transactions on Geoscience and Remote Sensing}, 

  title={Wasserstein Distance-Based Full-Waveform Inversion With a Regularizer Powered by Learned Gradient}, 

  year={2023},

  volume={61},

  number={},

  pages={1-13},

  doi={https://doi.org/10.1109/TGRS.2023.3241723}}

@article{LiVillaLiEtAl24,
title={Application of Learned Ideal Observers for Estimating Task-Based Performance Bounds for Computed Imaging Systems},
author={Li, Kaiyan and Villa, Umberto and Li, Hua and Anastasio, Mark},
journal={ Journal of Medical Imaging},
volume = {11},
number = {2},
publisher = {SPIE},
pages = {026002},
year = {2024},
doi = {https://doi.org/10.1117/1.JMI.11.2.026002},
URL = {https://doi.org/10.1117/1.JMI.11.2.026002}
}

@article{li21,
  title={Assessing the impact of deep neural network-based image denoising on binary signal detection tasks},
  author={Li, Kaiyan and Zhou, Weimin and Li, Hua and Anastasio, Mark A},
  journal={IEEE transactions on medical imaging},
  volume={40},
  number={9},
  pages={2295--2305},
  year={2021},
  publisher={IEEE},
doi = {https://doi.org/10.1109/TMI.2021.3076810}

}

@article{barrett93,
  title={Model observers for assessment of image quality.},
  author={Barrett, Harrison H and Yao, Jie and Rolland, Jannick P and Myers, Kyle J},
  journal={Proceedings of the National Academy of Sciences},
  volume={90},
  number={21},
  pages={9758--9765},
  year={1993},
  publisher={National Acad Sciences},
doi={https://doi.org/10.1073/pnas.90.21.9758}
}

@article{christianson15,
  title={An improved index of image quality for task-based performance of CT iterative reconstruction across three commercial implementations},
  author={Christianson, Olav and Chen, Joseph JS and Yang, Zhitong and Saiprasad, Ganesh and Dima, Alden and Filliben, James J and Peskin, Adele and Trimble, Christopher and Siegel, Eliot L and Samei, Ehsan},
  journal={Radiology},
  volume={275},
  number={3},
  pages={725--734},
  year={2015},
  publisher={Radiological Society of North America},
doi={https://doi.org/10.1148/radiol.15132091}
}

@article{zhang21,
  title={Impact of deep learning-based image super-resolution on binary signal detection},
  author={Zhang, Xiaohui and Kelkar, Varun A and Granstedt, Jason and Li, Hua and Anastasio, Mark A},
  journal={Journal of Medical Imaging},
  volume={8},
  number={6},
  pages={065501--065501},
  year={2021},
  publisher={Society of Photo-Optical Instrumentation Engineers},
doi={
https://doi.org/10.48550/arXiv.2107.02338}
}

@article{adler22,
  title={Task adapted reconstruction for inverse problems},
  author={Adler, Jonas and Lunz, Sebastian and Verdier, Olivier and Sch{\"o}nlieb, Carola-Bibiane and {\"O}ktem, Ozan},
  journal={Inverse Problems},
  volume={38},
  number={7},
  pages={075006},
  year={2022},
  publisher={IOP Publishing},
doi={https://doi.org/10.1088/1361-6420/ac28ec}
}

@article{he2013model,
  title={Model observers in medical imaging research},
  author={He, Xin and Park, Subok},
  journal={Theranostics},
  volume={3},
  number={10},
  pages={774},
  year={2013},
  publisher={Ivyspring International Publisher},
doi={https://doi.org/10.7150/thno.5138}

}

@inproceedings{
  song2021scorebased,
  title={Score-Based Generative Modeling through Stochastic Differential Equations},
  author={Yang Song and Jascha Sohl-Dickstein and Diederik P Kingma and Abhishek Kumar and Stefano Ermon and Ben Poole},
  booktitle={International Conference on Learning Representations},
  year={2021},
  url={https://openreview.net/forum?id=PxTIG12RRHS},
doi={
https://doi.org/10.48550/arXiv.2011.13456}
}

@article{neal2011mcmc,
  title={{MCMC} using Hamiltonian dynamics},
  author={Neal, Radford M and others},
  journal={Handbook of markov chain monte carlo},
  volume={2},
  number={11},
  pages={2},
  year={2011},
  publisher={Chapman and Hall/CRC},
doi={https://doi.org/10.1201/b10905}
}

@article{kodali2017convergence,
  title={On convergence and stability of {GAN}s},
  author={Kodali, Naveen and Abernethy, Jacob and Hays, James and Kira, Zsolt},
  journal={arXiv preprint arXiv:1705.07215},
  year={2017}
}

@article{srivastava2017veegan,
  title={Veegan: Reducing mode collapse in {GAN}s using implicit variational learning},
  author={Srivastava, Akash and Valkov, Lazar and Russell, Chris and Gutmann, Michael U and Sutton, Charles},
  journal={Advances in neural information processing systems},
  volume={30},
  year={2017}
}

@article{yacoby2020failure,
  title={Failure modes of variational autoencoders and their effects on downstream tasks},
  author={Yacoby, Yaniv and Pan, Weiwei and Doshi-Velez, Finale},
  journal={arXiv preprint arXiv:2007.07124},
  year={2020}
}
\bibliographystyle{IEEEtran}

\end{document}